\documentclass[Afour,sageh,times]{sagej}

\usepackage{moreverb,url}

\usepackage{graphics} 
\usepackage{subfigure}
\usepackage{epsfig} 
\usepackage{amsmath} 
\usepackage[T1]{fontenc}
\usepackage{bm}
\usepackage{cite}
\usepackage{booktabs}
\usepackage{xcolor}
\usepackage{multirow}
\usepackage{diagbox}
\usepackage{threeparttable}
\usepackage{colortbl}
\usepackage{algorithm}
\usepackage{algpseudocode}
\usepackage{stfloats}
\usepackage{makecell}
\usepackage{tabularx}
\usepackage{tablefootnote}
\usepackage{fixltx2e}
\usepackage{siunitx}
\usepackage{color}
\usepackage{amsthm}
\usepackage{paralist}
\usepackage[bookmarksopen,bookmarksnumbered,citecolor=red,urlcolor=red]{hyperref}
\hypersetup{
    hidelinks,
    }
\newcommand \naline {\raisebox{0.5ex}{{\rule{0.6cm}{0.4pt}}}}
\algrenewcommand{\algorithmiccomment}[1]{$//$ #1}

\DeclareMathOperator*{\argmin}{arg\,min}

\newcommand\BibTeX{{\rmfamily B\kern-.05em \textsc{i\kern-.025em b}\kern-.08em
T\kern-.1667em\lower.7ex\hbox{E}\kern-.125emX}}

\def\volumeyear{2023}

\begin{document}

\runninghead{Jiang et al.}

\title{{A Tactile-Informed Bi-Level Framework with Complementary Motion-Contact Planning and Tracking for Robust Model-Based In-Hand Manipulation}}

\author{Yongpeng Jiang\affilnum{1}, Mingrui Yu\affilnum{1}, Yi Ren\affilnum{2} and Xiang Li\affilnum{1}}

\affiliation{\affilnum{1}Department of Automation, Tsinghua University, Beijing, China\\
\affilnum{2}Advanced Manufacturing Lab, Huawei Technologies, Shenzhen, China}

\corrauth{Xiang Li, Department of Automation, Tsinghua University, Beijing, China}

\email{xiangli@tsinghua.edu.cn}

\begin{abstract}
Robotic dexterous in-hand manipulation, where multiple fingers dynamically make and break contact, represents a step toward human-like dexterity in robotic applications.
{Learning-based methods achieve strong task performance with increasing generalization to unseen objects. A complementary advantage of model-based methods, central to this work, is online adaptation to new task specifications without policy retraining, given suitable system models.}
However, due to the complexity of physical contacts and limited sensing modalities, existing model-based methods still struggle with modeling errors, even when using advanced contact models or optimization schemes, which limit their real-world applications.
To advance the effectiveness and robustness of model-based contact-rich in-hand manipulation, this paper proposes a novel bi-level framework with tactile incorporation to mitigate these limitations.
Specifically, at the high level, reference finger motions and contacts are jointly generated via contact-implicit model predictive control. 
The high-level module enables online reactive planning and disturbance recovery. At the low level, these references are concurrently tracked through a hand force-motion model and tactile feedback. The low-level module compensates for modeling errors and enhances manipulation robustness.
The complementarity between the two levels is manifested in two dimensions: 1) subsystem decomposition, with a slow optimization-based planning layer and a fast tactile-reactive control layer; and 2) functional decomposition, combining task-level optimality with contact-level stability.
Extensive experiments demonstrate that our approach outperforms existing model-based methods in accuracy and robustness, {while supporting online replanning during execution}. 
Our method successfully completes a diverse set of challenging tasks in real-world environments, even under appreciable external disturbances and modeling uncertainties.
Videos and code are available on the project website \href{https://director-of-g.github.io/in\_hand\_manipulation\_2/}{\textcolor{cyan!70!black}{https://director-of-g.github.io/in\_hand\_manipulation\_2/}}.

\end{abstract}


\maketitle

\section{Introduction}
\label{sec: introduction}
In-hand manipulation refers to changing the pose of a grasped object using finger, palm, and even external contacts.
This ability has become increasingly important recently in areas of robotics research and applications \citep{billard2019trends, cruciani2020benchmarking, yu2022dexterous, ma2011dexterity, gu2025humanoid}.
{The dexterity, robustness, and generalization of in-hand manipulation—the foundation of physical interactions—directly determine the range of tasks a robot can perform and its practical value.}

{In-hand manipulation involves the fingers \textit{making and breaking contacts} with (i.e., regrasping) the object \citep{sundaralingam2018geometric, qi2023general}, which is challenging for two reasons.}
First, real-time planning is crucial for in-hand manipulation due to external disturbances, such as unmodeled human interventions and stochastic contact dynamics, as noted by \citet{rostel2023estimator}.
However, achieving real-time planning is particularly challenging due to high degrees of freedom (DoFs) in multi-fingered robotic hands and the complex coordination required among numerous contacts. Second, modeling errors are inevitable during in-hand manipulation, because it is difficult to acquire both an accurate and efficient model for nonlinear and hybrid contact-rich dynamics \citep{le2024contact, todorov2012mujoco, makoviychuk2021isaac}. {The issue is further compounded by sensor noise and the challenge of generalizing across a wide range of object geometries and hand structures.}

\begin{figure*}
    \centering
    \includegraphics[width=0.95\linewidth]{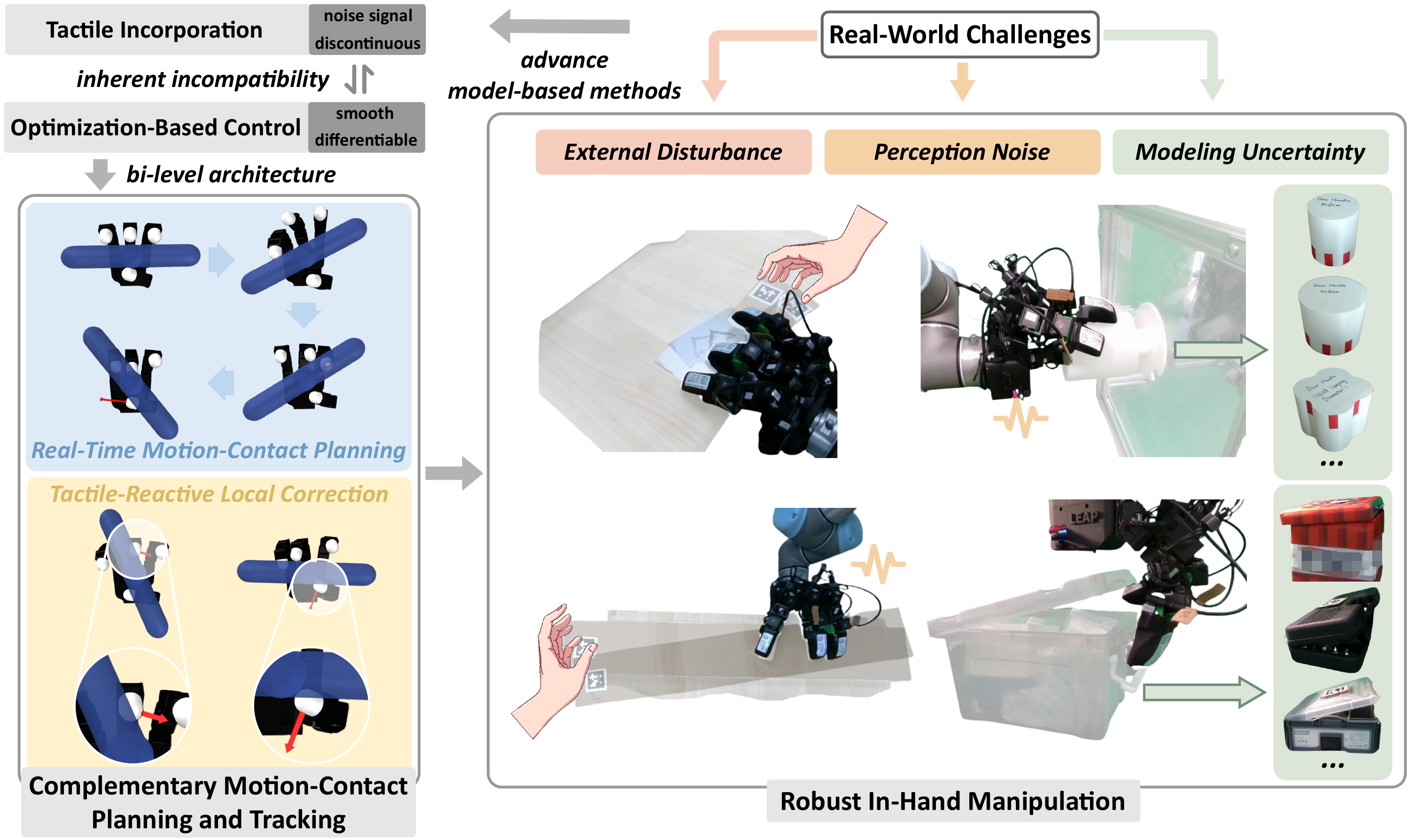}
    \caption{The proposed tactile-informed bi-level framework advances model-based in-hand manipulation under real-world challenges, including external disturbances, perception noise, and modeling uncertainties. With a bi-level architecture, it explicitly resolves the incompatibilities between tactile incorporation and optimization-based control. The framework couples real-time motion–contact planning with tactile-informed local corrections, enabling reactive, human-like finger gaiting, while mitigating modeling errors and ensuring contact-level stability. This combination results in robust, precise, and multi-task in-hand manipulation.}
    \label{fig: teaser}
\end{figure*}

Existing research to address these challenges can be broadly classified into learning-based and model-based paradigms.
Reinforcement learning (RL) shows state-of-the-art performance due to {multi-modal sensing} and domain randomization \citep{qi2023general, rostel2023estimator, chen2023visual, pitz2024learning, yang2024anyrotate, chen2024objectcentric}. Imitation Learning (IL) alleviates the difficulties of sim-to-real transfer by enabling real-world demonstrations, though acquiring them without haptic devices remains challenging \citep{guzey2023dexterity, si2024tilde}.
{Learning-based methods achieve strong task performance and increasingly generalize to unseen objects, although adapting policies across heterogeneous tasks may still require additional data and training.}

{In contrast, model-based methods offer a complementary advantage: with suitable models, they can adapt to new task specifications through online inference without policy retraining.}
These methods can be categorized as \textit{contact-explicit} and \textit{contact-implicit} methods, depending on whether contact sequences (i.e., contact locations, modes, and forces) are explicitly expressed as variables/constraints or implicitly encoded through general contact modeling.
\textit{Contact-explicit} methods transform contact-based manipulation into a hybrid problem that combines discrete contact sequences with continuous control \citep{cruciani2018dexterous, chen2021trajectotree, cheng2023enhancing, zhu2023efficient, sleiman2023versatile}, leading to combinatorial complexity.
Such methods are sensitive to perturbations during deployment due to the time-consuming replanning.
In contrast, \textit{contact-implicit} methods 
determine contact sequences implicitly by enforcing physics-induced complementarity constraints \citep{posa2014direct, pang2023global, kurtz2023inverse, aydinoglu2024consensus}, enabling unified reasoning over motion and contact, thus allowing for more flexibility.
To ensure real-time performance, these methods often adopt relaxed models, avoiding the use of exact but difficult-to-solve complementarity constraints, which may reduce physical fidelity. 
Consequently, such methods remain susceptible to modeling errors.

Despite the success of model-based methods in simulation, real-world contact-rich manipulation must cope with perception noise, modeling errors, and other uncertainties that cannot be resolved solely through improved contact modeling or more advanced optimization. Following insights from RL, we argue that incorporating tactile sensing to endow the policy with local corrective capability is a natural and promising step toward human-like in-hand manipulation. Although combining model-based control with tactile sensing is not entirely new, prior work has largely been limited to single-task settings \citep{hogan2020tactile, zhao2024tac, Yu2024InHandFO, xu2024letac}, and a systematic, multi-task in-hand manipulation framework with both theoretical significance and practical value for real-world applications remains lacking.

To this end, this paper proposes a bi-level framework that unifies real-time motion-contact planning with tactile-reactive tracking control (Fig.~\ref{fig: teaser}). We emphasize that the proposed framework is neither a trivial extension of existing approaches, nor a simple cascade of two modules. The proposed framework is a system-level methodology aimed at enabling {robust and precise in-hand manipulation with online replanning} under real-world challenges. Specifically, at a high level, we formulate a contact-implicit model predictive control (CIMPC) scheme to plan reference finger motions and desired contacts in real time. At a low level, the synergistic motion-contact references are tracked using MPC-based hybrid force-motion control (HFMC). This bi-level architecture yields two complementary subsystems: a slow reasoning layer that determines \textit{how desired object motion should be realized through coordinated, contact-rich interactions}, and a fast corrective layer that determines \textit{how desired contacts should be established and maintained through fine motor control}. Through this decomposition, the mismatch between nominal and real dynamics is significantly alleviated. The bi-level architecture further enables a functional decomposition. Specifically, the high-level module maintains task-level optimality by reasoning over long-term contact events and finger-gaiting patterns, while the low-level module ensures contact-level stability by compensating for modeling uncertainties through short-horizon motion corrections toward the desired contacts. Besides, we introduce several additional innovations—warm starting, trajectory shifting, multi-contact coupling, and adaptive weighting—that effectively coordinate the two levels and balance multiple tracking objectives with minimal cross-level interference.

Our key contributions are as follows.

\begin{enumerate}
    \item {To enable online reactive planning for in-hand manipulation, we propose a unified optimization framework that structurally couples an implicit contact model, task-oriented cost functions, and a customized DDP solver.} This approach adaptively discovers human-like finger-gaiting patterns by reasoning over long-term contact events.
    \item {To mitigate modeling errors from mathematical simplifications and multi-source uncertainties, we propose an MPC-based hybrid force--motion controller that uses tactile feedback to correct finger motions in real time, aligning actual and desired contacts.} We further introduce adaptive weighting that modulates force--motion tracking priorities for effective corrective motion during dynamic in-hand manipulation.
    \item Leveraging 1 and 2, we propose a bi-level framework for in-hand manipulation that unifies motion-contact co-planning with tactile-reactive local correction. The proposed framework benefits from both task-level optimality and contact-level stability.
    \item Through extensive simulations and real-world experiments, we show that the proposed framework {not only achieves high manipulation accuracy but also demonstrates strong robustness to modeling uncertainties and external perturbations, thereby outperforming existing model-based approaches. These strengths are comprehensively validated in diverse practical real-world scenarios.}
\end{enumerate}

{This work builds upon our previous work \citep{jiang2024contact}.
The improvements include: 1) A more systematic and rigorous development of the proposed bi-level framework, with its key advances stemming from new innovations that significantly improve cross-level coordination. 2) Further refining the proposed framework to enhance multi-contact coordination, adaptability to arbitrary object geometries, and {online planning capability}.
3) First-time real-world experiments on diverse in-hand manipulation tasks, crucial for validating the framework’s innovations beyond the conference version \citep{jiang2024contact}. 4) Conducting comprehensive simulation studies that include more rigorous and fair comparisons with additional state-of-the-art methods.
Videos and code are available on the project website\footnote{https://director-of-g.github.io/in\_hand\_manipulation\_2/}.
} 

The remainder of the paper is organized as follows. We begin with a review of related work (Sec.~\ref{sec: ralated_work}) and preliminaries (Sec.~\ref{sec: preliminaries}). We then provide an overview of the proposed framework (Sec.~\ref{sec: overview}), highlighting its key innovations, followed by a detailed discussion of {the real-time motion-contact planning} (Sec.~\ref{sec: high_level}) and {tactile-reactive tracking control} (Sec.~\ref{sec: low_level}) sub-modules. Comprehensive simulation and real-world results are presented in Sec.~\ref{sec: simulation_results} and Sec.~\ref{sec: real_world_experiments}. Finally, we discuss the framework's limitations and present conclusions in Sec.~\ref{sec: discussion_and_conlusion}.

\section{Related Work}
\label{sec: ralated_work}
\subsection{Contact-Rich In-Hand Manipulation}

This section reviews the existing approaches to in-hand manipulation. We discuss learning-based approaches with RL or IL and model-based approaches, including contact-explicit and contact-implicit methods.

\subsubsection{Learning-Based Approaches With RL or IL.}

RL-based approaches to in-hand manipulation typically involve offline training followed by real-world deployment \citep{chen2023visual, handa2023dextreme}. For this task, most RL methods adopt a model-free design. RL is further combined with tactile sensing to train multi-modal policies, which greatly alleviates the problem of visual occlusion {and enhances the robustness} \citep{qi2023general, pitz2024learning, yang2024anyrotate, suresh2024neuralfeels}. To improve data efficiency, sampling-based planning has been applied to assist exploration in the complex state space \citep{Khandate2023SamplingbasedEF}. Real-world RL \citep{luo2024precise} and IL \citep{chi2023diffusion} are possible future roadmaps for data-efficient learning, but their performance in dexterous in-hand manipulation remains to be determined. Despite recent breakthroughs in imitation policy learning, data collection remains an issue, considering the embodiment gap between human and robot \citep{qin2022dexmv, handa2020dexpilot, arunachalam2023dexterous}.
{Learning-based methods offer strong task performance and robustness, and recent approaches have also demonstrated zero-shot generalization to unseen objects \citep{qi2023general, yang2024anyrotate}. However, extending them across heterogeneous task specifications or setups may still require considerable adaptation effort, such as additional data collection or policy retraining.}

\subsubsection{Model-Based Approaches: Contact-Explicit Methods.}

{Several works predefine contact sequences using prior knowledge or solve them explicitly via mixed integer programming, exploiting the hybrid nature of contact-rich manipulation. The specific methods for determining contact sequences are highly diverse.} In \citet{cheng2023enhancing} and \citet{zhu2023efficient}, the authors employ Monte Carlo Tree Search (MCTS) with different heuristics to improve efficiency. In \citet{cheng2022contact} and \citet{dafle2018hand}, the authors exploited special structures of the constrained manifold to guide sampling-based planners under multiple contacts. In \citet{cruciani2018dexterous}, Cruciani et al. proposed a graph search method to plan the contact sequence 
on a Dexterous Manipulation Graph (DMG). With the planned contacts, the motion planning problem is simplified to determining the continuous control inputs under explicit constraints \citep{sleiman2023versatile}, such as ensuring force closure and maintaining specific contact modes. The motion planner can be optimization based \citep{hou2019robust, chen2021trajectotree} or hand crafted \citep{cheng2023enhancing, dafle2018hand, cruciani2018dexterous}. Such methods typically yield near-complete solutions, which is advantageous in planning problems.
{By explicitly separating contact-sequence planning from motion planning, the control problem can be formulated with explicit contact constraints, thereby reducing modeling errors.} However, solving the hybrid optimization is time consuming, posing significant challenges to real-time planning performance and robustness against disturbances.

\subsubsection{Model-Based Approaches: Contact-Implicit \texorpdfstring{Meth-\\ods.}{Methods.}}

{Several works do not explicitly solve contact sequences; instead, contact events are implicitly inferred through physics-based contact models during optimization.}
The original form of implicit contact modeling includes intractable complementarity constraints \citep{posa2014direct}. To improve real-time performance, Aydinoglu et al. proposed an operator splitting framework for parallelizing the inference of contact events \citep{aydinoglu2024consensus}. 
In addition, smoothing has been identified as an important procedure for efficient contact exploration; this includes specific approaches such as randomized smoothing \citep{pang2023global}, analytic smoothing \citep{pang2023global, Kim2023ContactimplicitMP, le2024fast}, and using softened contact models \citep{onol2019contact, kurtz2023inverse}. Using smoothing techniques, Pang et al. proposed a convex optimization-based dynamics model for general contact-rich manipulation planning \citep{pang2023global}.
{However, due to the hybrid and non-smooth nature of contact modeling, incorporating implicit contact models into an optimization framework is computationally expensive.}
To address this problem, Yang et al. approximated the dynamics model with signed distance functions (SDFs), leading to explicit time stepping \citep{yang2024contactsdf}. \citet{jin2024complementarity} further proposed a complementarity-free simplification. Recently, Zhang et al. extended contact-implicit methods to complex-shaped objects, but they did not demonstrate real-time control performance \citep{zhang2024simultaneous}.
Contact-implicit methods enable {unified} planning of motion and contact. However, they can introduce appreciable modeling errors, including the force-at-a-distance effect \citep{anitescu2006optimization, pang2023global, kurtz2023inverse, fazeli2017parameter}, due to model simplifications and approximations. The proposed method addresses these issues.

In addition to gradient-based optimization, predictive sampling \citep{howell2022predictive} (PS) has recently emerged as a simple yet effective approach for in-hand manipulation. This method samples and rolls out actions while enforcing contact constraints via a physics engine.
{The key advantage of this method is its computational efficiency, enabled by modern parallel simulation \citep{todorov2012mujoco, howell2022predictive}.}
Li et al. performed an in-hand cube reorientation task in the real world with PS \citep{li2024drop}. 
Hess et al. applied the method to a tendon-driven hand for in-hand ball manipulation \citep{hess2024sampling}. 
By adopting simulation-based dynamics, in-hand manipulation can be performed with dexterous hands and objects that were previously hard to model.
However, sampling-based planning generates actions with high variance, which can increase internal disturbances. {In addition, it is sensitive to discrepancies between simulated and real dynamics, which may degrade real-world performance.}

\subsection{Force-Aware Execution in Contact-Rich Manipulation}

Building on the above review of contact-rich in-hand manipulation, this subsection takes a broader view of force-aware execution, focusing on how force or tactile feedback regulates contact interactions and improves contact-level stability. In robotic interaction tasks, contact force regulation led to pure force control and hybrid force-motion control strategies \citep{gold2022model}. In dexterous manipulation, explicit force control and tactile servoing have been used for grasp adjustment, stable pinching, and uncertainty compensation \citep{liu2022multi,psomopoulou2021robust,khadivar2023adaptive}, while tactile feedback has also been exploited to build closed-loop manipulation primitives \citep{hogan2020tactile}. More methodologically close to our low-level controller, optimization-based methods embed interaction-force prediction into MPC or optimal control for force-closure grasping and hybrid force-motion behaviors \citep{gold2021model,gold2022model}. \citet{xu2024letac} further combined tactile representations with a differentiable MPC layer for tactile-reactive grasping. These works align with our method in formulating contact-force regulation as closed-loop feedback or optimization, but mainly focus on grasping or general robot-environment interaction, while our controller extends this idea to in-hand manipulation with adaptive force-motion weighting.

Recent learning-based methods also incorporate force awareness into dexterous manipulation policies. Tactile-feedback policies are related to our method in using contact feedback to stabilize manipulation, although force regulation is usually implicit in the learned policy \citep{guzey2023dexterity,qi2023general,yang2024anyrotate}. More closely aligned with our force-motion formulation, DexForce constructs impedance-based force-informed targets from measured contact forces \citep{chen2025dexforce}, while KineDex learns from tactile-enriched kinesthetic demonstrations and deploys policies with force control \citep{zhang2025kinedex}. These methods share the goal of complementing motion commands with force-aware execution, but their force behaviors or references are learned from task-specific data. In contrast, our controller computes motion and contact-force references from a generic model-based formulation and tracks them explicitly with MPC, without task-specific data collection or policy training.

\section{Preliminaries}
\label{sec: preliminaries}
\subsection{Assumptions and Notations}

This paper considers the task of quasi-dynamic in-hand manipulation through rigid frictional contacts.
We make the following assumptions in this paper: 1) In-hand manipulation is performed under \textit{quasi-dynamic} conditions \citep{mason2001mechanics}, which means that the system exhibits negligible inertia effects and is characterized by first-order dynamics. 2) The object is modeled as a single rigid body.
{The geometries of the object and robot hand are known}, whereas the inertia and contact parameters are approximately estimated.
Assumption 1) is reasonable because it is consistent with many daily objects due to frictional surfaces, damped hinges, or slow motions.
Assumption 2) can be relaxed to multi-body or soft objects with task-specific perception, which is beyond the scope of this paper.

In this paper, we use the superscripts $a$ and $u$ for the robot (actuated) and object (unactuated). In addition, $\Vert \cdot \Vert_{\bm{W}}$ is the weighted quadratic norm, $\left[\bm{a}; \bm{b}\right]$ is the vertical concatenation of vectors $\bm{a}$ and $\bm{b}$, $\left[\bm{A};\bm{B}\right]$ is the vertical stacking of matrices $\bm{A}$ and $\bm{B}$, and $\text{blkdiag}(\bm{A}_1,\cdots,\bm{A}_m)$ is the block diagonal matrix comprising $\bm{A}_1,\cdots,\bm{A}_m$. The operators $\oplus,\ominus$ denote SE(3) addition and subtraction; increments and derivatives use local coordinates. Several variables that can be easily confused are summarized in {Tab.}~\ref{tab: notations}.
\begin{table}[!t]
	\caption{Notations Used in this Paper}
        \label{tab: notations}
	\begin{tabularx}{\linewidth}{p{0.12\linewidth}X}
		\toprule	
		\underline{\textbf{\emph{High-Level Module}}} \\
		$\bm{x}$ & state of the full hand-object system $[\bm{x}^u;\bm{x}^a]$  \\
            $\bm{x}^u$ & object pose (the orientation is parameterized by quaternion) \\
            $\bm{x}^a$ & hand joint positions \\
            $\bm{x}^{+}$ & state of the next time step \\
            $\bm{x}_\text{init}$ & initial state of each optimization iteration \\
            $\bm{u}$ & control input (i.e., delta desired joint positions) \\
            $\bm{J}$ & contact Jacobian of the full hand-object system \\
            $N$    & prediction horizon of the discrete system \\
            $h$    & time step of the discrete system \\
		\underline{\textbf{\emph{Low-Level Module}}} \\  
		$\bm{s}$ & low-level state $[\bm{q};\bm{q}_d;\bm{\Lambda}_\text{ext}]$ \\
            $\bm{q}, \bm{q}_d$ & actual and commanded joint positions \\
            $\bm{\lambda}_\text{ext}, \bm{\Lambda}_\text{ext}$ & contact force \\ & (i.e., of a single contact and all contacts) \\
            $\bm{u}$ & control input \\ & (i.e., the change rate $\dot{\bm{q}}_d$ of joint commands) \\
            $\bar{\bm{K}}_s, \bar{\bm{K}}$ & equivalent stiffness \\ & (i.e., of a single contact $\bar{\bm{K}}_s$ and all contacts $\bar{\bm{K}}$) \\
            $\bm{J}_s, \bm{J}$ & hand kinematic Jacobian \\ & (i.e., of a single contact $\bm{J}_s$ and all contacts $\bm{J}$) \\
            $\bm{G_o}$ & grasp matrix \\
            $T$    & prediction horizon of the continuous system \\
            $\Delta t$ & sampling time of the continuous system \\
		\underline{\textbf{\emph{Others}}} \\ 
		$n_c$  & number of active contacts \\
            $n_q$  & number of hand joints \\
            $n_u$  & number of control inputs \\
		\bottomrule
	\end{tabularx}
\end{table}

\subsection{Implicit Dynamics Model with Contacts}
\label{subsec: implicit_contact_model_preliminaries}

Pang et al. proposed an implicit time stepping for general multi-contact systems \citep{pang2023global}. Under quasi-dynamic assumptions, the system dynamics is constructed as the Karush-Kuhn-Tucker (KKT) conditions of the following Second-Order Cone Program (SOCP):
\begin{equation}
    \begin{aligned}
        \delta\bm{x}^{*}=\argmin_{\delta\bm{x}} \, & \frac{1}{2}\delta\bm{x}^\top\bm{Q}\delta\bm{x}+h\cdot\bm{b}^\top\delta\bm{x} \\
        \mbox{s.t.}\quad
        & \bm{J}_i\delta\bm{x}
        +\left[
            \begin{matrix}
                \phi_i \\
                \bm{0}_2
            \end{matrix}
        \right]\in \mathbf{FC}^{*}(\mu_i), \forall i\in\{1,\dots,n_c\} \\
    \end{aligned}
    \label{eq: CQDC_model_SOCP_form}
\end{equation}
where $h$ is the discrete time step, $n_c$ is the number of active contacts, and $\mathbf{FC}^{*}(\mu)=\left\{(\alpha,\bm{\beta})\in\mathbb{R}\times\mathbb{R}^{2}\vert\mu\Vert\bm{\beta}\Vert_2\leq\alpha\right\}$ is the dual friction cone with friction coefficient $\mu$. The contact distance $\phi(\bm{x})$ and contact Jacobian $\bm{J}(\bm{x})$ are computed with collision detection.
In addition, $\bm{Q}$ and $\bm{b}(\bm{x},\bm{u})$ are computed using the robot stiffness and object inertia (i.e., model parameters).
The implicit time stepping is derived as $\bm{x}^{+}=\bm{f}(\bm{x},\bm{u})=\bm{x}\oplus\delta\bm{x}(\bm{x},\bm{u})$. In this paper, we refer to $\bm{f}(\bm{x},\bm{u})$ as the CQDC model. Note that the SOCP form (\ref{eq: CQDC_model_SOCP_form}) simplifies the actual contact dynamics.

\subsection{Differential Dynamic Programming (DDP)}

DDP is an iterative algorithm for solving optimal control problems (OCPs) using the concept of dynamic programming \citep{tassa2014control}. Consider the following OCP with finite horizon and discrete-time dynamics:
\begin{equation}
    \begin{aligned}
        \min_{\bm{U}} \quad & \sum_{i=0}^{N-1}l(\bm{x}_i,\bm{u}_i)+l_f(\bm{x}_N) \\
        \mbox{s.t.} \quad & \bm{x}_{i+1}=\bm{f}(\bm{x}_i,\bm{u}_i)
    \end{aligned}
    \label{eq: DDP_problem}
\end{equation}
The objective function comprises running costs $l$ and the final cost $l_f$. DDP enforces the dynamics constraint by computing the full state trajectory $\bm{X}=\{\bm{x}_i\}_{i=0}^{N}$ from the integration of $\bm{f}$ with the control trajectory $\bm{U}=\{\bm{u}_i\}_{i=0}^{N-1}$. Integration starts from the known initial state $\bm{x}_0$. DDP iteratively corrects $\bm{U}$ with forward and backward passes.

A state feedback law is derived in the backward pass:
\begin{equation}
    \delta\bm{u}^{*}(\delta\bm{x})=\argmin_{\delta\bm{u}}P(\delta\bm{x},\delta\bm{u})=\bm{k}+\bm{K}\delta\bm{x}
    \label{eq: DDP_state_feedback}
\end{equation}
where $P$ is the discrete-time analogue of the Hamiltonian,
$\bm{k}=-P_{\bm{uu}}^{-1} P_{\bm{u}}$ and $\bm{K}=-P_{\bm{uu}}^{-1} P_{\bm{ux}}$ are constructed from its partial derivatives.
The backward pass starts from step $N-1$. During the backward pass, the costs and dynamics are differentiated around the trajectory $\hat{\bm{X}}, \hat{\bm{U}}$ solved during the last iteration. These derivatives are needed in the computation of $\{P_{\bm{x}}, P_{\bm{u}}, P_{\bm{xx}}, P_{\bm{ux}}, P_{\bm{uu}}\}$.

The forward pass then updates the trajectory $\hat{\bm{X}}, \hat{\bm{U}}$ with a control law based on (\ref{eq: DDP_state_feedback}):
\begin{equation}
    \begin{aligned}
        \hat{\bm{x}}_0 &= \bm{x}_0 \\
        \hat{\bm{u}}_i &= \bm{u}_i+\alpha\bm{k}_i+\bm{K}_i(\hat{\bm{x}}_i\ominus\bm{x}_i) \\
        \hat{\bm{x}}_{i+1} &= \bm{f}(\hat{\bm{x}}_i,\hat{\bm{u}}_i), \quad \forall i \in \{0,\dots,N-1\}
    \end{aligned}
    \label{eq: DDP_forward_pass}
\end{equation}
where $\bm{x}_i,\bm{u}_i$ are solved state and control obtained from the last iteration.

Note that the original DDP requires the costs and dynamics to be second-order differentiable, which does not always hold in contact-rich manipulation as piecewise continuous functions are common in physical contact models. 
In practice, it is reasonable to neglect the Hessians of $\bm{f}$ using the Gauss-Newton variant of DDP, which aligns with the iLQR approach
\citep{tassa2014control}.

\section{Overview of the Proposed Framework}
\label{sec: overview}
\begin{figure*}[ht]
    \centering
    \includegraphics[width=1.0\linewidth]{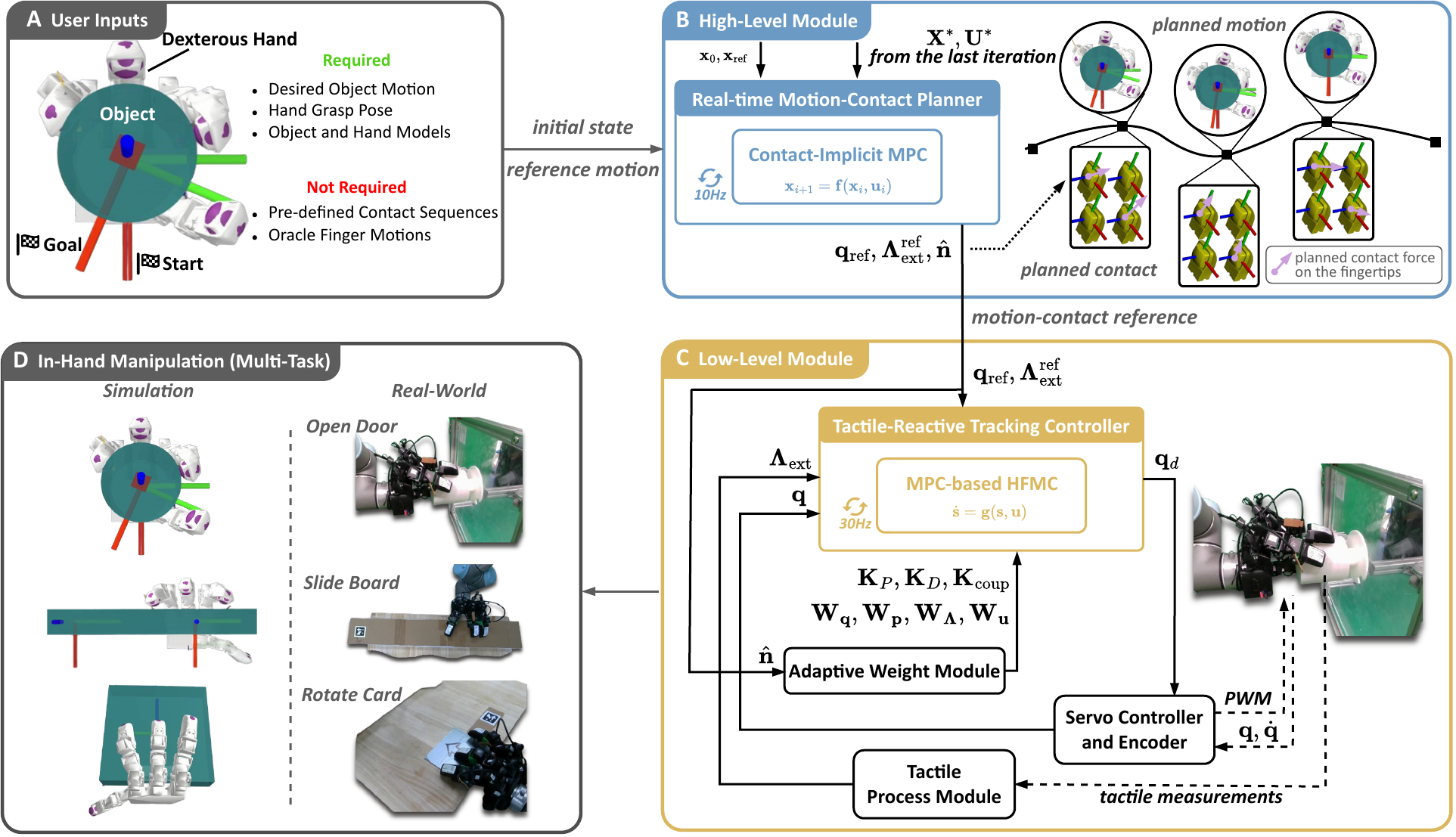}
    \caption{Proposed bi-level framework for {robust in-hand manipulation with online motion-contact replanning} toward real-world challenges. (A) The user inputs the desired object motion, hand grasp pose (i.e., nominal configuration), and corresponding models. (B) The high-level motion-contact planner employs contact-implicit MPC to generate motion-contact references from the initial state $\bm{x}_0$, state reference $\bm{x}_\text{ref}$, and the previous iteration's solution $\bm{X}^{*}, \bm{U}^{*}$. (C) The low-level tactile-reactive tracking controller leverages adaptive-weighted MPC-based HFMC to synergistically track motion–contact references with minimal mutual interference. (D) Both levels complement each other, forming a systematic, tactile-inclusive, multi-task framework for in-hand manipulation.}
    \label{fig: framework_overview}
\end{figure*}

\begin{figure}[t]
    \centering
    \includegraphics[width=1.0\linewidth]{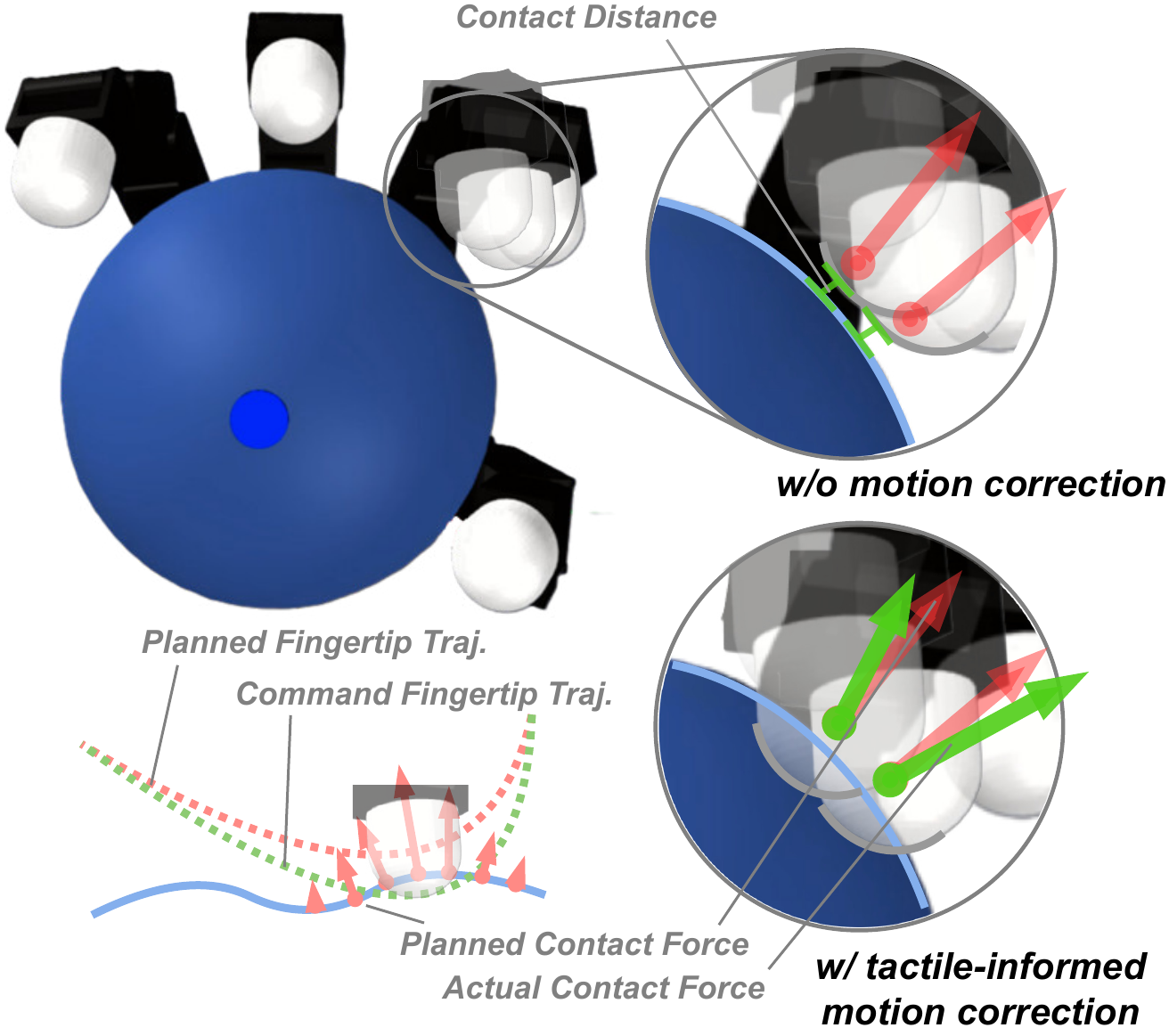}
    \caption{Illustration of the inter-level coordination. The top figures show the high-level motion-contact planner, which {generates finger-gaiting patterns online} with embedded contact information (visualizing only the index finger and contact forces). The right-side figures show results without (w/o) and with (w/) motion correction enabled by the proposed low-level controller. The top-right figure illustrates how modeling errors lead to the force-at-a-distance effect \citep{anitescu2006optimization}, where non-zero planned forces appear even when contact is inactive. Modeling errors can be mitigated through low-level motion-contact tracking with tactile incorporation (shown in the bottom-right figure). The bottom-left figure shows how motion correction is achieved by deforming the command fingertip trajectory (traj.). Best viewed in color.}
    \label{fig: example_reference_modification}
\end{figure}

{This section provides an overview of the proposed bi-level framework for dexterous in-hand manipulation, as illustrated in Fig.~\ref{fig: framework_overview}.} 
The framework takes as input the desired object motion, the hand grasp pose {(i.e., nominal configuration)}, and the models of both the object and hand.
{No predefined contact or oracle finger motions are needed.}
{The two levels, including a high-level real-time motion-contact planner and a low-level tactile-reactive controller, complement each other and are related through the motion-contact references.}

{We articulate the \textit{rationale} behind the proposed framework, covering its necessity, challenges, and contributions.}

{\textbf{Necessity}: Incorporating tactile information meaningfully enhances model-based in-hand manipulation. However, prior studies largely focus on single-task cases, and a systematic, multi-task framework is still lacking, which has high practical value in real-world applications. Moreover, state-of-the-art contact-implicit MPC is not directly compatible with tactile incorporation, motivating a bi-level formulation that can leverage tactile feedback without disrupting the underlying optimization structure.}

{\textbf{Challenges}: 1) \textit{Complex multi-contact dynamics}: High-dimensional, nonlinear, and nonsmooth system behaviors render real-time planning and updates of finger motion modalities computationally demanding. 2) \textit{Dynamics/modeling inconsistency}: Discrepancies between real and simulated dynamics—as well as mismatches between upper- and lower-level models—complicate coordination and increase compensation burden. 3) \textit{Inter-level coupling}: High-level delays and low-level tracking imperfections can induce oscillations and slow convergence due to the tight coupling between the two levels. 4) \textit{Motion-contact coordination}: Stable and responsive coordination of motion and contact tracking is challenging due to their nonlinear coupling via contact dynamics.}

{\textbf{Contributions}: 1) We propose a unified optimization framework that structurally couples an implicit contact model, computationally efficient cost functions, and a customized DDP solver. This framework effectively avoids being trapped in local optima and enables {online, adaptive generation of finger-gaiting patterns}. 2) We introduce a functional cross-level model decomposition in which the upper level uses an implicit contact model to reason about task-centric motion patterns, while the lower level applies an explicit high-frequency force–motion mapping for precise finger adjustments. This separation enables each level to operate within its respective strengths, thereby helping mitigate real–sim discrepancies. 3) We develop a coherent set of improvements—including a structured motion–contact reference between levels and stabilization strategies such as warm-starting and trajectory shifting—that address suboptimal convergence and reduce delays impeding inter-level coordination. 4) We propose an MPC-based hybrid force–motion controller with adaptive weighting, leveraging MPC’s soft-constraint formulation to naturally resolve conflicts between motion and contact objectives.}

{The coordination between the high- and low-level modules is illustrated in Fig.~\ref{fig: example_reference_modification}.}

\section{{Real-Time Motion-Contact Planning}}
\label{sec: high_level}
This section proposes a unified optimization framework for synergistically generating finger movements with embedded contact information (i.e., contact locations and contact forces) {for online in-hand manipulation}.

\subsection{Problem Formulation}

The adaptive and coordinated finger motions and contact information {are generated} based on desired object motion.
Such a real-time motion-contact planning process can be realized by solving the following optimal control problem (OCP) using a receding-horizon approach:
\begin{equation}
    \begin{aligned}
        \min_{\bm{X},\bm{U}} \quad & \sum_{i=0}^{N-1}l(\bm{x}_i,\bm{u}_i)+l_f(\bm{x}_N) \\
        \mbox{s.t.} \quad & \bm{x}_0=\bm{x}_\text{init} \\
        & \bm{x}_{i+1}=\bm{f}(\bm{x}_i,\bm{u}_i) \\
        & \bm{x}_i \in \mathcal{X}_\text{jnt} \cap \mathcal{X}_\text{sc} \\
        & \bm{u}_i \in \mathcal{U}
    \end{aligned}
    \label{eq: high_level_OCP}
\end{equation}
where $\bm{X}=\left\{\bm{x}_0,\cdots,\bm{x}_N\right\}, \bm{U}=\left\{\bm{u}_0,\cdots,\bm{u}_{N-1}\right\}$ are the state and control variables and $\bm{x}_\text{init}$ is the known initial state. The feasible sets are formulated as
\begin{equation}
    \begin{aligned}
        \mathcal{X}_\text{jnt} &= \left\{\bm{x} \vert \underline{\bm{x}^a} \leq \bm{x}^a \leq \overline{\bm{x}^a} \right\}, \\
        \mathcal{X}_\text{sc} &= \left\{\bm{x} \vert \mathcal{G}_1 \cap \mathcal{G}_2=\varnothing, \forall \mathcal{G}_1\ne\mathcal{G}_2 \in \mathrm{CG}(\bm{x}) \right\}, \\
        \mathcal{U} &= \left\{\bm{u} \vert \underline{\bm{u}} \leq \bm{u} \leq \overline{\bm{u}} \right\}
    \end{aligned}
    \label{eq: high_level_feasible_sets}
\end{equation}
where $\mathcal{X}_\text{jnt}$, $\mathcal{X}_\text{sc}$ represent the joint limits and the self-collision constraints, $\mathcal{U}$ denotes the input bounds, and $\mathrm{CG}(\bm{x})$ is the set of collision geometries.
{To implicitly encode and adaptively infer contact events,}
we employ the CQDC model (introduced in Sec.~\ref{subsec: implicit_contact_model_preliminaries}) as the system dynamics $\bm{f}$. When solved with DDP, the dynamics constraint is enforced with the forward integration of $\bm{f}$. In addition, $\mathcal{X}_\text{jnt}, \mathcal{X}_\text{sc}$ are converted to soft constraints for the purpose of DDP solving:
\begin{equation}
    \begin{aligned}
        l(\bm{x},\bm{u}) &= l_\text{reg}(\bm{x})+l_\text{reg}(\bm{u})+l_\text{jnt}(\bm{x})+l_\text{sc}(\bm{x}) \\
        l_f(\bm{x}_N) &= l_\text{reg}(\bm{x}_N)
    \end{aligned}
    \label{eq: high_level_cost_term_definition}
\end{equation}
The running cost $l(\bm{x},\bm{u})$ comprises the regulation costs of states and control, the cost of joint limits, and the cost of self-collision, whereas the final cost $l_f(\bm{x}_N)$ comprises only the regulation cost of states. The specific form of each cost term is introduced in Sec.~\ref{sec: cost_terms}.

\subsection{Contact-Implicit Dynamics Model}

\subsubsection{Smoothing of the Contact-Based Dynamics Model.}
Due to the nonsmooth contact modeling, directly solving (\ref{eq: CQDC_model_SOCP_form}) results in shortsighted outcomes that inadequately explore the possible contact modes. {Inspired by \citet{pang2023global}, the dynamics in (\ref{eq: CQDC_model_SOCP_form}) is smoothed in our framework using a barrier function, yielding the following formulation:}
\begin{equation}
    \begin{aligned}
        \delta\bm{x}^{*}=\argmin_{\delta\bm{x}} \quad &\frac{1}{2}\delta\bm{x}^\top\bm{Q}\delta\bm{x}+h\cdot\bm{b}^\top\delta\bm{x} \\
        -&\frac{1}{\kappa}\sum_{i=1}^{n_c}\log\left(z_{i,1}^{2}-\mu_i^{2}
        \left\Vert\begin{matrix}z_{i,2}\\z_{i,3}\end{matrix}\right\Vert_2^{2}\right)
    \end{aligned}
    \label{eq: CQDC_model_barrier}
\end{equation}
where $\bm{z}_i=\bm{J}_i\delta\bm{x}+[\phi_i;\bm{0}_2]$ with $z_{i,1}>0$, and $\kappa$ controls smoothing. Due to this smoothing, a small $\kappa$ (e.g., on the order of hundreds) can generate non-negligible forces even when the constraint remains inactive. {This phenomenon is known as the force-at-a-distance effect \citep{anitescu2006optimization}, which is commonly reported in many model-based methods \citep{suh2025dexterous, shirai2025linear}.}
Although beneficial for planning, it poses challenges for control by inducing sliding and missed contacts, which the proposed framework is designed to mitigate.

\subsubsection{Gradient Computation.}
An important property of the CQDC model is its differentiability, which is required by the backward pass of DDP.
{The partial derivatives with respect to states and controls are computed using sensitivity analysis and automatic differentiation.}
\begin{equation}
    \begin{aligned}
        \frac{\partial\bm{f}}{\partial\bm{x}}&=\bm{I}+\frac{\partial\delta\bm{x}}{\partial\bm{x}} \\
        &=\bm{I}+\frac{\partial\delta\bm{x}}{\partial\bm{b}}\frac{\partial\bm{b}}{\partial\bm{x}}+\sum_{i=1}^{n_c}\left(\frac{\partial\delta\bm{x}}{\partial\bm{J}_i}\frac{\partial\bm{J}_i}{\partial\bm{x}}+\frac{\partial\delta\bm{x}}{\partial\phi_i}\frac{\partial\phi_i}{\partial\bm{x}}\right) \\
        \frac{\partial\bm{f}}{\partial\bm{u}}&=\frac{\partial\delta\bm{x}}{\partial\bm{u}}=\frac{\partial\delta\bm{x}}{\partial\bm{b}}\frac{\partial\bm{b}}{\partial\bm{u}}
    \end{aligned}
    \label{eq: CQDC_model_derivative}
\end{equation}
{The partial derivatives $\frac{\partial \bm{J}}{\partial \bm{x}}, \frac{\partial \phi}{\partial \bm{x}}$ are computed using collision detection, while our framework imposes no assumptions on the objects’ geometries. Further algorithmic developments and performance results for handling concave and complex objects are presented in Appendix B.}

\begin{table*}[tb]
\centering
\caption{Definitions and purposes of cost terms in the high-level MPC.}
\label{tab: ddp_cost_terms_summary}
\renewcommand{\arraystretch}{1.5}
\setlength{\tabcolsep}{3pt}
\begin{threeparttable}[b]
\begin{tabular}{c|c|l} 
\toprule
\multicolumn{1}{c|}{\textbf{Cost}} & 
\multicolumn{1}{c|}{\textbf{Definition}} & 
\multicolumn{1}{c}{\textbf{Purpose}} \\
\hline
\multirow{2.5}{*}{$l_\text{reg}(\bm{x})$} &
    $\Vert \bm{x}^a-\bm{x}_\text{ref}^a \Vert_{\bm{W}_{\bm{x}}^a}^2$ &
    \makecell[l]{Restricts the fingers to local exploration, and guides \\ the fingers to break contacts.} \\
 &
    $\Vert \bm{x}^u\ominus\bm{x}_\text{ref}^u \Vert_{\bm{W}_{\bm{x}}^u}^2$ &
    Achieves the desired object motion. \\
$l_\text{reg}(\bm{u})$ &
    $\Vert \bm{u} \Vert_{\bm{W}_u}^2$ &
    Control regularization term. \\
$l_\text{jnt}(\bm{x})$ &
    $(w_{\mathrm{jnt}}/2)(\Vert \min(\bm{x}^a-\underline{\bm{x}^a}, \bm{0}) \Vert_2^2 + \Vert \max(\bm{x}^a-\overline{\bm{x}^a}, \bm{0}) \Vert_2^2)$ &
    Avoids the hand joint limits. \\
$l_\text{sc}(\bm{x})$ &
    $w_{\mathrm{sc}}\sum \sigma \left(d\left(\mathrm{FK}_i(\bm{x}), \mathrm{FK}_j(\bm{x})\right)\right)$ &
    Avoids the hand self-collisions. \\
\bottomrule
\end{tabular}
\end{threeparttable}
\end{table*}

\subsection{Cost Terms}
\label{sec: cost_terms}

We construct a set of smooth and computationally efficient cost terms that {facilitate online generation of human-like finger-gaiting patterns}, as summarized in Tab.~\ref{tab: ddp_cost_terms_summary}.

\subsubsection{Regulation Cost.}
\label{subsec: regulation_cost}

{The regulation cost includes two parts, the state-dependent part $l_\text{reg}(\bm{x})$ and the control-dependent part $l_\text{reg}(\bm{u})$, which collectively infer the desired finger motion patterns that produce the expected object motion.}
The regulation cost is formulated as the weighted quadratic norm:
\begin{equation}
    \begin{aligned}
        l_\text{reg}(\bm{x}) &= \Vert \bm{x}^a-\bm{x}_\text{ref}^a \Vert_{\bm{W}_{\bm{x}}^a}^2 + \Vert \bm{x}^u\ominus\bm{x}_\text{ref}^u \Vert_{\bm{W}_{\bm{x}}^u}^2 \\
        l_\text{reg}(\bm{u}) &= \Vert \bm{u} \Vert_{\bm{W}_u}^2 \\
        l_f(\bm{x}_N) &= \Vert \bm{x}_N^a-\bm{x}_{N,\text{ref}}^a \Vert_{\bm{W}_{\bm{x}_N}^a}^2 + \Vert \bm{x}_N^u\ominus\bm{x}_{N,\text{ref}}^u \Vert_{\bm{W}_{\bm{x}_N}^u}^2
    \end{aligned}
    \label{eq: regulation_cost}
\end{equation}
where $\bm{x}_\text{ref}, \bm{x}_{N, \text{ref}}$ are the state references.
Note that the notations such as $\bm{x}^a, \bm{x}^u$ have been defined in {Tab.}~\ref{tab: notations}.

Merely using the object tracking cost $\Vert \bm{x}^u\ominus\bm{x}_\text{ref}^u \Vert_{\bm{W}_{\bm{x}}^u}^2$ often causes the system to fall into local optima; i.e., the fingers simply follow the object motion and fail to actively break contacts. Typically, the expected finger motion follows the periodic pattern (e.g., finger gaiting). Thus, the finger regularization cost $\Vert \bm{x}^a-\bm{x}_\text{ref}^a \Vert_{\bm{W}_{\bm{x}}^a}^2$ is introduced to encourage local exploration and contact breaking of the fingers. {We emphasize that $\bm{x}_\text{ref}^a$ is not derived from a predefined trajectory, but represents a nominal hand configuration (i.e., usually the initial grasp configuration) that remains fixed throughout the manipulation.}
In practice, $\bm{x}_\text{ref}^a$ can be generated using grasp synthesis methods (e.g., \citet{chen2024bodex}) or manually set using a graphical user interface. {The basic idea behind $\bm{x}_\text{ref}^a$ is inspired by the natural way humans manipulate objects, forming a pre-grasp to establish potential contacts for subsequent motions.}

{The weighting matrices \(\bm{W}_{\bm{x}}^a\), \(\bm{W}_{\bm{x}}^u\), and \(\bm{W}_{\bm{u}}\) are diagonal. Their magnitudes are initialized according to the typical numerical ranges of the corresponding tracking and regularization errors, so that no term dominates the optimization merely because of its scale. The state-related weights \(\bm{W}_{\bm{x}}^a\) and \(\bm{W}_{\bm{x}}^u\) are then coarsely rescaled across tasks according to the relevant translational and rotational error ranges. The control regularization weight \(\bm{W}_{\bm{u}}\) is kept fixed across tasks, while the terminal-state weights \(\bm{W}_{\bm{x}_N}^a\) and \(\bm{W}_{\bm{x}_N}^u\) are set larger than their corresponding running-state weights to facilitate convergence. The joint-limit and self-collision penalty weights \(w_{\mathrm{jnt}}\) and \(w_{\mathrm{sc}}\), introduced in Secs.~\ref{subsec: joint_limits_cost} and \ref{subsec: self_collision_cost}, are increased during tuning only when the corresponding violations are observed.}

Gradients on \(\bm{x}^a\) arise directly from finger regularization and indirectly from object tracking through the differentiable contact dynamics, enabling the desired object motion.

\subsubsection{Joint Limits Cost.}
\label{subsec: joint_limits_cost}

Dexterous hands typically have mechanical joint limits. Violation of these joint limits results in the collision of mechanical structures and overloading.
We thus design the joint limits cost as
\begin{equation}
    \begin{split}
        l_\text{jnt}(\bm{x}) &= \frac{w_\text{jnt}}{2} \left( \left\Vert \min(\bm{x}^a-\underline{\bm{x}^a}, \bm{0}) \right\Vert_2^2 \right. \\
        &\quad \left. + \left\Vert \max(\bm{x}^a-\overline{\bm{x}^a}, \bm{0}) \right\Vert_2^2 \right)
    \end{split}
    \label{eq: joint_limits_cost}
\end{equation}
where $w_\text{jnt}$ is the weighting parameter, and $\min, \max$ are element-wise operations. From (\ref{eq: joint_limits_cost}), the joint limits cost is designed as a barrier function. In other words, $l_\text{jnt}$ remains zero if $\bm{x}^a$ stays within the limits and penalizes the dimensions that violate limits.

\subsubsection{Self-Collision Cost.}
\label{subsec: self_collision_cost}

{Finally, a self-collision cost is introduced to avoid collisions between fingers}
\begin{equation}
    l_\text{sc}(\bm{x})= w_\text{sc} \sum_{\substack{i,j \in \{1,\cdots,n_l\} \\ i \neq j}} \sigma \left(d\left(\mathrm{FK}_i(\bm{x}), \mathrm{FK}_j(\bm{x})\right)\right)
    \label{eq: self_collision_cost}
\end{equation}
where $w_\text{sc}$ is the weighting parameter, and $n_l$ is the number of monitored links. To reduce the computational burden, rather than including all links in (\ref{eq: self_collision_cost}), we can first solve (\ref{eq: high_level_OCP}) without $l_\text{sc}$ and only monitor the links that are about to collide. Moreover, $\mathrm{FK}_i$ represents the forward kinematics of the $i^{th}$ link that the collision geometry is attached to, and $d(\cdot, \cdot)$ denotes the translational distance. The activation function $\sigma$ is written in quadratic barrier form as
\begin{equation}
    \sigma(a)=\left\{
        \begin{aligned}
            \frac{1}{2}(a - \gamma)^2 &, \quad a < \gamma \\
            0 &, \quad \text{otherwise}
        \end{aligned}
    \right.
    \label{eq: quadratic_barrier}
\end{equation}
According to (\ref{eq: quadratic_barrier}), the state $\bm{x}$ is penalized only if the distance between any two links falls below the threshold $\gamma$.

\begin{figure}
    \centering
    \includegraphics[width=0.7\linewidth]{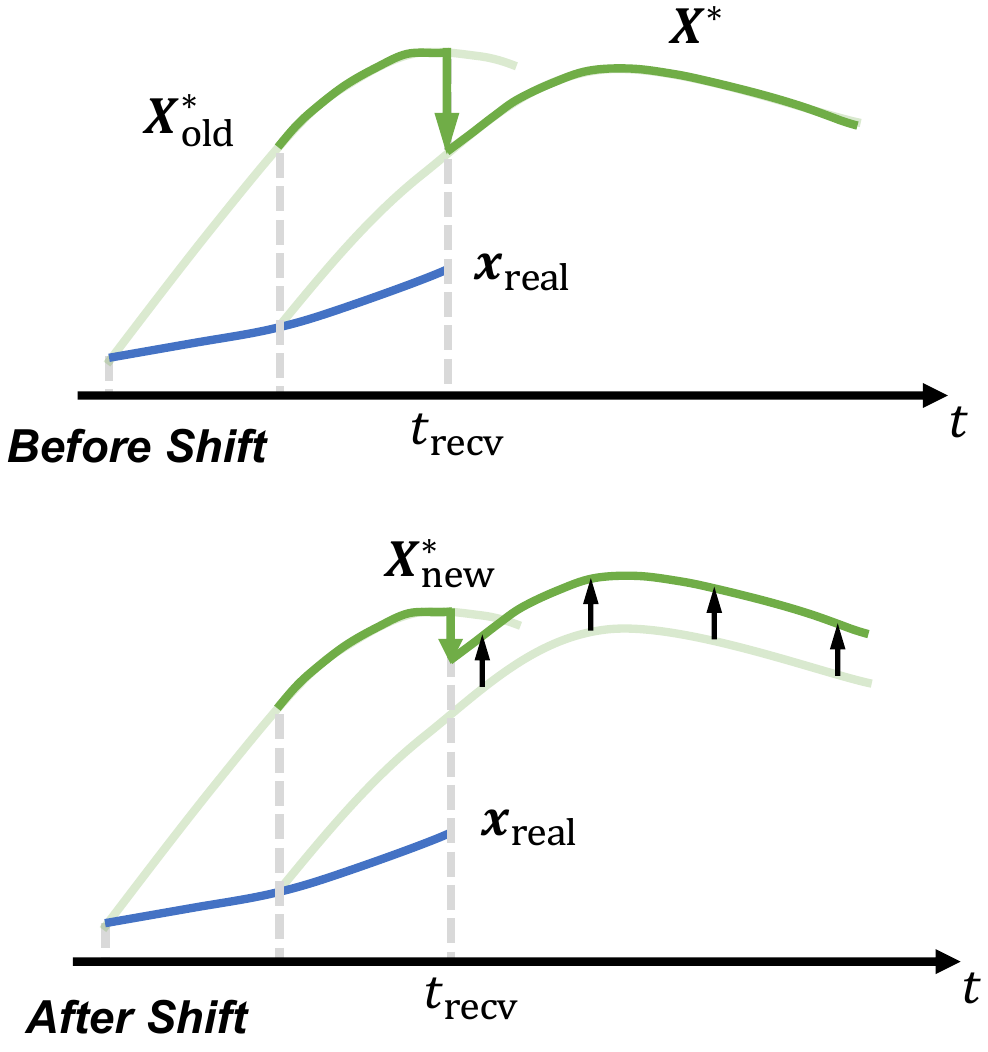}
    \caption{Illustration of trajectory shift for action smoothness. The high-level MPC computes optimal state sequence $\bm{X}^*$ with delays, shown by light green curves. The low-level module interpolates the previous solution $\bm{X}_\text{old}^{*}$ until a new one is received at $t_\text{recv}$. The interpolated trajectory is shown in dark green, and the arrow indicates discontinuity. To reduce the discontinuity and ensure smoother hardware commands, we shift the new solution to $\bm{X}_\text{new}^{*}$.
    Note that due to multiple delays, the real system states $\bm{x}_\text{real}$ always lag the commanded states.}
    \label{fig: trajectory_shift}
\end{figure}

\subsection{Model Predictive Control}

The proposed OCP (\ref{eq: high_level_OCP}) with soft constraints (\ref{eq: high_level_cost_term_definition}) is solved using the control-limited DDP \citep{tassa2014control} with our customized dynamics model and optimization cost. Although the implicit CQDC model encodes all possible contact modes, the solution of (\ref{eq: high_level_OCP}) yields a unique plan of desired motions and contacts within the prediction horizon, serving as the motion-contact reference that guides low-level tracking.
The optimal state and control trajectories, denoted as $\bm{X}^{*}, \bm{U}^{*}$, represent the desired motion.
The desired contacts include the contact locations and normals $\hat{\bm{n}}$—obtained via the embedded collision detection procedure—and the contact forces $\bm{\lambda}^{*}$, given by the dual solution of the conic constraint in (\ref{eq: CQDC_model_SOCP_form}). We denote $\bm{\Lambda}^{*}=\left\{\bm{\lambda}_0^{*},\cdots,\bm{\lambda}_{N-1}^{*}\right\}$.

The high-level MPC, built upon an implicit contact model, carefully designed cost terms, and a customized solver, provides a unified framework for {online inference of task-required finger-gaiting patterns} through long-term prediction of contact events. Several strategies are proposed to improve the convergence and action smoothness under imperfect low-level tracking and solution delays, thereby greatly enhancing inter-level coordination.

\subsubsection{Warm Start for Improved Convergence.}

Due to its local nature, optimization-based control generally requires a good initial guess to converge to a satisfactory solution.
Therefore, we propose warm-starting the OCP (\ref{eq: high_level_OCP}) to accelerate convergence. Specifically, the initial guess $\bm{U}_0$ is obtained by interpolating the previous optimal control sequence $\bm{U}^{*}$, leveraging the predictive capabilities of the method. Due to imperfect tracking, the hardware state typically lags behind the prediction; therefore, we stretch the timeline of $\bm{U}^{*}$ by a factor $\beta_1 \geq 1$. A larger $\beta_1$ slows the evolution of the warm start, resulting in smoother and slower finger motions.
Additionally, the control initial guess $\bm{U}_0$ concludes with zero-padding at the last time step; this zero-padding is crucial as the solution may diverge without it. The state initial guess $\bm{X}_0$ is then derived by rolling out $\bm{U}_0$.
This process guarantees that the initial guess precisely starts from the current system state and follows the trajectory of the previous solution. By easing the reliance on per-step convergence, the warm-start design allows the solution to stabilize gradually across iterations. In our experiments, we observe that this leads to more stable finger-gaiting patterns, which can adjust in response to external disturbances.

\subsubsection{{Trajectory Shift for Action Smoothness.}}

{High-level solution delays naturally introduce discontinuities between the current command and the newly received solution trajectory (Fig.~\ref{fig: trajectory_shift}), which} results in jerky motions in hardware.
{To ensure action smoothness, which is critical for preventing contact impacts that degrade tracking performance, we introduce a \textit{trajectory shift} to mitigate such effects.}
The \textit{trajectory shift} is described as $\bm{X}^{*}_\text{new}=\bm{X}^{*}+\beta_2\left(\bm{X}^{*}_\text{old}(t_\text{recv})-\bm{X}^{*}(t_\text{recv})\right)$, where $\bm{X}_\text{new}^{*}$ is the shifted solution trajectory, 
and $\bm{X}^{*}_\text{old}(t_\text{recv})-\bm{X}^{*}(t_\text{recv})$ measures the discontinuity at time $t_\text{recv}$.
Note that $\beta_2$ balances the smoothness and timeliness of the solution.

\section{Tactile-Reactive Motion-Contact Tracking}
\label{sec: low_level}
{When relying solely on high-level planning, the method can perform reasonably well provided that the modeling is accurate enough. However, its performance becomes sensitive to typical real-world factors such as non-physical effects from model relaxations, object-shape uncertainties, and perception noise. This motivates incorporating tactile feedback, which complements motion tracking by ensuring that the desired contacts are correctly established.}
Thus, this section proposes a tactile-feedback controller for synergistically tracking the motion-contact references.

\subsection{Force-Motion Model: Single Contact Case}

\begin{figure}[!tb]
    \centering
    \includegraphics[width=0.95\linewidth]{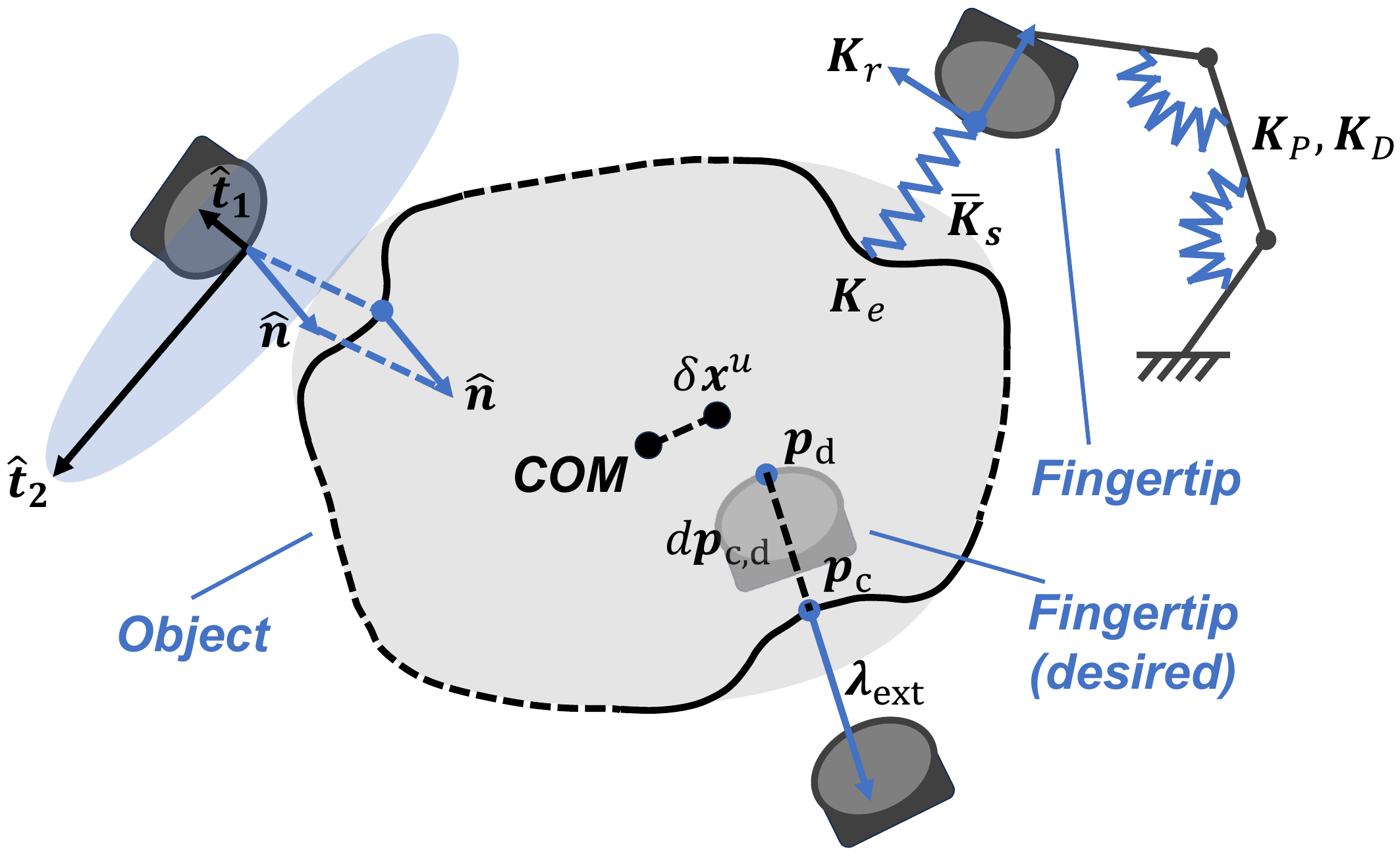}
    \caption{Notations of the force-motion model. The finger uses joint proportional-derivative (PD) control with stiffness $\bm{K}_P$ and damping $\bm{K}_D$. Due to controller compliance, the desired fingertip location differs from the actual location. At the contact point, the environment stiffness $\bm{K}_e$ and the finger's Cartesian stiffness $\bm{K}_r$ result in the equivalent stiffness $\bar{\bm{K}}_s$. The single contact stiffness relates the desired motions $d\bm{p}_{c,d}$ with the contact force $\bm{\lambda}_\text{ext}$. The forces and motions of all contacts are correlated due to the resultant displacement $\delta\bm{x}^u$ of the object's center of mass (COM). The force controlled direction $\hat{\bm{n}}$ and motion controlled directions $\{\hat{\bm{t}}_1, \hat{\bm{t}}_2\}$ are determined using the contact normal. This figure has been adapted from the conference version of this work \citep{jiang2024contact}.}
    \label{fig: coupling_complex_case}
\end{figure}

To derive the tactile-feedback controller, we require a force-motion model, which relates contact forces and finger motions. The following assumptions are made: 1) The finger's built-in controller operates under joint-space proportional-derivative (PD) control with stiffness $\bm{K}_P$ and damping $\bm{K}_D$. 2) There is mechanical compliance in both the object and fingers, characterized by environment stiffness $\bm{K}_e$. 3) Inertia, Coriolis, and centripetal effects are neglected at the scale of the contact forces. 4) A point contact model is used, and contact moments are ignored.
Inspired by \citet{gold2022model}, we derive the relationship between contact force and finger motion:
\begin{equation}
    \bm{\lambda}_\text{ext}=\bar{\bm{K}}_s d\bm{p}_{c,d}=\left(\bm{I}+\bm{K}_e\bm{K}_r^{-1}\right)^{-1}\bm{K}_e d\bm{p}_{c,d}
    \label{eq: single_contact_force_motion_law}
\end{equation}
where $\bm{\lambda}_\text{ext} \in \mathbb{R}^3$ is the contact force, $d\bm{p}_{c,d}=\bm{p}_d-\bm{p}_c$, and $\bm{p}_c, \bm{p}_d \in \mathbb{R}^3$ are contact point positions on the object and robot finger, respectively. Note that $\bm{p}_d$ is determined with the \textit{desired} finger configurations. Because of controller compliance and contact force, the \textit{desired} and \textit{real} finger configurations are usually different. The advantage of introducing $\bm{p}_d$ is to obtain a bounded equivalent stiffness $\bar{\bm{K}}_s$, which has a finite value even if $\bm{K}_e \rightarrow \infty$ (i.e., rigid environment). Moreover, $\bm{K}_r$ is defined as the controller's Cartesian stiffness at the finger contact point, where $\bm{K}_r^{-1}=\bm{J}_s(\bm{q})\bm{K}_P^{-1}\bm{J}_s(\bm{q})^\top$ is configuration dependent. The kinematic Jacobian $\bm{J}_s$ has been defined in {Tab.}~\ref{tab: notations}. Refer to Fig.~\ref{fig: coupling_complex_case} for a graphical illustration of all notations. The above model only applies to the single-contact case. In addition, the following equation is derived from joint-space PD control:
\begin{equation}
    \dot{\bm{q}}=\dot{\bm{q}}_d+\bm{K}_D^{-1}\left(\bm{K}_P(\bm{q}_d-\bm{q})-\bm{J}_s^\top\bm{\lambda}_\text{ext}\right)
    \label{eq: single_contact_joint_angle_law}
\end{equation}

\subsection{Force-Motion Model: Full Hand Case}

At some point, there may be multiple contacts between the dexterous hand and the object. To be consistent with the high-level module, we assume there are $n_c$ contacts. Note that the number and locations of contacts change continuously. The force-motion model of the complete system can be derived by repeating (\ref{eq: single_contact_force_motion_law}) for all $n_c$ contacts:
\begin{equation}
    \bm{\Lambda}_\text{ext}=\bar{\bm{K}}\left(\bm{P}_d-\bm{P}_c\right),
    \label{eq: full_hand_force_motion_law}
\end{equation}
The joint motion equation (\ref{eq: single_contact_joint_angle_law}) is now
\begin{equation}
    \dot{\bm{q}}=\dot{\bm{q}}_d+\bm{K}_D^{-1}\left(\bm{K}_P(\bm{q}_d-\bm{q})-\bm{J}^\top\bm{\Lambda}_\text{ext}\right)
    \label{eq: full_hand joint_angle_law}
\end{equation}
where the matrices are defined as
\begin{equation}
    \begin{aligned}
        \bm{\Lambda}_\text{ext} &=
            \left[ \begin{matrix}
                \bm{\lambda}_{\text{ext},1} \\
                \vdots \\
                \bm{\lambda}_{\text{ext},n_c} \\
            \end{matrix} \right], 
        \bm{P}_\text{d}=
            \left[ \begin{matrix}
                \bm{p}_{d,1} \\
                \vdots \\
                \bm{p}_{d,n_c} \\
            \end{matrix} \right],
        \bm{P}_\text{c}=
            \left[ \begin{matrix}
                \bm{p}_{c,1} \\
                \vdots \\
                \bm{p}_{c,n_c} \\
            \end{matrix} \right], \\
        \bar{\bm{K}} &=
            \mathrm{blkdiag}\left(\bar{\bm{K}}_{s,1},\cdots,\bar{\bm{K}}_{s,n_c}\right),
        \bm{J} = \left[ \begin{matrix}
                \bm{J}_{s,1} \\
                \vdots \\
                \bm{J}_{s,n_c} \\
            \end{matrix} \right]
    \end{aligned}.
    \label{eq: full_hand_stack_matrices}
\end{equation}
The dimensions of these vectors and matrices are $\bm{\Lambda}_\text{ext} \in \mathbb{R}^{3n_c \times 1},  \bm{P}_d, \bm{P}_c \in \mathbb{R}^{3n_c \times 1}, \bar{\bm{K}} \in \mathbb{R}^{3n_c \times 3n_c}, \bm{J} \in \mathbb{R}^{3n_c \times n_q}$.

\begin{figure}[!tb]
    \centering
    \includegraphics[width=1\linewidth]{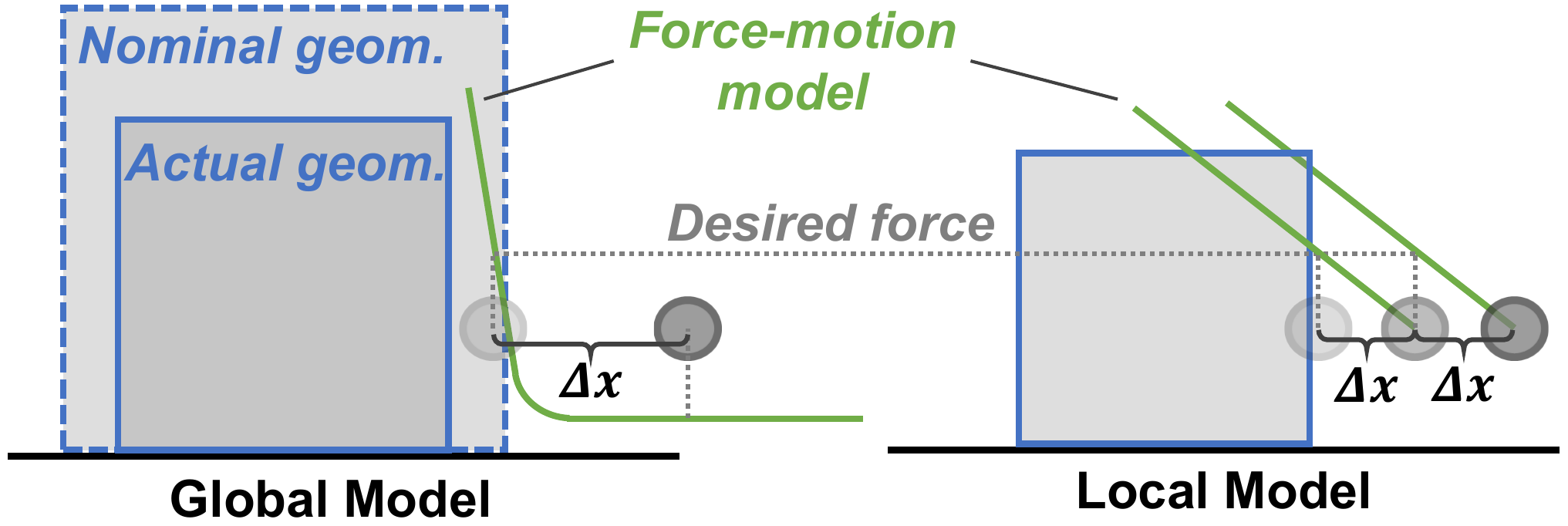}
    \caption{{Comparison of \textit{global} (e.g., the CQDC model) and \textit{local} force–motion models. The global model, assuming nominal object geometry (geom.), defines forces based on absolute finger positions, whereas the local model conditions forces on relative finger displacements. Consequently, only the local model forms a tactile feedback pathway, iteratively updating finger positions ($\Delta\bm{x}$) until contact is achieved. In contrast, the global model computes fixed finger positions, which may not result in actual contact.}}
    \label{fig: difference_global_local_model}
\end{figure}

\subsection{Force-Motion Model: with the Coupling Effect}

In the above analysis, we assume the object is fixed with contact points $\bm{p}_c$ unchanged, at least within the prediction horizon. However, supposing that the object is not fully constrained, the motion of each finger also affects the contact forces of the other fingers through the object's movement,
as shown in Fig.~\ref{fig: coupling_complex_case}.
In other words, the force-motion model is no longer in decoupled form (\ref{eq: full_hand_force_motion_law}). To analyze the coupling effect, we introduce the grasp matrix $\bm{G}_o \in \mathbb{R}^{6 \times 3n_c}$. The following mappings hold:
\begin{equation}
    \bm{w}_\text{ext}=\bm{G}_o\bm{\Lambda}_\text{ext}, \quad \delta\bm{P}_c=\bm{G}_o^\top\delta\bm{x}^u
    \label{eq: grasp_matrix_mappings}
\end{equation}
where $\bm{w}_\text{ext}$ is the resultant wrench applied on the object. Here, $\delta\bm{x}^u$ is the local object pose increment (six-dimensional for free motion). We ignore the local deformation and assume that the contact points move with the object. Under quasi-dynamic assumptions and with a constant external force such as gravity, the resultant wrench should remain constant, denoted by $\bm{w}_\text{ext} = \text{Const}$. From (\ref{eq: full_hand_force_motion_law}) and (\ref{eq: grasp_matrix_mappings}), it is then derived that
\begin{equation}
    \delta\bm{w}_\text{ext}=\bm{G}_o\delta\bm{\Lambda}_\text{ext}=\bm{G}_o\bar{\bm{K}}\left(\delta\bm{P}_d-\delta\bm{P}_c\right)=0
    \label{eq: force_motion_coupling}
\end{equation}
We apply the implicit function theorem to (\ref{eq: force_motion_coupling}) and obtain
\begin{equation}
    \begin{aligned}
        \frac{\partial\bm{x}^u}{\partial\bm{P}_d} &= -\left[\frac{\partial(\delta\bm{w}_\text{ext})}{\partial\bm{x}^u}\right]^{-1}\frac{\partial(\delta\bm{w}_\text{ext})}{\partial\bm{P}_d} \\
        &= \left(\bm{G}_o\bar{\bm{K}}\bm{G}_o^\top\right)^{-1}\bm{G}_o\bar{\bm{K}}
    \end{aligned}
    \label{eq: partial_xu_partial_pd}
\end{equation}
We then take the derivative of (\ref{eq: full_hand_force_motion_law}) with respect to $\bm{P}_d$ and derive the coupled stiffness $\bm{K}_\text{coup} \in \mathbb{R}^{3n_c \times 3n_c}$ as
\begin{equation}
    \begin{aligned}
        \bm{K}_\text{coup} = \frac{\partial\bm{\Lambda}_\text{ext}}{\partial\bm{P}_d} &= \bar{\bm{K}}\left(\bm{I}-\frac{\partial\bm{P}_c}{\partial\bm{x}^u}\frac{\partial\bm{x}^u}{\partial\bm{P}_d} \right) \\
        &= \bar{\bm{K}}-\bar{\bm{K}}\bm{G}_o^\top\left(\bm{G}_o\bar{\bm{K}}\bm{G}_o^\top\right)^{-1}\bm{G}_o\bar{\bm{K}}
    \end{aligned}
    \label{eq: coupled_force_motion_law}
\end{equation}
Note that the second term is the correction term due to the coupling effect. The matrix $\bm{G}_o\bar{\bm{K}}\bm{G}_o^\top$ may be rank deficient depending on the number and configuration of active contacts. For partially constrained objects, we retain the rows of $\bm{G}_o$ associated with unconstrained motion coordinates and adjust the dimensions in (\ref{eq: grasp_matrix_mappings})--(\ref{eq: coupled_force_motion_law}) accordingly.
If the resulting matrix is singular or ill-conditioned, we use $(\bm{G}_o\bar{\bm{K}}\bm{G}_o^\top+\varepsilon\bm{I})^{-1}$, with a small $\varepsilon>0$, to obtain a regularized approximation of the coupling model.

Using the substitutions $\dot{\bm{P}}_d=\bm{J}(\bm{q}_d)\dot{\bm{q}_d}$ and $\bm{u}=\dot{\bm{q}}_d$, the complete force-motion model with coupling effect can be derived using (\ref{eq: full_hand joint_angle_law}) and (\ref{eq: coupled_force_motion_law})
\begin{equation}
    \begin{aligned}
        \left[\begin{matrix}
            \dot{\bm{q}} \\
            \dot{\bm{q}}_d \\
            \dot{\bm{\Lambda}}_\text{ext} \\
        \end{matrix} \right] = 
        \left[\begin{matrix}
            \bm{u}+\bm{K}_D^{-1}\left(\bm{K}_P(\bm{q}_d-\bm{q})-\bm{J}(\bm{q})^\top\bm{\Lambda}_\text{ext}\right) \\
            \bm{u} \\
            \bm{K}_\text{coup}\bm{J}(\bm{q}_d)\bm{u} \\
        \end{matrix} \right]
    \end{aligned}
    \label{eq: final_force_motion_model}
\end{equation}
Defining the state vector as $\bm{s} \triangleq \left[\bm{q}; \bm{q}_d; \bm{\Lambda}_\text{ext}\right] \in \mathbb{R}^{2n_q+3n_c}$, the system dynamics can be expressed in continuous-time form as $\dot{\bm{s}}=\bm{g}(\bm{s}, \bm{u})$. Note that the dynamics are inherently nonlinear, primarily due to the dependence on Jacobians.

{It is worth noting that (\ref{eq: final_force_motion_model}) represents a local force–motion model, directly relating contact forces to the desired finger motions. The absence of prescribed object geometry in this formulation enables the tactile feedback pathway. The distinctions between global (e.g., the CQDC model) and local force–motion models are summarized in Fig.~\ref{fig: difference_global_local_model}.}

\begin{table*}[!b]
\centering
\caption{Determination of weighting matrices in the low-level MPC.}
\label{tab: weighting_matrices}
\renewcommand{\arraystretch}{1.5}
\begin{threeparttable}[b]
\begin{tabular}{c|c|c|c} 
\toprule
\multirow{2}{*}{\textbf{Matrix}} & \multicolumn{2}{c|}{\textbf{HFMC}} & \multirow{2}{*}{\textbf{JSC}} \\
\cline{2-3}
 & $\Vert \bm{\lambda}_\text{ext}^\text{ref} \Vert \geq \underline{\bm{\lambda}}$ & $\Vert \bm{\lambda}_\text{ext}^\text{ref} \Vert < \underline{\bm{\lambda}}$ & \\
\hline
$\bm{W}_{\bm{q}} \in \mathbb{R}^{n_q \times n_q}$ & \multicolumn{2}{c|}{$\bm{0}$} & $k_{\bm{q}}\bm{I}$ \\
\hline
$\hat{\bm{W}}_{\bm{p}} \in \mathbb{R}^{6 \times 6}$ &
  $[\hat{\bm{T}}\hat{\bm{T}}^\top \ \ \bm{0};\ \ \bm{0} \ \ w_\text{ori}\bm{I}]$
 & 
  $[\bm{I} \ \ \bm{0};\ \ \bm{0} \ \ w_\text{ori}\bm{I}]$ & $\bm{0}$ \\
\hline
$\hat{\bm{W}}_{\bm{\Lambda}} \in \mathbb{R}^{3 \times 3}$ & $\hat{\bm{n}}\hat{\bm{n}}^\top$ &
    $\bm{0}$ & 
    $\hat{\bm{n}}\hat{\bm{n}}^\top$ \\
\hline
$\bm{W}_{\bm{u}} \in \mathbb{R}^{n_u \times n_u}$ & 
  \multicolumn{3}{c}{$k_{\bm{u}}\bm{I}$} \\
\bottomrule
\end{tabular}
\end{threeparttable}
\end{table*}

\subsection{Model Predictive Control}
\label{subsec: low_level_model_predictive_control}

{Given the complex nonlinear coupling between contact forces and finger motions, we employ a low-level MPC that leverages soft constraints and adaptive weighting to jointly coordinate contact and motion tracking. The MPC is formulated based on the deduced force–motion model (\ref{eq: final_force_motion_model})}
\begin{equation}
    \begin{aligned}
        \min_{\bm{u}}\! \int_{t_0}^{t_0+T} & \bigg( \Vert \bm{q}\!-\!\bm{q}_\text{ref} \Vert_{\bm{W_q}}^2 \!+\! 
        \sum_{i=1}^{n_c} \left( \Vert \mathrm{FK}_i\!(\bm{q}) \!\ominus\! \mathrm{FK}_i\!(\bm{q}_\text{ref}) \Vert_{\bm{W}_{\bm{p}_i}}^2 \right. \\
        & \left. + \Vert \bm{\lambda}_{\text{ext}, i} - \bm{\lambda}^\text{ref}_{\text{ext}, i} \Vert_{\bm{W}_{\bm{\Lambda}_i}}^2 \right) + \Vert \bm{u} \Vert_{\bm{W_{u}}}^2 \bigg)\,d\tau \\
        \mbox{s.t.} \quad & \dot{\bm{s}}=\bm{g}(\bm{s},\bm{u})
    \end{aligned}
    \label{eq: low_level_MPC}
\end{equation}
where the subscript $i$ denotes the $i^{th}$ contact, $\bm{q}_\text{ref}$ and $\bm{\lambda}^\text{ref}_\text{ext}$ are the high-level reference finger motions and contact forces, respectively. Besides, $\mathrm{FK}(\cdot)$ refers to the forward kinematics, and $\ominus$ computes the pose error. The time dependency in (\ref{eq: low_level_MPC}) is omitted for clarity.
In addition, $\bm{q}_\text{ref}$ and $\bm{\lambda}^\text{ref}_\text{ext}$ are obtained by fitting splines through the discrete solutions $\bm{X}^*$ and $\bm{\Lambda}^*$.
A schematic of the proposed controller is shown in Fig.~\ref{fig: framework_overview} (C).

Note that the MPC is a finite-horizon optimal control problem with continuous-time dynamics. At each iteration, integration is initialized with the measured joint positions $\bm{q}$, the last command $\bm{q}_d$, and the measured external contact forces $\bm{\Lambda}_\text{ext}$. 
{The measured contact points are used to construct the grasp matrix \(\bm G_o\) in (\ref{eq: grasp_matrix_mappings}) and thus the coupled stiffness matrix \(\bm K_{\rm coup}\) in (\ref{eq: coupled_force_motion_law}), thereby incorporating the measured contact geometry into the low-level prediction.}
Following \citet{gold2022model}, we solve the problem with efficient third-party solvers. 

The MPC design constitutes a unified and flexible framework, supporting diverse control modes and the integration of additional constraints.
For instance, with nonzero $\bm{W}_q$, the MPC behaves like Joint-Space Control (JSC), governing large-range finger motions such as initial grasping.
Furthermore, when the pose and contact force error terms are adaptively weighted by $\bm{W_p}$ and $\bm{W_\Lambda}$ (Sec.~\ref{subsec: determination_of_low_level_matrices}), the MPC functions as Hybrid Force-Motion Control (HFMC), coordinating motion and contact tracking with minimal mutual interference.

\subsection{Adaptive Weighting for Hybrid Force-Motion Control}
\label{subsec: determination_of_low_level_matrices}

The weighting matrices $\bm{W}_{\bm{\Lambda}}, \bm{W}_{\bm{p}}$ adaptively define the force–motion control subspaces based on the planned contact normal $\hat{\bm{n}}$ from the high-level module.
Using QR decomposition, we obtain the orthogonal basis $\left\{\hat{\bm{n}},\hat{\bm{t}}_{1},\hat{\bm{t}}_{2}\right\}$ and define $\hat{\bm{T}} \triangleq \left[\hat{\bm{t}}_{1}, \hat{\bm{t}}_{2} \right]$, which spans the contact tangent space (Fig.~\ref{fig: coupling_complex_case}).
The element matrices $\hat{\bm{W}}_{\bm{\Lambda}}$ and $\hat{\bm{W}}_{\bm{p}}$ are then constructed as:
\begin{equation}
    \hat{\bm{W}}_{\bm{\Lambda}}=\hat{\bm{n}}\hat{\bm{n}}^\top, \quad \hat{\bm{W}}_{\bm{p}}=
    \left[ \begin{matrix}
        \hat{\bm{T}}\hat{\bm{T}}^\top&\bm{0} \\
        \bm{0}&w_\text{ori}\bm{I} \\
        \end{matrix} \right]
    \label{eq: HFMC_matrices}
\end{equation}
The weighting matrices are defined as $\bm{W}_{\bm{\Lambda}} = k_{\bm{\Lambda}} \hat{\bm{W}}_{\bm{\Lambda}}$ and $\bm{W}_{\bm{p}} = k_{\bm{p}} \hat{\bm{W}}_{\bm{p}}$, and the environment stiffness as $\bm{K}_e = k_e \hat{\bm{W}}_{\bm{\Lambda}}$.
{The directional structures of \(\bm{W}_{\bm{\Lambda}}\) and \(\bm{W}_{\bm{p}}\) are thus determined online by the planned contact normal, while the scalar gains \(k_{\bm{\Lambda}}\) and \(k_{\bm{p}}\) are coarsely tuned once and kept fixed across tasks.}
For brevity, the subscript $i$ is omitted. The scalar $w_\text{ori}$ controls the orientation weight.
The stiffness $\bm{K}_e$ specifies the direction of steepest force growth, and the force-controlled direction (the principal eigenvector of $\bm{W}_{\bm{\Lambda}}$) aligns with the contact normal. In other words, only the \textit{normal contact force} is regulated to maintain the planned contact, while \textit{tangential motions} are used to generate object motion and therefore remain in the motion-controlled subspace.

It should be noted that the force-motion relationship, as described in (\ref{eq: single_contact_force_motion_law}), is only effective for active contacts. Thus, all contacts are classified into two categories according to the force threshold $\underline{\bm{\lambda}}$. The matrices $\hat{\bm{W}}_{\bm{\Lambda}}, \hat{\bm{W}}_{\bm{p}}$ in (\ref{eq: HFMC_matrices}) are designed for contacts with $\Vert \bm{\lambda}_\text{ext}^\text{ref} \Vert \geq \underline{\bm{\lambda}}$, whereas for contacts with $\Vert \bm{\lambda}_\text{ext}^\text{ref} \Vert < \underline{\bm{\lambda}}$, the matrices are defined as $\hat{\bm{W}}_{\bm{\Lambda}}=\bm{0}, \hat{\bm{W}}_{\bm{p}}=[\bm{I} \ \ \bm{0};\ \ \bm{0} \ \ w_\text{ori}\bm{I}]$
to perform position tracking only. 
During contact breaking, the reference contact force from the high-level planner drops below the threshold, so the controller switches to position tracking for fingertip lift-off and relocation. Since the switch is governed by the reference contact force rather than noisy tactile measurements, the transition follows the planned contact schedule and avoids noise-induced chattering, while the input regularization term in (24) further smooths the commanded motion.
In addition, we set $\bm{W}_{\bm{q}}=\bm{0}$ for HFMC. As for JSC, we set $\bm{W}_{\bm{q}}=k_{\bm{q}}\bm{I}$ and $\bm{W}_{\bm{p}}=\bm{0}$. The determination of weighting matrices is summarized in {Tab.}~\ref{tab: weighting_matrices}.

Once (\ref{eq: low_level_MPC}) is solved and the optimal control $\bm{u}^{*}$ is obtained, the hardware joint commands are computed as $\bm{q}_d(t_0)+\bm{u}^{*}(t_0) \Delta t$, where $\Delta t$ is the sampling time used for solving (\ref{eq: low_level_MPC}). The next MPC iteration is warm-started by shifting $\bm{u}^{*}$. {The low-level module refines the planned finger motions through tactile-informed contact tracking. This feedback mechanism complements the high-level planner by ensuring that intended contacts are realized, thereby strengthening overall performance—particularly in the presence of modeling errors.}

\begin{figure*}[!t]
    \centering
    \includegraphics[width=0.87\linewidth]{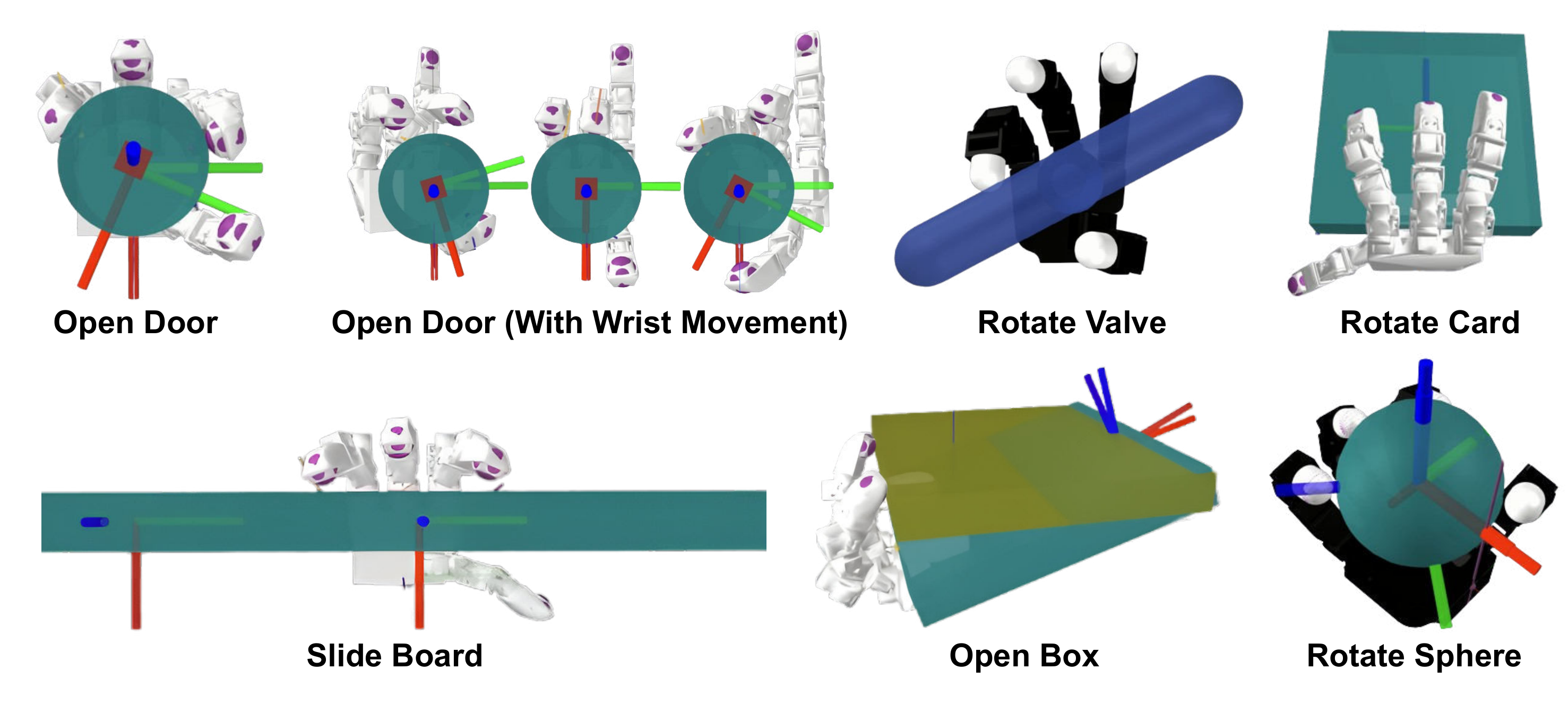}
    \caption{Snapshots of the environment-constrained manipulation tasks used for evaluation in simulation and on hardware.
    {We evaluate our method on two dexterous hands—the Allegro Hand and the LEAP Hand—across a variety of tasks to demonstrate the task-level adaptability of the proposed model-based framework to different manipulation setups.}
    }
    \label{fig: task_snapshots}
\end{figure*}

\section{Simulation Results}
\label{sec: simulation_results}
\subsection{Simulation Setup}

We use MuJoCo \citep{todorov2012mujoco} for simulation and adopt ROS to parallelize the high-level and low-level modules. The high-level module generates the unified motion-contact references at 10 Hz, whereas the low-level module performs tactile-informed motion correction at 30 Hz. Appendix~B further details the performance of the upper-level unified motion-contact planning module and the selection of key parameters in simulation.

\subsection{In-Hand Manipulation Tasks}

We consider two complementary evaluation settings: \textbf{environment-constrained manipulation}, where external structures restrict object motion, as shown in Fig.~\ref{fig: task_snapshots}; and \textbf{prehensile in-hand manipulation}, where the object is maintained within the hand through coordinated finger contacts and deliberately reoriented relative to the palm and fingers (Sec.~\ref{subsec: prehensile_tasks}).
The \textbf{Rotate Sphere} task in Fig.~\ref{fig: task_snapshots}, performed with the Allegro Hand, is used as the simulation benchmark for comparative analysis in Sec.~\ref{subsec: highlevel_comparison_baselines}. The environment-constrained tasks are described below. Please refer to the \href{https://drive.google.com/file/d/1ZMQ43PDZnjn1DYf48m6V3kgo47HlhPUA/view?usp=sharing}{\textcolor{cyan!70!black}{attached video}} for visual demonstrations.

\subsubsection{\textbf{Open Door}.}
The door handle rotates around its z-axis (i.e., the blue axis), which is perpendicular to the palm. The system has $n_x = 17$ DoFs, representing the combined DoFs of the hand and the object.
\subsubsection{\textbf{Open Door (With Wrist Movement)}.}
This task is similar to Open Door except that 
the wrist is allowed to translate freely and rotate around the vertical direction within -30$^\circ$$\sim$30$^\circ$.
The hand executes a three-fingered grasp, utilizing the thumb and two additional fingers.
The system has $n_x=17$ DoFs.
\subsubsection{\textbf{Rotate Valve}.}
The capsule-shaped valve rotates around its z-axis which is perpendicular to the palm. The system has $n_x=17$ DoFs.
\subsubsection{\textbf{Rotate Card}.}
The card moves freely on the table. The dexterous hand rotates the card around the vertical direction while minimizing the card's translation. The thumb is not used in this task. The system has $n_x = 19$ DoFs.
\subsubsection{\textbf{Slide Board}.}
The board moves freely on the table. The dexterous hand slides the board along its long edge toward a target location. Three fingers are used similar to the Rotate Card task. Translation and rotation in the other two orthogonal directions should be minimized. The system has $n_x = 19$ DoFs.
\subsubsection{\textbf{Open Box}.}
The box is fixed on the table, and the lid rotates around the hinge joint. The dexterous hand should make contact with the lid using the same three fingers as in the Rotate Card task, and open it until it reaches a specified angle. The system has $n_x = 17$ DoFs.
\subsubsection{\textbf{Rotate Sphere}.}
The sphere rotates freely in the SO(3) space with its center fixed. The dexterous hand rotates the sphere toward a target SO(3) orientation. The system has $n_x=19$ DoFs.

\begin{table*}[b]
\centering
\arrayrulecolor{black}
{
\caption{{Performance comparison of different methods in MuJoCo simulation of the \textbf{Rotate Sphere} task.}}
\label{tab: sim_comparison_sphere_rot_task}
\begin{threeparttable}[b]
\resizebox{\textwidth}{!}{
    \begin{tabular}{c|cccc}
    \toprule
    Method\tnote{a} & Success rate $\uparrow$ & \begin{tabular}[c]{@{}c@{}}Average task error ($\mathrm{rad}$) $\downarrow$ \end{tabular} & \begin{tabular}[c]{@{}c@{}}Average task error\\ of successful trials\\ ($\mathrm{rad}$) $\downarrow$\end{tabular} & \begin{tabular}[c]{@{}c@{}}Average task error S.D. \\ of successful trials\\($\times 10^{-1}\mathrm{rad}$) $\downarrow$\end{tabular} \\
    \hline
    Ours (CQDC)\tnote{b} & 100 / 100 & 4.954$\times 10^{-5}$ & 4.954$\times 10^{-5}$ & 0.003 \\
    \textbf{Ours (2\textsuperscript{nd})} & \textbf{100 / 100} & \textbf{0.024} & 0.024 & 0.034 \\
    Ours (2\textsuperscript{nd}, openloop) & 14 / 100 & 0.644 & 0.069 & 0.004 \\
    \hline
    Pang et al. (CQDC)\tnote{b} & 93 / 100 & 0.029 & \textbf{0.020} & \naline \tnote{c} \\
    Jin & \textbf{100 / 100} & 0.049 & 0.049 & \textbf{0.000} \\
    MJPC & 83 / 100 & 0.099 & 0.050 & 4.505 \\
    iLQG & 14 / 100 & 0.464 & 0.084 & 5.601 \\
    \midrule
    Method\tnote{a} & \begin{tabular}[c]{@{}c@{}}Average task time\\ of successful trials ($\mathrm{s}$) $\downarrow$\end{tabular} & \begin{tabular}[c]{@{}c@{}} Average joint \\ acceleration ($\mathrm{rad}/\mathrm{s}^2$) $\downarrow$\end{tabular} & \begin{tabular}[c]{@{}c@{}} Average frequency \\ high-level/low-level\\ ($\mathrm{Hz}$) $\uparrow$ \end{tabular} &  \\
    \hline
    Ours (CQDC)\tnote{b} & 28.824 & 0.247 & 10.007 / \naline & \\
    \textbf{Ours (2\textsuperscript{nd})} & 35.550 & \textbf{0.618} & 9.884 / 30.006 & \\
    Ours (2\textsuperscript{nd}, openloop) & 47.801 & 0.651 & 9.456 / \naline & \\
    \hline
    Pang et al. (CQDC)\tnote{b} & 39.678 & 0.666 & \naline\, / \naline \tnote{c} & \\
    Jin & \textbf{7.473} & 4.549 & 34.904 / \naline & \\
    MJPC & 21.926 & 174.578 & 99.996 / \naline & \\
    iLQG & 24.999 & 198.478 & \textbf{99.998} / \naline & \\
    \bottomrule
    \end{tabular}
}
\begin{tablenotes}
     \item[a] \begin{minipage}[t]{0.95\textwidth} The first three methods are implemented in our codebase, while the other four use publicly available implementations. \end{minipage}
     \item[b] \begin{minipage}[t]{0.95\textwidth} Ours (CQDC) refers to employing the CQDC model (\ref{eq: CQDC_model_barrier}) both for planning and simulation, which serves as a performance upper bound. \end{minipage}
     \item[c] \begin{minipage}[t]{0.95\textwidth} We omit the task error S.D. and frequency metrics for Pang et al. (CQDC), as it reflects only the planning results. \end{minipage}
\end{tablenotes}
\end{threeparttable}
}
\end{table*}

\subsection{{Comparative Analysis with Other In-Hand Manipulation Approaches}}
\label{subsec: highlevel_comparison_baselines}

{To fairly position the proposed framework within the literature, we compare it against several state-of-the-art model-based methods for in-hand manipulation: 1) \textit{Global Contact-Rich Planning (Pang et al.)} \citep{pang2023global}: An RRT-based motion planning method that uses a CQDC-based local reachability metric for node extension. The resulting trajectory is further refined via trajectory optimization to mitigate non-physical artifacts introduced during smoothing. 2) \textit{Complementarity-Free Optimization (Jin)} \citep{jin2024complementarity}: An MPC-based method that employs a complementarity-free contact model, which relaxes the cone constraints of the CQDC model. 3) \textit{Predictive Sampling (MJPC)} \citep{howell2022predictive}: A sampling-based MPC approach that has been demonstrated to be effective. 4) \textit{Nonsmooth Gradient-Based MPC (iLQG)} \citep{howell2022predictive}: A gradient-based method that exploits the non-smooth contact gradients provided by MuJoCo. For fairness, all baselines above are implemented using their publicly available code. In addition, we also include ablations of our framework for comparison.}

{We use the \textbf{Rotate Sphere} task as a benchmark for comparison. For fairness, all methods are evaluated in the same MuJoCo environment, except Ours (CQDC) and Pang et al. (CQDC), which report planning results only.} We generate 100 random target orientations and evaluate each method by recording the system states within a 60-second time budget per target.

\subsubsection{Evaluation Metrics.}
The following metrics are used for comparison:
\begin{itemize}
    \item [-] \textbf{Success Rate}: The number of successful trials over all 100 trials. The trial is judged a success if the minimum orientation error is below 8 degrees.
    \item [-] \textbf{Average Task Error}: The average minimum orientation error among all 100 trials or the successful trials.
    \item [-] \textbf{Average Task Error S.D.}: The standard deviation of orientation error from the first attainment of its minimum over the trial to the trial end, averaged over successful trials. This metric reflects stability near the achieved orientation.
    \item [-] \textbf{Average Task Time}: The average time taken before the trial succeeds. This metric is averaged over all successful trials.
    \item [-] \textbf{Average Joint Acceleration}: The average joint acceleration over all joints and all 100 trials.
    This metric reflects the smoothness of {finger motions}.
\end{itemize}

\subsubsection{Results and Discussion.}
{
As shown in {Tab.}~\ref{tab: sim_comparison_sphere_rot_task}, the proposed method \textbf{Ours (2\textsuperscript{nd})} achieves the highest precision and success rate while producing the smoothest finger motions. Among the ablation methods, \textbf{Ours (CQDC)} serves as an upper-performance bound, since it uses the same CQDC dynamics model for both planning and simulation; therefore, its metrics are provided only for reference rather than direct comparison. In contrast, \textbf{Ours (2\textsuperscript{nd})} and \textbf{Ours (2\textsuperscript{nd}, openloop)} use MuJoCo’s second-order dynamics for forward simulation, which more closely reflect real contact dynamics. However, the mismatch between the planning and simulation models induces a force-at-a-distance effect, impeding reliable contact establishment and degrading convergence as the simulated states drift away from the previously optimized ones.
Consequently, the performance of both methods remains below the upper bound.
Notably, \textbf{Ours (2\textsuperscript{nd})} narrows the performance gap with the upper bound compared to \textbf{Ours (2\textsuperscript{nd}, openloop)}, which disables the low-level contact tracking module. This demonstrates that the proposed method effectively addresses modeling gaps in contact-implicit methods by generating sufficient contact force and restraining sliding contacts through tactile-informed motion corrections.
}

{Regarding the baselines, \textbf{Pang et al. (CQDC)} fails in 7 trials, unable to find a path when the RRT search tree reached its maximum of 5,000 nodes. Note that \textbf{Pang et al. (CQDC)} reflects only the offline planning results, although its performance approaches that of our method; all trials fail under open-loop execution in MuJoCo due to modeling errors in the CQDC model and the lack of feedback.}

\begin{figure}[!tb]
    \centering
    \includegraphics[width=1\linewidth]{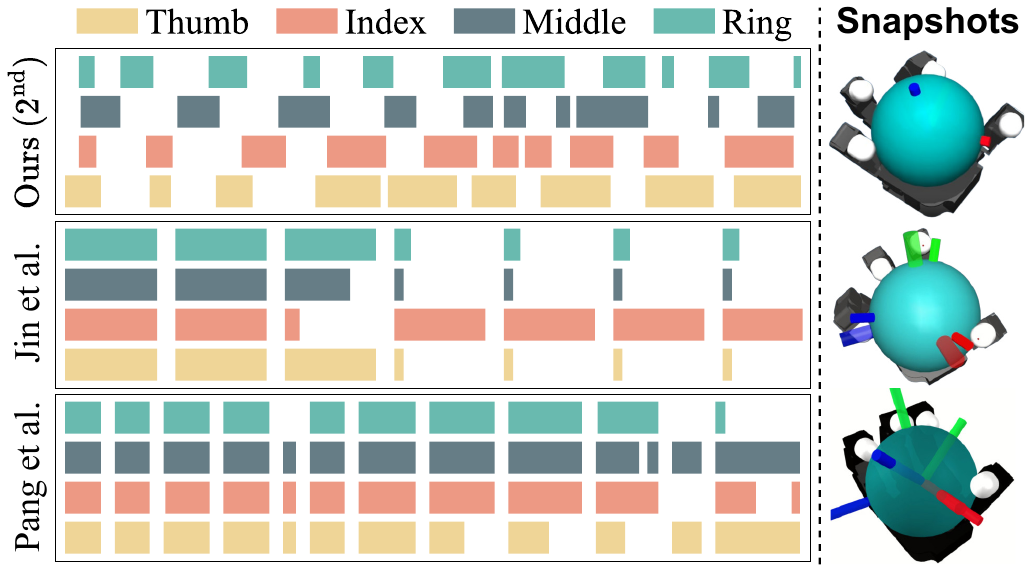}
    \caption{{Illustration of finger gaiting patterns across different methods. The colored patches denote time intervals during which the contact distance of each finger falls below 1 mm. Representative snapshots are shown on the right. For Jin (2025) and Pang et al., the fingers move almost simultaneously with periodic manual regrasping, as reflected by the gaps, whereas the proposed framework exhibits natural, human-like finger gaiting behavior. For a more intuitive visualization of this phenomenon, we refer interested readers to the \href{https://drive.google.com/file/d/1hMf0XQUl73HWsqHGEH-KdrDBSFd2GL1v/view?usp=sharing}{\textcolor{cyan!70!black}{attached video}}.}}
    \label{fig: finger_gaiting_comparison}
    \vspace{-0.3cm}
\end{figure}

{In addition, the proposed method achieves performance comparable to \citet{jin2024complementarity}, while attaining slightly higher precision. While Jin's relaxation of the CQDC model alleviates some force-at-a-distance effects, residual modeling errors persist due to necessary approximations for differentiability. This indicates that refining contact modeling alone is insufficient to address modeling errors, and that tactile incorporation is essential for accurate and robust manipulation. A major advantage of leveraging complementarity-free modeling is the higher control frequency, which also reduces the time to reach the target. It also attains the lowest standard deviation in task error, as the fingers remain largely stationary once the orientation target is reached. Additionally, \textbf{Jin (2025)} succeeds only when the fingers are manually reset to the initial configuration periodically. In contrast, our method automatically generates finger-gaiting behavior (Fig.~\ref{fig: finger_gaiting_comparison}), eliminating the need for explicit regrasping, likely enabled by the interplay between the contact model, the proposed cost formulation, and the customized solver. Detailed studies of this finger-gaiting behavior would be valuable and are left for future work.}

{Finally, compared with \textbf{MJPC}, our method achieves lower task error, as sampling-based approaches are less precise, especially near convergence.} Our method also exhibits significantly lower task error standard deviation and joint acceleration, suggesting potential advantages for hardware deployment. In contrast, {\textbf{iLQG}} performs the worst, since gradients computed via MuJoCo’s finite differences are highly non-smooth and often vanish in contact-rich manipulation, highlighting the importance of smoothing. Notably, {\textbf{iLQG}} performs better on locomotion tasks \citep{howell2022predictive, zhang2025whole}, where foot contacts are easily established and the gradients are typically non-zero.

\begin{figure}[!t]
    \centering
    \includegraphics[width=1\linewidth]{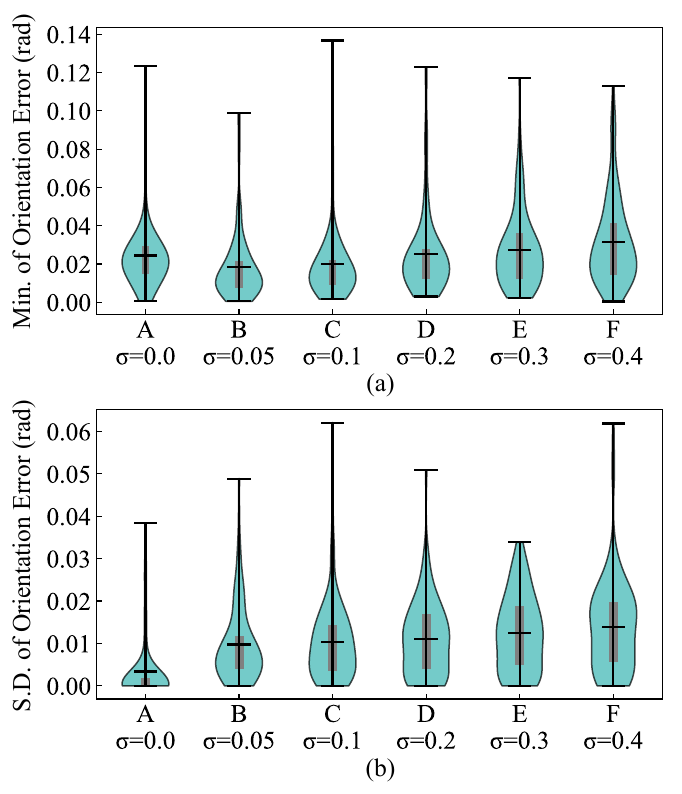}
    \caption{Task error distribution for the \textbf{Rotate Sphere} task with noisy object orientation. The x-axis labels indicate the standard deviation of the added Gaussian noise. The bars represent mean and extreme values, whereas the gray stripes show the intervals between the first and third quartiles.}
    \label{fig: object_error_rotate_sphere}
\end{figure}

\subsection{Robustness against Noise and Modeling Errors}
\label{subsec: robustness_noise_modeling_error}
First, to study the performance of the proposed framework under sensor noise, we again use the \textbf{Rotate Sphere} benchmark, and randomly perturb the observed sphere orientation at each time step. The perturbation comprises a random axis, with the angle following a normal distribution $\mathcal{N}(0,\sigma^2)$. We track the same 100 random target orientations in Sec.~\ref{subsec: highlevel_comparison_baselines}. The task error and task error S.D. under different $\sigma$ are shown in Fig.~\ref{fig: object_error_rotate_sphere}. Task error varies modestly and nonmonotonically with noise; its S.D. increases considerably. The results show that sensor noise causes unwanted finger and object motions.
In addition, larger sensor noise results in longer convergence times, as demonstrated in the \href{https://drive.google.com/file/d/1crsK6mrv8RD1j8DSZDTmUYkpKgaWSTH0/view?usp=sharing}{\textcolor{cyan!70!black}{attached video}}.
Overall, the proposed method is robust against sensor noise, owing to the online planning and long-term predictive ability that reduce sensitivity to noisy observations.

\begin{figure}[!tb]
    \centering
    \includegraphics[width=1\linewidth]{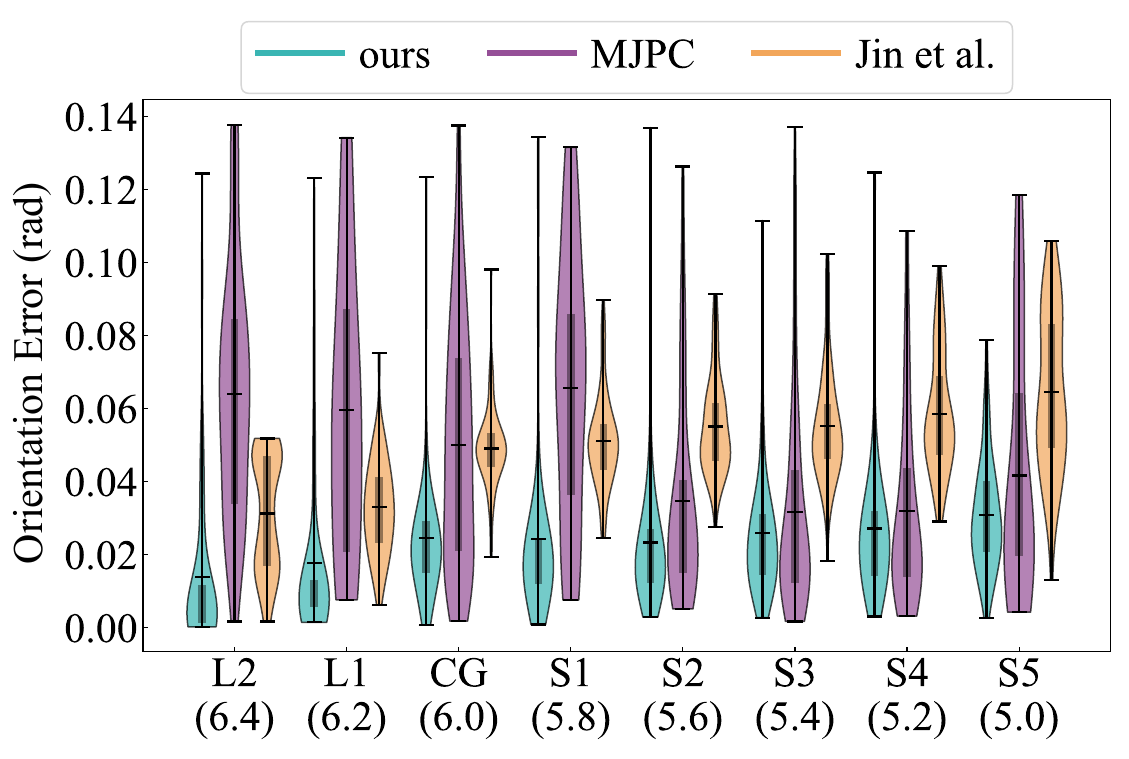}
    \caption{{Distribution of the task error across 100 trials in the \textbf{Rotate Sphere} task, under modeling errors. We use a consistent radius of $r=6.0\mathrm{cm}$ in the planning model across different groups. The x-axis labels indicate the actual radius used in the MuJoCo simulation ($5.0\sim6.4 \mathrm{cm}$). The control group (CG) has zero modeling error. Groups L1 to L2 have a larger radius, whereas groups S1 to S5 have a smaller radius for the simulated sphere.}}
    \label{fig: ori_error_sphere_rotate_model_error}
\end{figure}

Second, to explore the effects of modeling errors, we vary the radius of the sphere in the simulation from $5.0$ to $6.4 \mathrm{cm}$. We compare the proposed framework with MJPC and the method of \citet{jin2024complementarity}, which demonstrate comparable performance to Ours (2\textsuperscript{nd}) in Tab.~\ref{tab: sim_comparison_sphere_rot_task}. {The nominal sphere radius remains $6.0 \mathrm{cm}$ in the planning module for all methods.} The distribution of task error among all 100 trials is shown in Fig.~\ref{fig: ori_error_sphere_rotate_model_error}. For the proposed framework, as the simulated sphere becomes smaller, the task error slightly increases because the fingers need to make more significant adjustments to the planned trajectory to generate sufficient contact forces. Meanwhile, the task error decreases when the simulated sphere becomes larger.
{In contrast, MJPC exhibits a much wider error distribution due to jerky motions, indicating less stable performance.}
The task error decreases at $r<5.8\mathrm{cm}$ because the fingers are in contact with the sphere less often, resulting in fewer unwanted motions. In addition, the method of \citet{jin2024complementarity} exhibits a noticeably wider error distribution as the modeling error increases, indicating degraded performance. Overall, the proposed method achieves higher accuracy and increased stability under modeling errors compared to the baselines, indicating that tactile-informed motion correction effectively strengthens model-based control against modeling uncertainties.

Besides sensor noise and modeling errors, complex geometry is another major challenge for real-world in-hand manipulation. Additional results and analysis under complex geometries are provided in Appendix~B.

\subsection{Comparison of Different Weighting Schemes for Hybrid Force-Motion Control}
\label{subsec: comparison_weighting_scheme}

\begin{figure}[!t]
    \centering
    \includegraphics[width=0.95\linewidth]{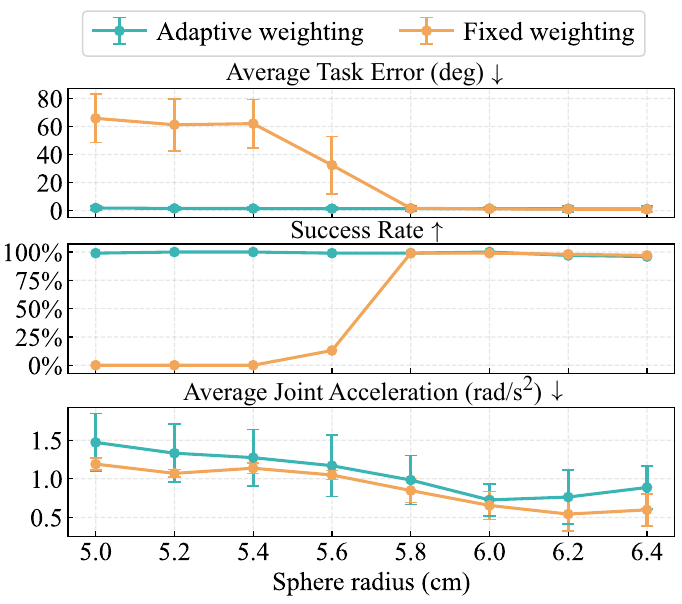}
    \caption{Performance comparison between the proposed adaptive weighting scheme and the fixed-weight baseline on the \textbf{Rotate Sphere} task under modeling errors. The modeling error is introduced by using different sphere radii in simulation and the planning model, as in Fig.~\ref{fig: ori_error_sphere_rotate_model_error}. Markers denote mean values, and error bars represent standard deviations.}
    \label{fig: comparison_weight_scheme}
\end{figure}

To examine whether the proposed adaptive weighting design is truly beneficial, we further compare it with a carefully tuned fixed-weight baseline on the \textbf{Rotate Sphere} benchmark. The comparison is conducted using the same 100 target orientations as in Sec.~\ref{subsec: highlevel_comparison_baselines}. The fixed-weight baseline is tuned by parameter sweeping beforehand (Appendix~C), and the best-performing setting is used for comparison. As shown in Fig.~\ref{fig: comparison_weight_scheme}, the proposed adaptive weighting scheme achieves clear advantages under larger modeling errors, particularly when the actual sphere radius is smaller than that assumed in the planning model and contact loss occurs more frequently. In such cases, adaptive weighting can better adjust the relative emphasis between force tracking and position tracking according to the contact normal direction, which helps recover and maintain contact under noticeable modeling errors. The fixed-weight scheme, in contrast, relies on a single compromise throughout execution. We also observe slightly higher average joint acceleration under adaptive weighting, which reflects more responsive compensation for contact recovery. Overall, the proposed adaptive weighting mechanism preserves comparable performance when the modeling error is mild, while providing clear advantages in the more challenging and practically relevant cases where contact loss is more likely.

To further illustrate the mechanism behind these advantages, Appendix~D visualizes the evolution of the adaptive weights during representative tasks and analyzes their correlation with key contact events such as establishment and breaking.

\subsection{Motion and Contact-Force Tracking with Tactile Feedback}
\label{subsec: exp_tac_feedback_ctrl}

\begin{figure}[!t]
    \centering
    \includegraphics[width=0.95\linewidth]{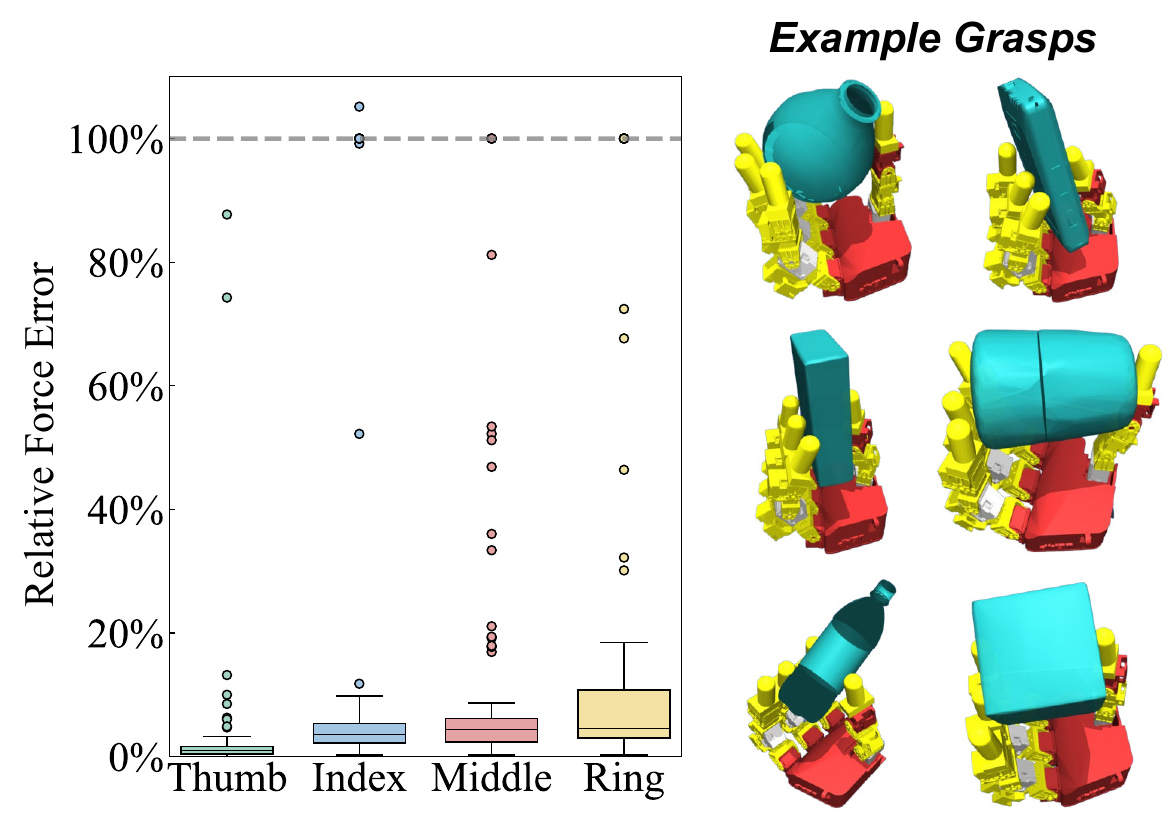}
    \caption{Distribution of the relative force tracking error when executing 100 grasps generated using BODex \citep{chen2024bodex}. The scatter points represent outliers that lie beyond 1.5 times the inter-quartile range (IQR) from the box. Points that exactly fall on the 100\% line correspond to zero contact force (i.e., missing contacts), while points above the 100\% line indicate excessively large contact forces (i.e., grasp failure). Examples of the hand configuration at the pre-grasping stage are shown on the right.
    }
    \label{fig: grasp100_force_error}
\end{figure}

\begin{figure}[!t]
    \centering
    \includegraphics[width=0.85\linewidth]{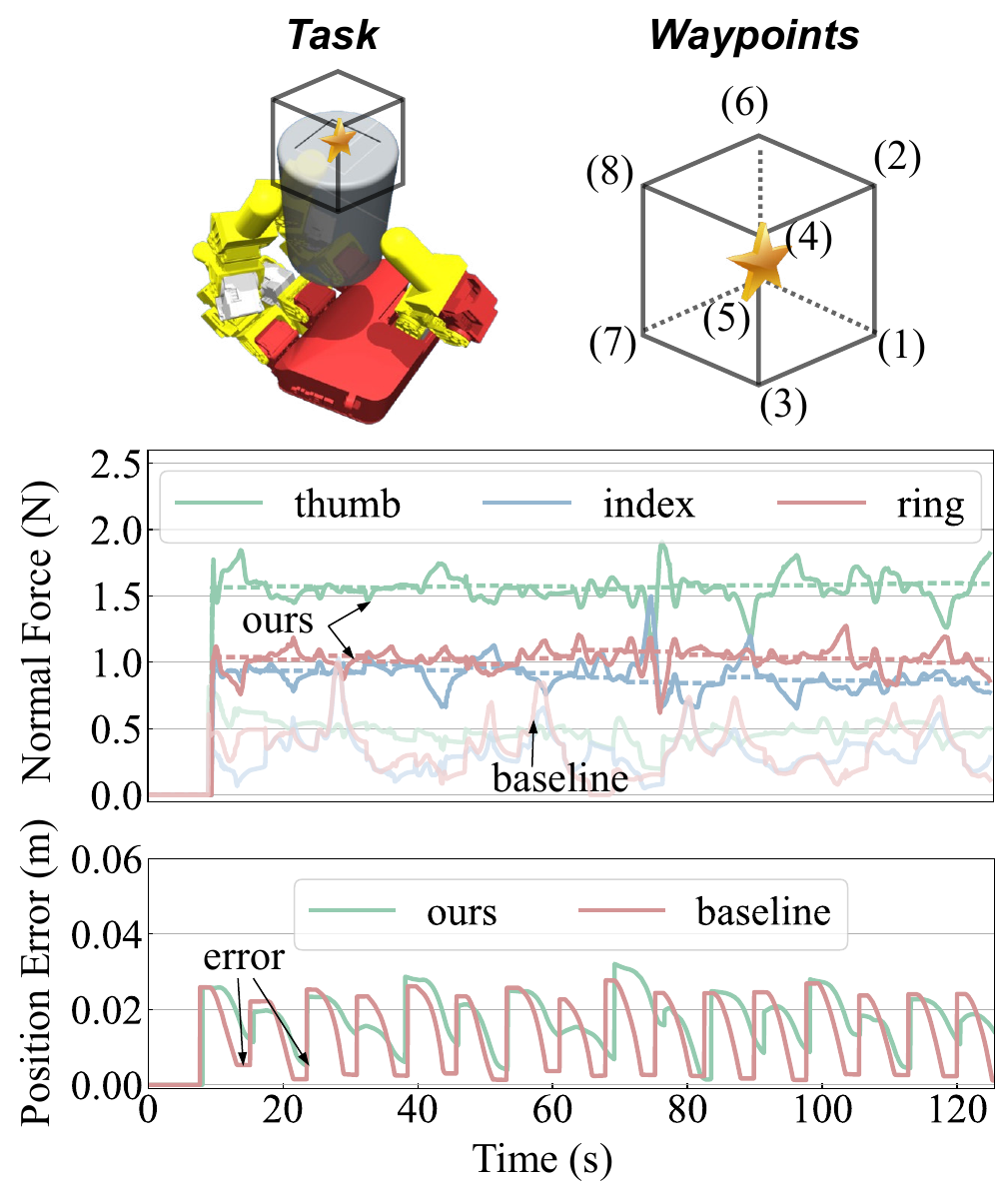}
    \caption{
    Results of the in-grasp object movement experiment. \textbf{Top: Task snapshot and waypoints}. The marker at the top center of the cylinder successively moves to the eight corners of the cube and back to the initial position. \textbf{Middle: Normal contact force}. Dashed lines represent the desired values, whereas solid lines show the execution results. The three dark lines indicate the results of the proposed controller, and the three light lines represent the baseline results. \textbf{Bottom: Position tracking error}. The local minima indicate tracking errors when the marker reaches the waypoints.
    }
    \label{fig: in_grasp_result_force_and_com}
\end{figure}

While the preceding experiments demonstrate simultaneous motion and force tracking in environment-constrained tasks, object motion remains partly supported by task-specific constraints. We further test the low-level tactile-feedback controller in scenarios where the object is constrained only by hand contacts, making successful execution more directly dependent on contact-force regulation. To isolate this evaluation from the high-level module, we conduct two experiments, static grasping and in-grasp manipulation, with predefined reference finger motions and contact forces.

\begin{figure}[!t]
    \centering
    \includegraphics[width=1\linewidth]{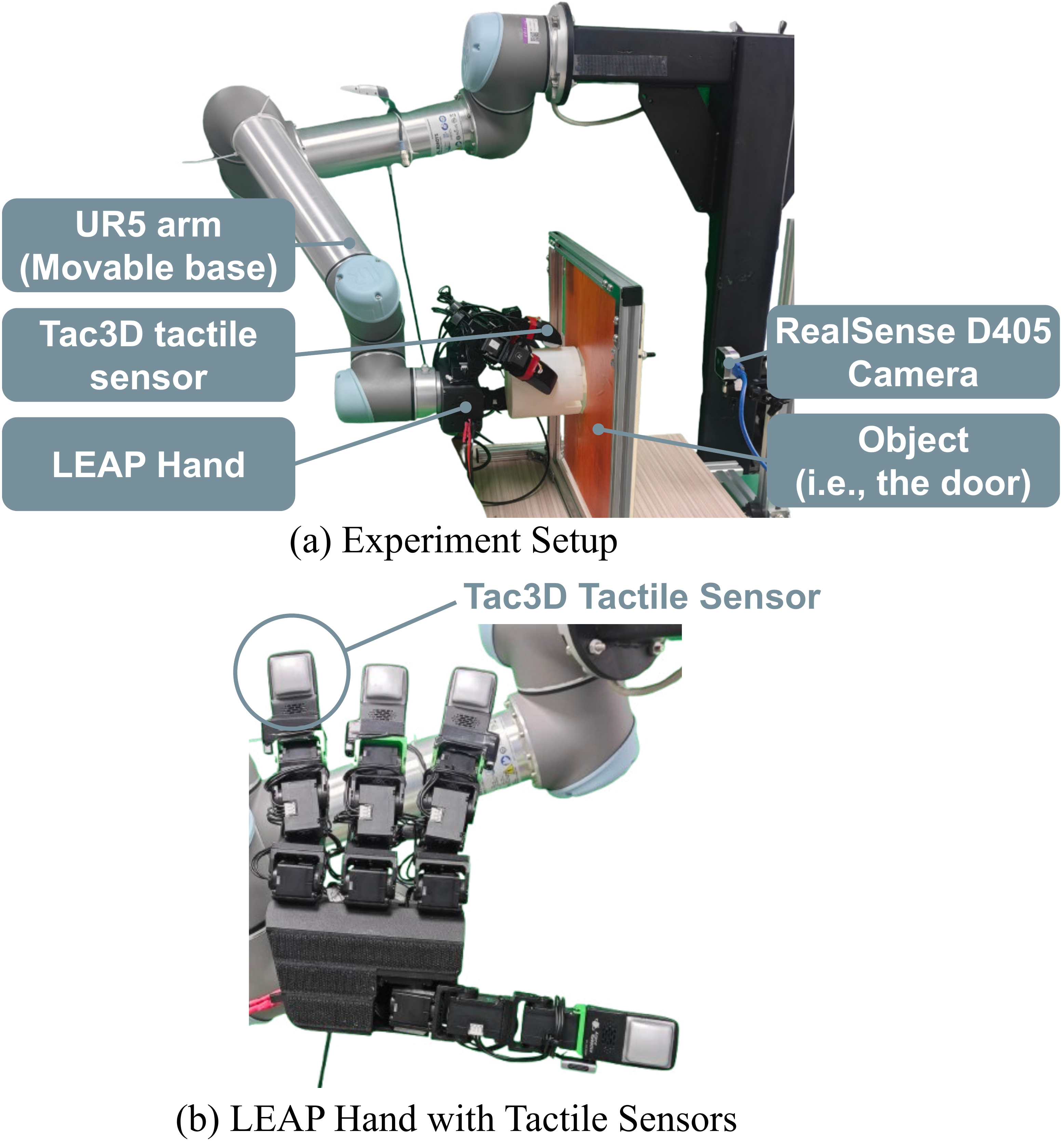}
    \caption{Experimental setup: (a) The LEAP Hand is mounted on the flange of a UR5 arm. An AprilTag and a RealSense D405 camera are used for perception. (b) The fingertips of the LEAP Hand are replaced with four Tac3D tactile sensors.
    }
    \label{fig: experiment_setup}
\end{figure}

\begin{figure*}[!t]
    \centering
    \includegraphics[width=\linewidth]{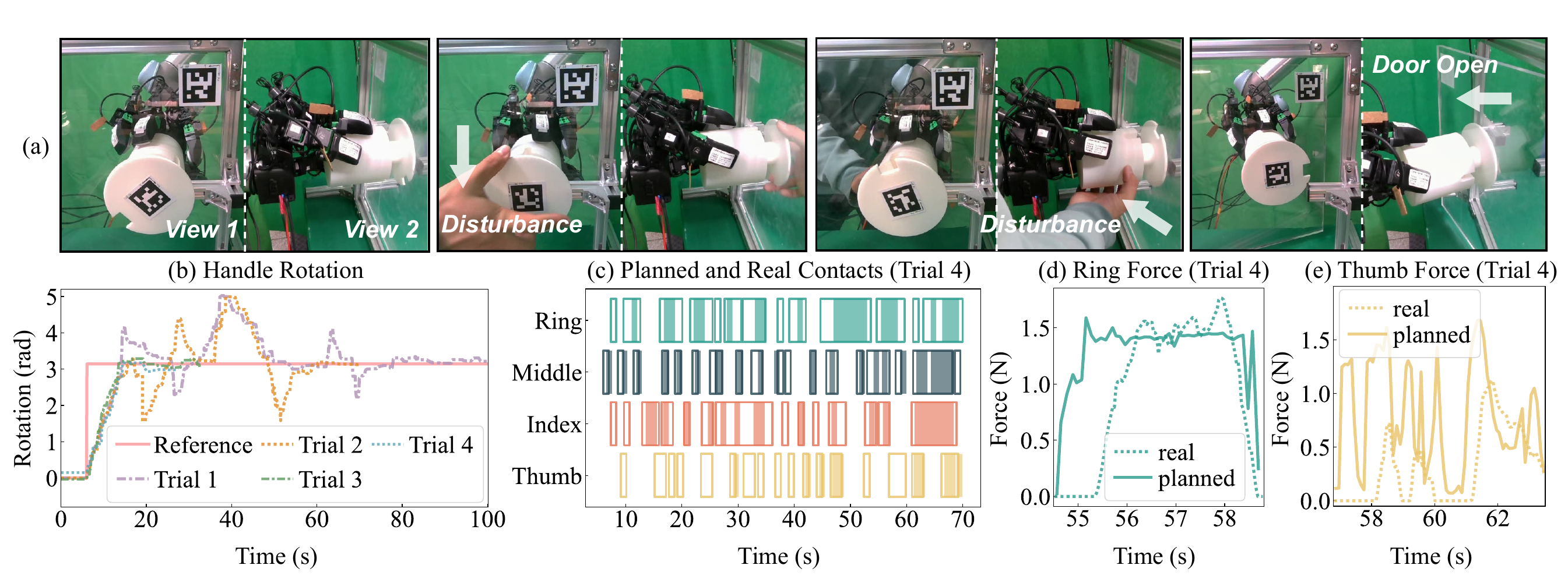}
    \caption{Experimental results for the \textbf{Open Door} task with human disturbance. (a) Snapshots from two views. (b) Door handle rotation in different trials. (c) Planned and real contacts (Trial 4) represented as time intervals during which the contact forces exceed a given threshold. (d)(e) Planned and real contact forces.}
    \label{fig: open_door_real_results}
\end{figure*}

\subsubsection{Static Grasping Experiment.}
\label{subsubsec: static_grasping_experiment}

The static grasping experiment is designed to primarily validate the capability of force tracking. We generate one grasp per object for the first 100 objects in the DGN test set using BODex \citep{chen2024bodex}. As we are not proposing a grasping controller, we simplify the experiments by adding objects with damped free-floating joints and disabling gravity. In this way, we focus on force tracking and neglect other effects. The static grasping is divided into three stages. 1) \textbf{Pre-Grasping}: The hand is initialized at the grasp pose with fingers slightly open to avoid contacts. 2) \textbf{Normal Estimation}: The fingers close gradually until the contact forces reach the pre-defined threshold. The stacked contact normals $\bm{N}=\left[\bm{n}_\text{thumb},\bm{n}_\text{index},\bm{n}_\text{middle},\bm{n}_\text{ring}\right] \in \mathbb{R}^{3 \times 4}$ are obtained. 3) \textbf{Grasp Control}: The proposed controller tracks desired contact forces. The desired forces are in the directions of the contact normals, and the magnitudes $\bm{f} \in \mathbb{R}^{4}$ are decided by solving the following optimization:
\begin{equation}
    \min_{\bm{f}} \quad \Vert \bm{G}_o\bm{N}_\text{aug}\bm{f} \Vert + \lambda \Vert \bm{f}-m\bm{1} \Vert
    \label{eq: exp_force_magnitude_optimization}
\end{equation}
where $\bm{1}$ is an all-ones vector, $\bm{N}_\text{aug} \in \mathbb{R}^{12 \times 4}$ is the augmentation of $\bm{N}$, and $\lambda$ is a weighting factor. Note that (\ref{eq: exp_force_magnitude_optimization}) penalizes unbalanced wrenches and regulates the force magnitudes around a pre-set value $m$.

Distribution of the relative force tracking error across 100 grasp trials is shown in Fig.~\ref{fig: grasp100_force_error}. The proposed controller achieves precise tracking of contact forces and exhibits robust generalization capabilities across various object shapes. Imperfections in force tracking primarily stem from two sources: contacts that are either missing or unstable at the object's outer edges, and contacts occurring on parts of the finger other than the fingertip, where no force feedback is available (i.e., only forces on the fingertips are sensed).

\subsubsection{In-Grasp Object Movement Experiment.}
\label{subsubsec: in_grasp_manipulation_experiment}

The in-grasp object movement experiment is designed to evaluate the ability to jointly track finger forces and motions.
In this experiment, the LEAP Hand grasps a cylinder using three fingers, excluding the middle finger. The marker at the cylinder's top center is directed to sequentially reach waypoints at the eight vertices of a 3-cm cube. After reaching each waypoint, the marker returns to its initial position. This task can be accomplished by planning and tracking finger motions through optimization. The finger movements are generated using code from \citet{yu2025rgmc}, and we re-calculate the desired contact forces using (\ref{eq: exp_force_magnitude_optimization}) before each waypoint. We compare our proposed controller with the baseline from \citet{yu2025rgmc}, which focuses exclusively on finger motion tracking. The normal contact force and position error are illustrated in Fig.~\ref{fig: in_grasp_result_force_and_com}. Compared with the baseline, the proposed controller slightly sacrifices position accuracy to prioritize force tracking. Overall, the proposed low-level controller effectively tracks both finger motion and contact forces.
It accurately tracks the desired normal force, whereas the baseline exhibits appreciable fluctuations.

\section{Real-World Experiment Results}
\label{sec: real_world_experiments}
{The real-world experiments evaluate the proposed framework under two complementary settings. In the environment-constrained tasks (Sec.~\ref{subsec: non_prehensile_tasks}), external structures restrict object motion; within this setting, the experiments primarily test online adaptability and robustness to modeling uncertainties and disturbances. In the prehensile in-hand tasks (Sec.~\ref{subsec: prehensile_tasks}), the object is maintained within the hand through coordinated finger contacts, providing a more direct evaluation of simultaneous motion tracking and contact-force regulation.}

\begin{figure}[!t]
    \centering
    \includegraphics[width=0.77\linewidth]{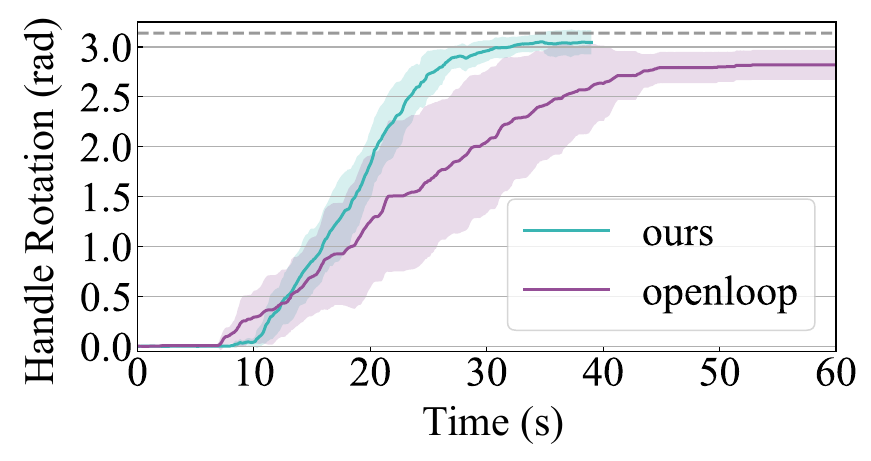}
    \caption{Real-world \textbf{Open Door} task—Comparison with the ablation without contact tracking. Here, `ours' refers to \textbf{Ours (2\textsuperscript{nd})}, and `openloop' refers to \textbf{Ours (2\textsuperscript{nd}, openloop)} in Tab.~\ref{tab: sim_comparison_sphere_rot_task}. The curves represent the mean handle rotation, while the shaded areas indicate the standard deviation. The dashed horizontal line marks the target rotation.}
    \label{fig: compare_real_open_door}
\end{figure}

\subsection{Experimental Setup}
The experimental setup is shown in Fig.~\ref{fig: experiment_setup}. We mount the LEAP Hand on the flange of a UR5 arm, which serves as a movable base, and the wrist movements are not utilized during in-hand manipulation. The joint states of the LEAP Hand are queried at 60 Hz. {The original fingertips of the LEAP Hand are replaced with four Tac3D vision-based tactile sensors \citep{zhang2022tac3d}, which internally process marker-array images to provide contact points and forces at 30 Hz. The measured forces are further low-pass filtered before being supplied to the low-level MPC.}
{Furthermore, explicit object-pose feedback is obtained at 30 Hz using AprilTags and a RealSense D405 camera and supplied as state feedback to the high-level MPC.}
Key parameters used in the real-world experiments are summarized in Appendix~E.

\subsection{Environment-Constrained Manipulation}
\label{subsec: non_prehensile_tasks}

\begin{figure}[!t]
    \centering
    \includegraphics[width=\linewidth]{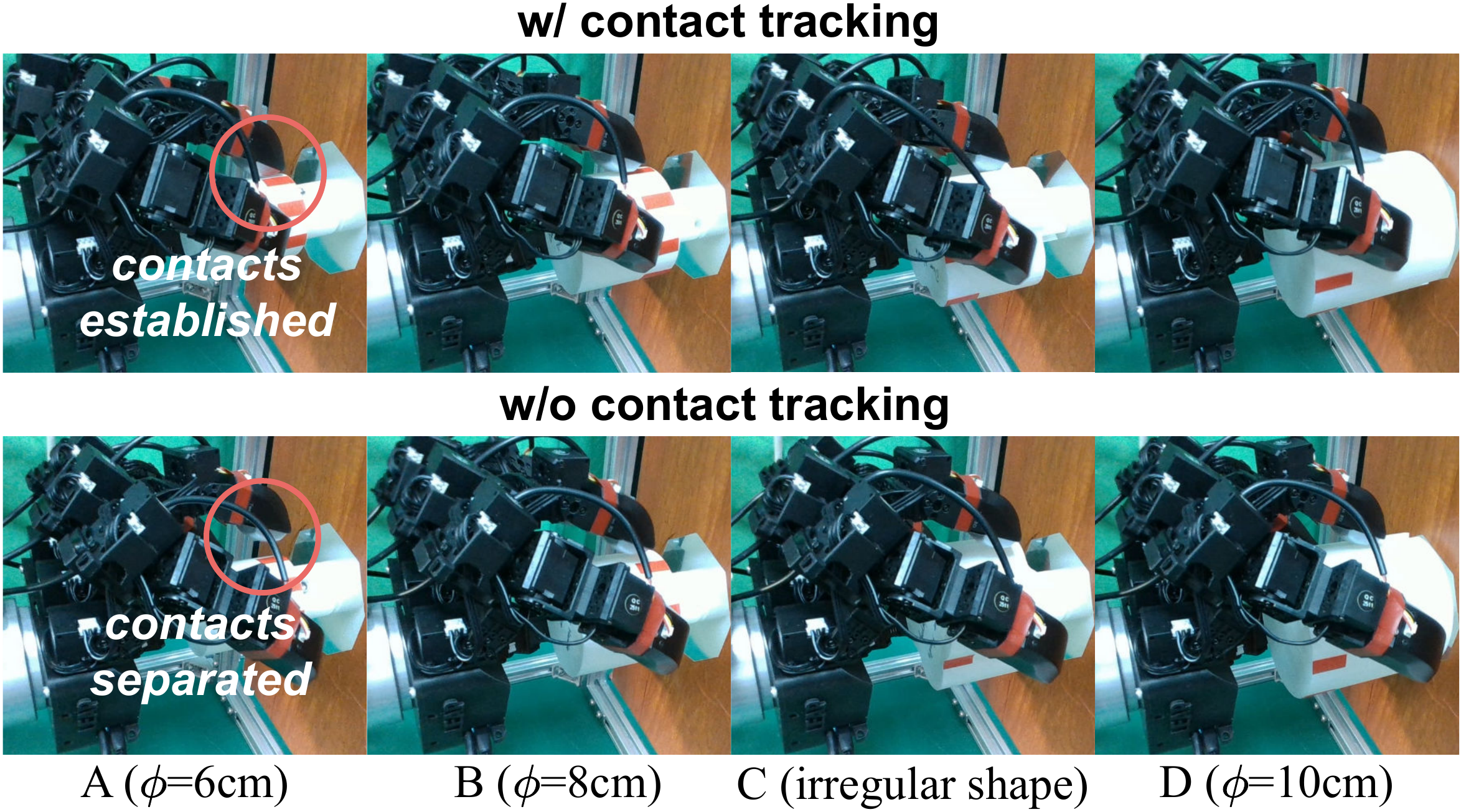}
    \caption{{Snapshots of the \textbf{Open Door} task under size and shape uncertainties. The proposed framework with contact tracking establishes the planned contacts despite substantial modeling errors, while the ablation without contact tracking shows insufficient contact force or even missed contacts.}}
    \label{fig: open_door_uncertainty_snapshots}
\end{figure}

\begin{figure*}[!t]
    \centering
    \includegraphics[width=0.99\linewidth]{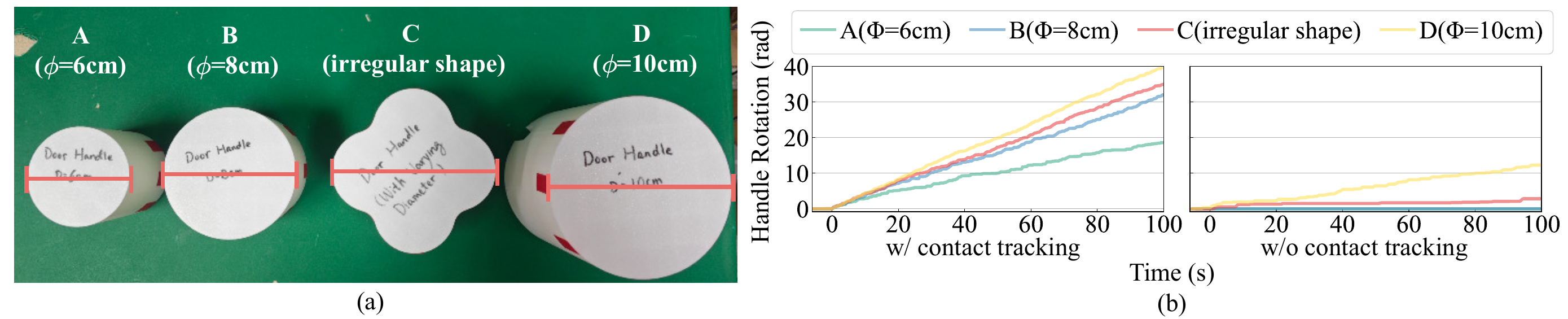}
    \caption{{(a) Real-world door handles of different sizes and shapes used in the \textbf{Open Door} task. Handles A ($\phi = 6$ cm) and B ($\phi = 8$ cm) introduce object-size uncertainties, while handle C—an irregular shape with a maximum diameter of 10 cm—represents partially defective objects. (b) Handle rotation angles over time for different handles and control methods, where `w/ contact tracking' and `w/o contact tracking' refer to \textbf{Ours (2\textsuperscript{nd})} and \textbf{Ours (2\textsuperscript{nd}, openloop)} in Tab.~\ref{tab: sim_comparison_sphere_rot_task}, respectively.}}
    \label{fig: open_door_uncertainty_plot}
    \vspace{-0.2cm}
\end{figure*}

\subsubsection{Experiments Involving the Open Door Task.}

We customize a door model with a cylindrical handle and hinge joint to accommodate the LEAP Hand, as shown in Fig.~\ref{fig: open_door_real_results} (a). The task first requires in-hand manipulation to turn the door handle 180 degrees so that the notch is aligned with the door latch, which requires high precision. The hand then pulls open the door as shown in the snapshot. The rotation of the door handle during four trials is shown in Fig.~\ref{fig: open_door_real_results} (b). Human intervention is applied in Trials 1 and 2, as indicated by the peaks. The door handle is turned to the target orientation even under disturbances. The planned and real contact forces during Trial 4 are visualized in Fig.~\ref{fig: open_door_real_results} (c)$\sim$(e). In Fig.~\ref{fig: open_door_real_results} (c), the solid-bordered boxes and the shaded regions indicate time intervals during which the planned and real contact forces exceed the specified threshold $\SI{0.1}{N}$ (time intervals shorter than $\SI{1}{s}$ are regarded as noise and neglected), respectively. If the planned forces are ideally tracked, the shadows will fill 100\% of the boxes. However, there is an obvious tracking delay, especially when the planned forces are not maintained for some time (i.e., high-frequency oscillations), as shown in Fig.~\ref{fig: open_door_real_results} (d)(e). The imperfections are primarily due to the relatively low frame rate of the tactile sensors and inaccurately modeled servo characteristics. We further demonstrate one advantage of the model-based scheme in Appendix~E: its ability to readily adapt to modified task setups.

\begin{figure*}[!b]
    \centering
    \includegraphics[width=1\linewidth]{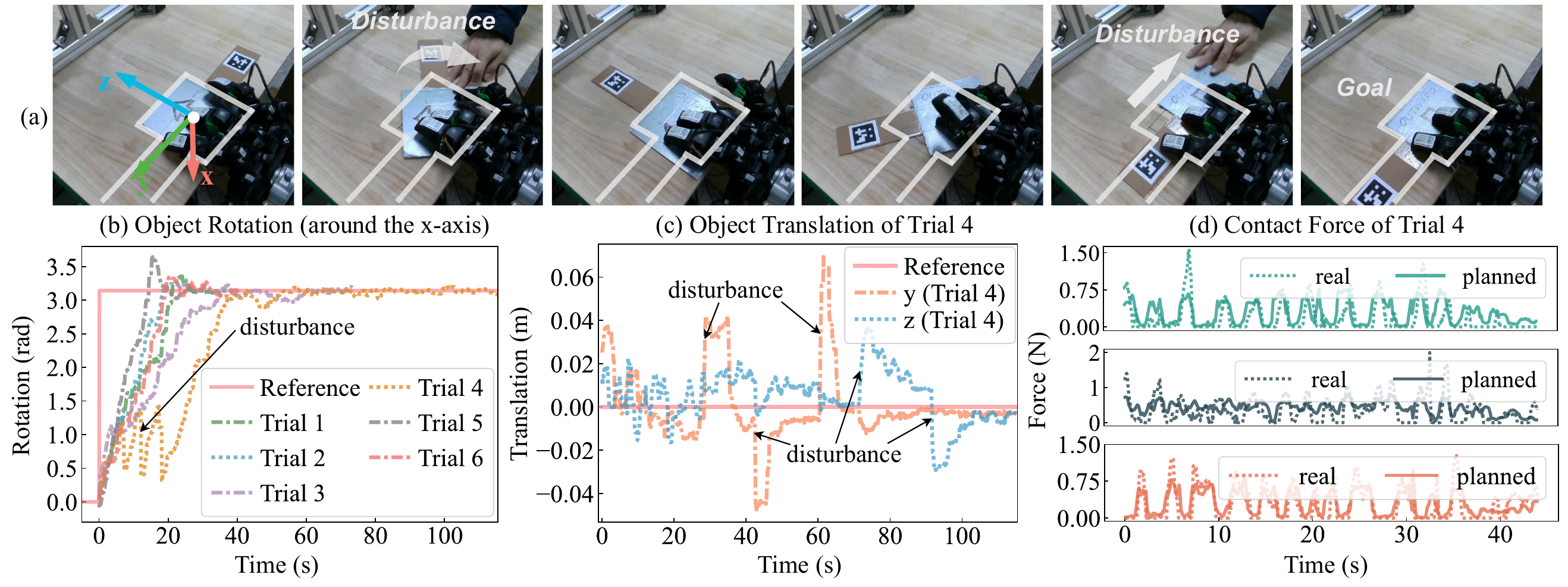}
    \caption{Experimental results for the \textbf{Rotate Card} task with disturbance. (a) Snapshots of the task. The white border represents the target pose. Rotational and translational disturbances are exerted by the human operator. (b)(c) Object rotation (around the x-axis) and translation (on the yOz plane) refer to the coordinate system shown in the first snapshot in (a). (d) Planned and measured contact forces of the ring, middle, and index fingers (from the top down).}
    \label{fig: rotate_card_real_results}
\end{figure*}

\subsubsection{{Ablation Studies of Contact tracking}.}
\label{subsubsec: comparison_with_baselines_open_door}

To demonstrate the importance of tactile-informed contact tracking, we conducted the open door task 10 times using both the proposed method {and the ablation \textbf{Ours (2\textsuperscript{nd}, openloop)}}. As shown in Fig.~\ref{fig: compare_real_open_door}, the proposed method converges faster and exhibits lower orientation error. In contrast, {\textbf{Ours (2\textsuperscript{nd}, openloop)}} performs poorly and inconsistently due to the fingers exerting insufficient contact force without contact tracking.
{Additionally, as shown in Fig.~\ref{fig: open_door_uncertainty_plot} (a), we use door handles with different sizes and shapes to evaluate the performance under modeling uncertainties, similar to the \textbf{Rotate Sphere} task in simulation (Fig.~\ref{fig: ori_error_sphere_rotate_model_error}). In the planning and control model, the handle diameter is set to 10 cm, consistent with the real handle D ($\phi = 10$ cm). The handle rotation curves over time are shown in Fig.~\ref{fig: open_door_uncertainty_plot} (b), with representative snapshots included in Fig.~\ref{fig: open_door_uncertainty_snapshots}. With the low-level controller for contact tracking, the dexterous hand is able to establish contacts and manipulate the handle despite substantial modeling errors. In contrast, without contact tracking, \textbf{Ours (2\textsuperscript{nd}, openloop)} simply follows the planned finger trajectories generated for the ``large and cylindrical'' handle, and thus often fails to establish proper contacts. Consequently, the handle barely rotates when its diameter is reduced by 20\% or 40\%.}

\subsubsection{Experiments Involving the Rotate Card Task.}
The Rotate Card task requires a paper card to be rotated 180 degrees while keeping the COM position stationary. The start and goal pose are visualized as the physical card and the ghost white border in the first snapshot of Fig.~\ref{fig: rotate_card_real_results} (a), respectively. We attach an AprilTag to the extended part of the card to avoid occlusions. The task is challenging because it relies on the proper finger-gaiting pattern to produce torques that create the correct center of rotation. To test the robustness of the algorithm, we exert disturbances (rotational and translational) as shown in Fig.~\ref{fig: rotate_card_real_results} (a). The card's x-axis rotation in six trials is displayed in Fig.~\ref{fig: rotate_card_real_results} (b).
All six trials successfully rotate the card to its target pose even under disturbances, as shown in Fig.~\ref{fig: rotate_card_real_results} (b). The card translation on the yOz plane is plotted in Fig.~\ref{fig: rotate_card_real_results} (c). 
The error decreases immediately after disturbances, thanks to the high-level planner's online (re)planning capability.
In addition, the planned and measured contact forces of the ring, middle, and index fingers are shown in Fig.~\ref{fig: rotate_card_real_results} (d) (from the top down). Overall, the measured forces follow the references well. The planned middle finger force is almost always non-zero, whereas the index and ring fingers alternately make contact, showing auto-generated finger gaiting.

\begin{figure*}[!t]
    \centering
    \includegraphics[width=1\linewidth]{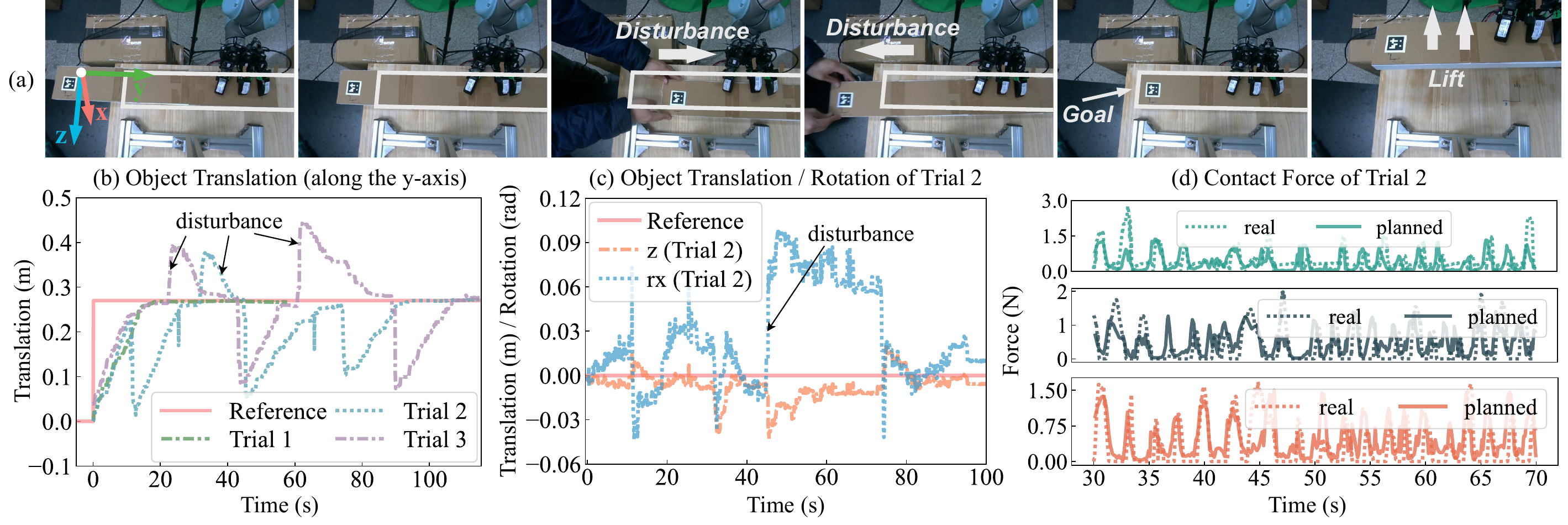}
    \caption{Experimental results for the \textbf{Slide Board} task with disturbance. (a) Snapshots of the task. The white border represents the target pose. External disturbances are exerted by the human operator. (b)(c) Object rotation and translation refer to the coordinate system shown in the first snapshot in (a), 'z' refers to translation along the z-axis, and 'rx' refers to the rotation around x-axis. (d) Planned and measured contact forces of the ring, middle, and index fingers (from the top down).}
    \label{fig: slide_board_real_results}
    \vspace{-0.3cm}
\end{figure*}

\begin{figure*}[!b]
    \centering
    \vspace{-0.3cm}
    \includegraphics[width=1\linewidth]{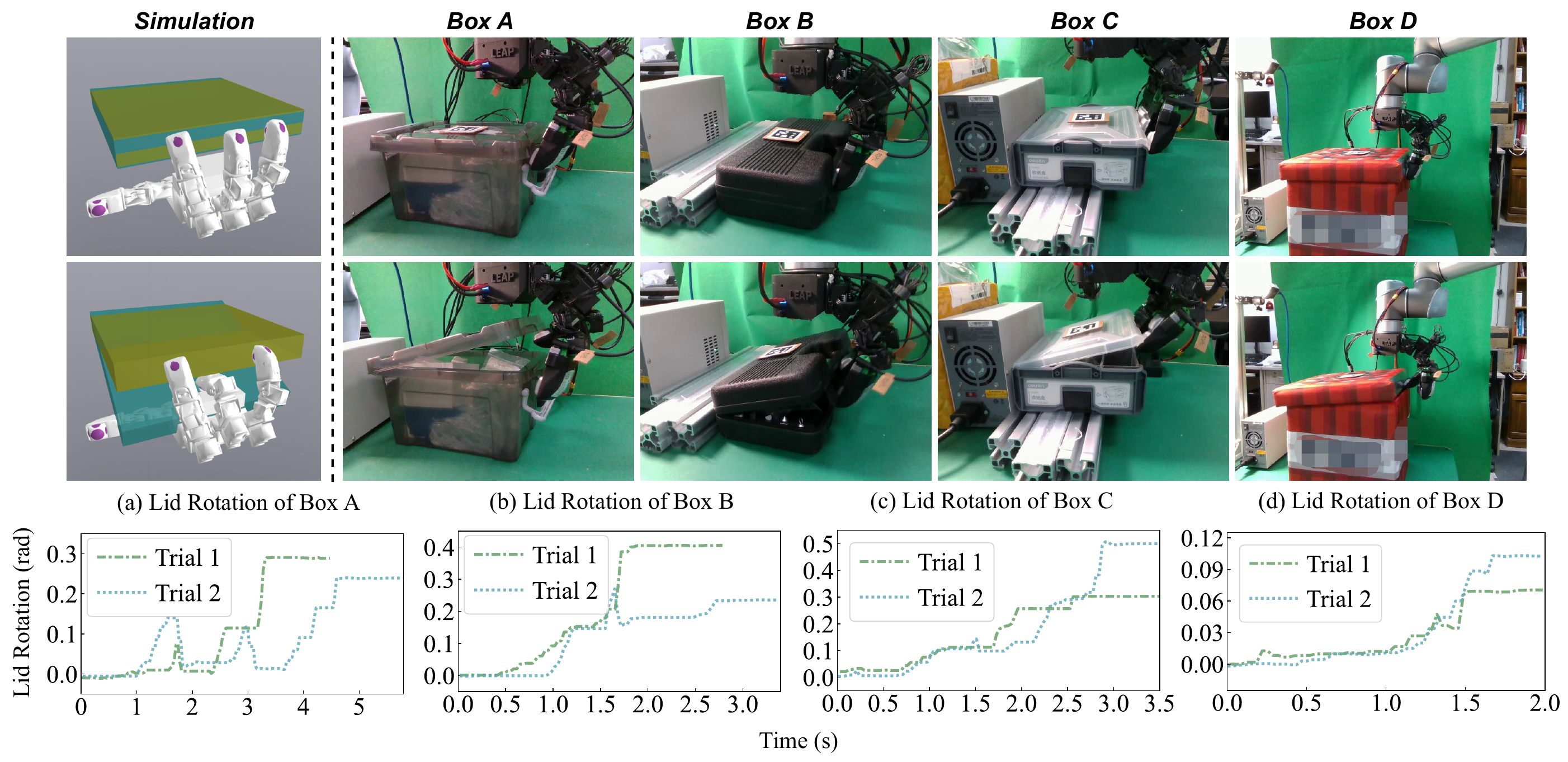}
    \caption{Experimental results for the \textbf{Open Box} task. At the top are snapshots of the simulation and the real-world experiments on four boxes A$\sim$D. The first row shows the initial state, and the second row shows the state when the box is opened. The yellow and cyan boxes represent the initial and opened positions of the lid, respectively. (a)$\sim$(d) Lid rotation of different boxes.}
    \label{fig: open_box_real_results}
\end{figure*}

\subsubsection{Experiments Involving the Slide Board Task.}
The Slide Board task requires the board to be translated approximately 27 cm along the y-axis so that it can be lifted by grasping the middle. Task snapshots are shown in Fig.~\ref{fig: slide_board_real_results} (a). The target pose is visualized by the white border. 
The board translation and rotation during the manipulation are shown in Fig.~\ref{fig: slide_board_real_results} (b) and (c). The board quickly recovers from translational disturbance but recovers more slowly from rotational disturbance because the frictional torque is greater. The rotational disturbance changes suddenly at approximately 74 s in Fig.~\ref{fig: slide_board_real_results} (c) due to human intervention. Fig.~\ref{fig: slide_board_real_results} (d) presents both the planned and measured forces, along with a visualization of the finger gaiting behavior.

\subsubsection{Experiments Involving the Open Box Task.}
To validate the ability of the proposed method to generate non-periodic finger motions, we implement the Open Box task. The snapshots and results are shown in Fig.~\ref{fig: open_box_real_results}. This task requires the fingers to lift the lid to a certain angle using friction, and then lift the lid with one of the fingers, as shown in the simulation snapshots. The angle is set differently for different boxes. This task has a relatively large modeling error because the contact parts of the lid have geometries different from those of the simplified cube in the model, as shown in the snapshots. We observe that this task has a low success rate (fluctuating between 20\% and 50\% for different boxes). For each box, we plot the results of two successful trials in Fig.~\ref{fig: open_box_real_results} (a)--(d). Box D is the most difficult to open because the lid is heavy and requires large fingertip movement (i.e., the radius of rotation is large). Despite the challenges, the task is successfully completed using the proposed method, which demonstrates generalization capabilities and robustness against modeling errors.

\subsubsection{{Demonstration of the Rotate Valve Task.}}

In addition to the real-world tasks above, Appendix~E presents a Rotate Valve experiment under a modified hardware setup, demonstrating long-horizon in-hand manipulation.

\begin{figure*}[tp]
    \centering
    \includegraphics[width=1\linewidth]{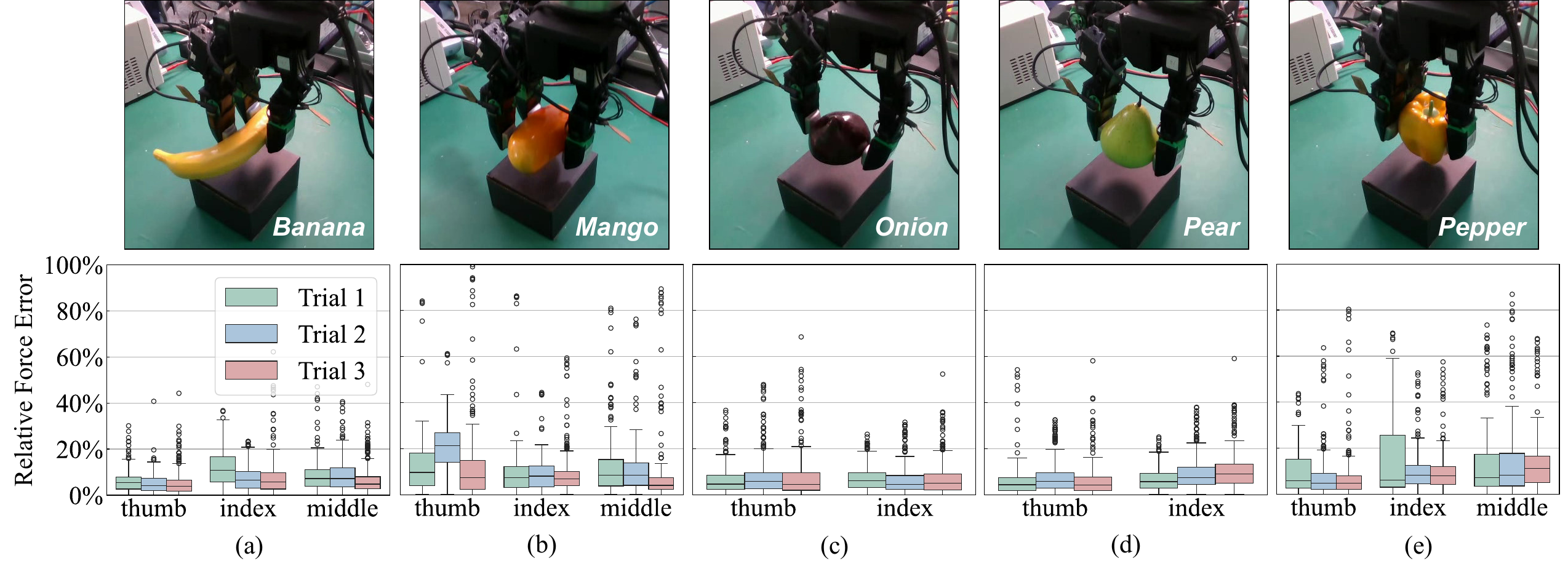}
    \caption{Relative force tracking error and snapshots of the grasping experiment. The objects are (a) banana, (b) mango, (c) onion, (d) pear, and (e) pepper. Scatter points denote outliers beyond the 1.5-IQR whiskers.}
    \label{fig: real_grasp_exp_5_fruits}
    \vspace{-0.2cm}
\end{figure*}

\subsection{Prehensile In-Hand Manipulation}
\label{subsec: prehensile_tasks}

Beyond the environment-constrained experiments, we further evaluate the proposed method on three real-world prehensile manipulation tasks in which the object is maintained within the hand through coordinated finger contacts.
{Compared with environment-constrained tasks, these prehensile experiments require regulating object pose relative to the hand while maintaining contact forces, providing a more direct evaluation of motion–force coordination.}
For prehensile manipulation, the LEAP Hand is mounted on a fixed frame instead of the movable UR5 flange.

\subsubsection{Real-World Static Grasping Experiment.}
As a controlled prehensile setting, we first evaluate the low-level controller in a real-world static grasping experiment, using a setup similar to that in Sec.~\ref{subsubsec: static_grasping_experiment}.
We execute grasping three times for each of the five plastic objects. 
The results are shown in Fig.~\ref{fig: real_grasp_exp_5_fruits}. 
The relative force tracking error is small and remains consistent between multiple trials, in alignment with the simulation studies.
The experimental results demonstrate that the proposed tactile-feedback controller is reliable for tracking desired contact forces using the real-world hardware.

\begin{figure}[!t]
    \centering
    \includegraphics[width=1\linewidth]{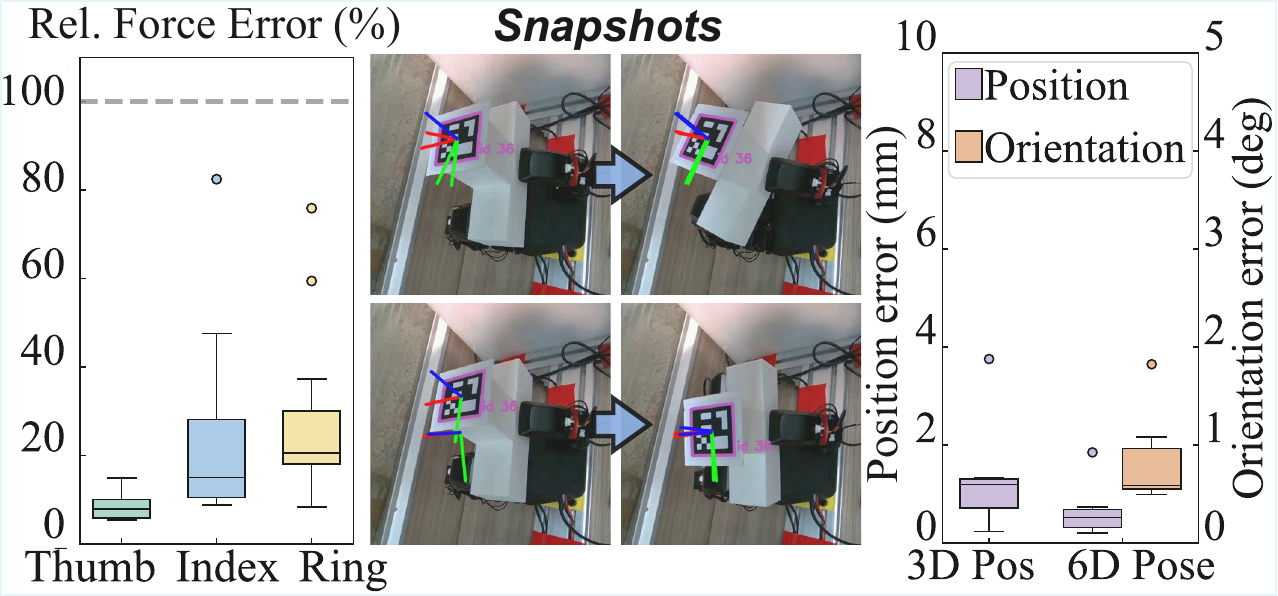}
    \caption{Results of the in-grasp object movement experiment. The object tracks target positions or poses through rolling contacts while maintaining contact-force regulation. \textbf{Left}: relative force tracking errors of the three contacting fingers. \textbf{Middle}: snapshots of target pose tracking. \textbf{Right}: object tracking errors for the position-only (3D Pos) and full-pose target (6D Pose) groups.}
    \label{fig: real_in_grasp_manip_task}
\end{figure}

\subsubsection{Real-World In-Grasp Object Movement Experiment.}

We conduct an in-grasp object movement experiment, as shown in Fig.~\ref{fig: real_in_grasp_manip_task}, with a setting similar to Sec.~\ref{subsubsec: in_grasp_manipulation_experiment}. In this task, the object is securely grasped by the fingers, and the AprilTag-marked feature frame is controlled to track either 3D target positions or 6D target poses through rolling contacts. The targets include six positions and six poses, with translational and rotational displacements of 2.5--4 cm and 14--23 deg from the initial state, respectively. Fig.~\ref{fig: real_in_grasp_manip_task} reports the relative contact-force tracking errors of the three contacting fingers, as well as the object position and orientation tracking errors. The smaller position errors in hardware are partly attributable to the use of three replanning trials for each target. Overall, the results validate the controller’s ability to coordinate object motion tracking and contact-force regulation during prehensile in-hand manipulation.

\subsubsection{Real-World Prehensile Manipulation with Finger Gaiting.}

\begin{figure}[!t]
    \centering
    \includegraphics[width=0.95\linewidth]{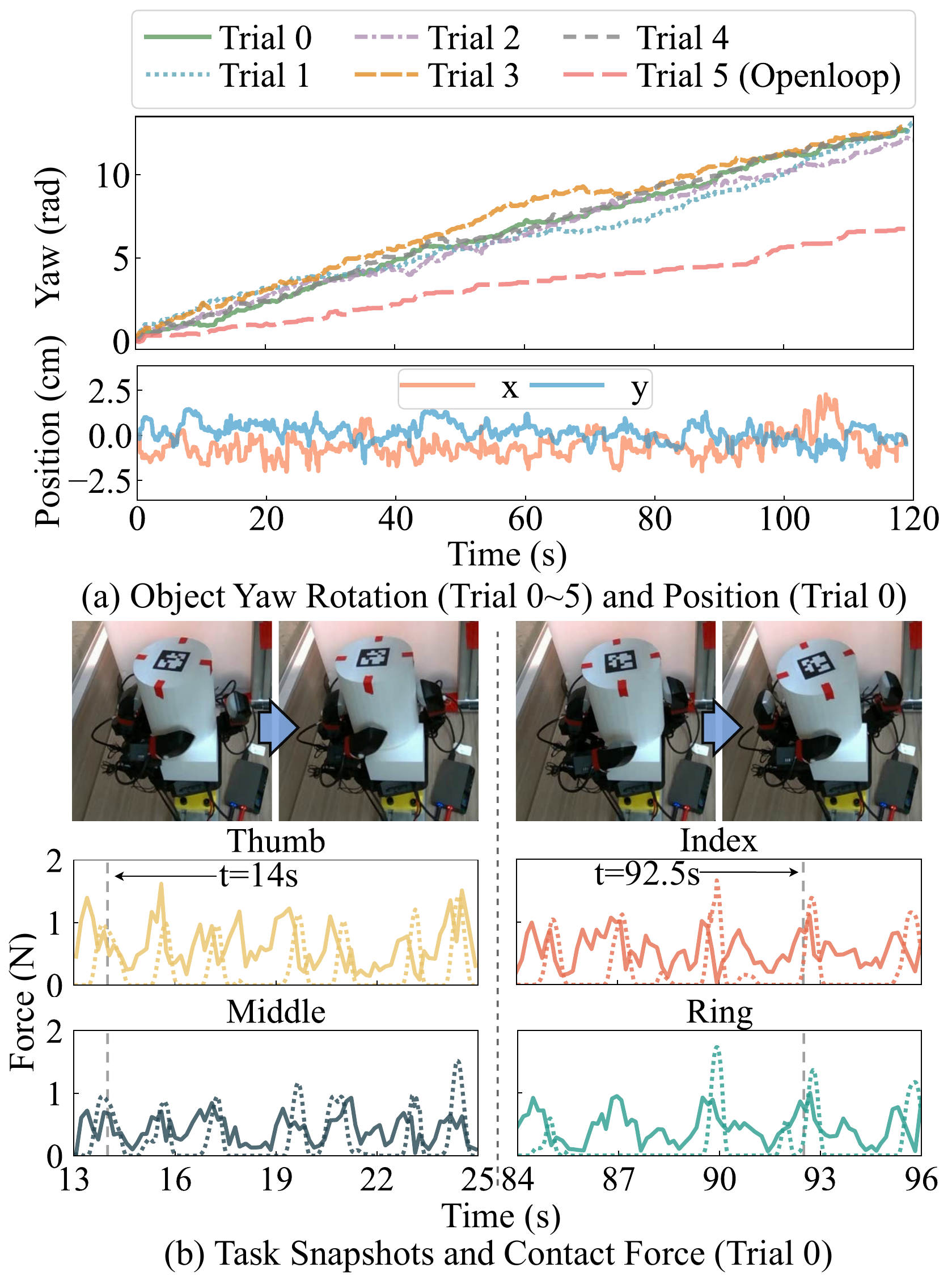}
    \caption{Results of the prehensile in-hand manipulation task with finger gaiting. (a) Object yaw angle over time for all six trials; for Trial 0, the object’s x-y position is also shown. Trial 5 is conducted with the ablation \textbf{Ours (2\textsuperscript{nd}, openloop)} in Tab.~\ref{tab: sim_comparison_sphere_rot_task} without contact tracking. (b) Snapshots and finger contact forces for Trial 0. Solid lines denote desired contact forces, and dashed lines denote measured contact forces.}
    \label{fig: in_hand_manip_cylinder}
    \vspace{-0.3cm}
\end{figure}

\begin{figure}[!t]
    \centering
    \includegraphics[width=1\linewidth]{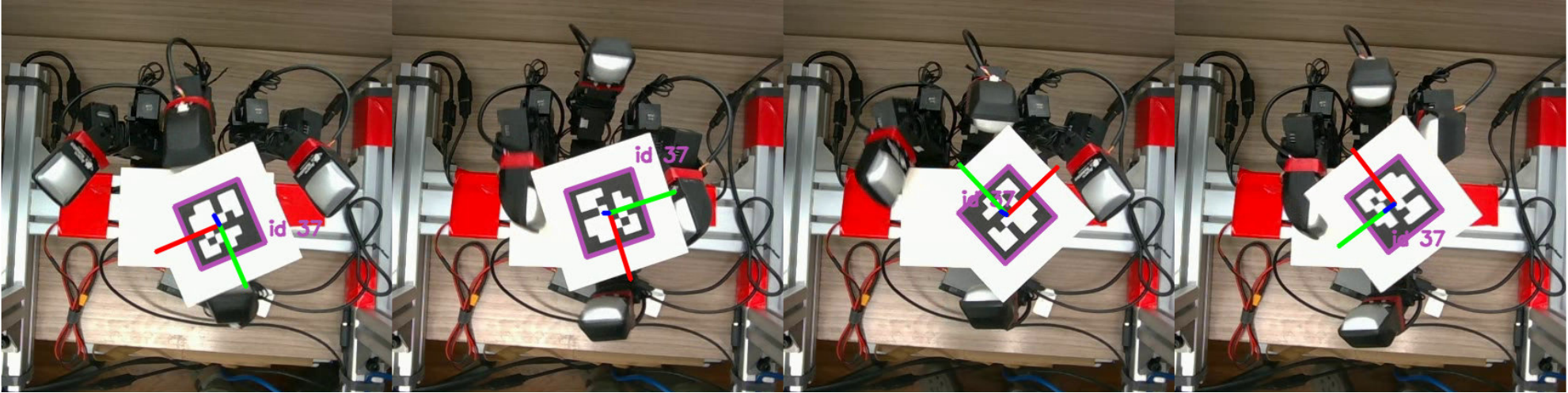}
    \caption{Top-view snapshots of the same in-hand manipulation task as in Fig.~\ref{fig: in_hand_manip_cylinder}, with the object replaced by a cube. The detected marker frame is visualized.}
    \label{fig: in_hand_manip_cube}
    \vspace{-0.3cm}
\end{figure}

We further evaluate the proposed framework on a more challenging prehensile in-hand manipulation task with finger gaiting, as shown in Fig.~\ref{fig: in_hand_manip_cylinder}. The object is continuously rotated through repeated contact breaking, re-establishment, and finger relocation. Active force-balanced grasp is also required to prevent the object from dropping, making this task more demanding than the non-prehensile ones. As shown in Fig.~\ref{fig: in_hand_manip_cylinder}(a), the proposed framework achieves continuous cylinder rotation in Trials 0--4, whereas the pose-tracking-only baseline exhibits much slower rotation in Trial 5 due to frequent contact loss and insufficient contact-force regulation. Fig.~\ref{fig: in_hand_manip_cylinder}(b) further shows a human-like finger-gaiting pattern that maintains force balance during rotation, where two opposing fingers synchronously make and break contact, as indicated by the measured contact forces. We also successfully perform the same task with a cubic object, as shown in Fig.~\ref{fig: in_hand_manip_cube}, where the non-smooth geometry makes contact transitions more challenging.
{Overall, these results demonstrate prehensile in-hand manipulation with automatically generated human-like finger-gaiting, complementing recent model-based advances in dexterous in-hand manipulation~\citep{yang2024contactsdf, suh2025dexterous}.}

\section{Discussion And Conclusion}
\label{sec: discussion_and_conlusion}
\subsection{Discussion}

The proposed bi-level framework for in-hand manipulation has several limitations, which motivate future research. 
1) This framework depends on explicit system modeling, which may not be entirely accurate. Thus, engineering effort is required to provide at least coarse object and hand models for new tasks. Although the proposed method compensates for the modeling errors algorithmically, manipulation performance can be enhanced through system identification.
2) The CQDC model at the high level is based on quasi-dynamic assumptions. Moreover, despite our efforts to achieve a practical control frequency, the current computational speed remains insufficient for highly dynamic manipulation. It is therefore challenging to handle tasks where inertia cannot be neglected and substantially higher planning frequencies are required. {Accordingly, our use of “real-time” refers to online replanning for the quasi-dynamic manipulation considered in this work, rather than high-speed or highly dynamic manipulation.} 3) The gradient-based MPC is well known to suffer from local optimality. Moreover, the solver at each time step is terminated before convergence to {maintain the online planning rate}.
Therefore, the proposed method may adapt finger gaiting patterns slowly under frequent disturbances.
Possible solutions include combination with sampling-based methods and more advanced solvers.

In addition, the imperfect properties of real-world hardware may limit the manipulation performance.
In this work, we choose the low-cost LEAP Hand as the hardware platform. We observe several deficiencies with the LEAP Hand. 
First, it has imperfect mechanical kinematics, as there exist slight passive movements that are not along the joint axis.
Second, the servo has imperfect characteristics. The transmission creates friction that resists changes in joint positions, which affects the contact force control. Third, gravity affects the joint positions due to the backlash and controller compliance. Despite these deficiencies, the proposed method still completes all the hardware tasks, demonstrating robustness against modeling errors. Moreover, the tactile sensors restrict the types of tasks that can be completed. As Tac3D provides only finger pulp touch, tasks that require the side and back of the fingertip are avoided. A possible solution is to consider omnidirectional sensors, such as Digit 360 \citep{lambeta2024digit360}.
{Finally, the current real-world implementation uses AprilTags and an RGB-D camera for object-pose feedback; general marker-free perception is beyond the scope of this work.}

\subsection{Conclusion}
This paper presents a systematic model-based solution for in-hand manipulation that incorporates tactile feedback through a bi-level architecture. The high-level module ensures task-level optimality by {inferring contact events and finger-gaiting patterns online}, while the low-level module provides contact-level stability through tactile-informed motion corrections toward the desired contacts.
{The two modules complement each other and enable in-hand manipulation with strong robustness, precision, and adaptability across diverse tasks, particularly under real-world challenges, as demonstrated in experiments conducted with explicit object-pose feedback.}

Specifically, at the high level, we formulate a unified optimization framework with implicit contact modeling that generates reference finger motions and desired contacts from the target object motion. At the low level, we formulate an MPC-based HFMC scheme that leverages tactile feedback to track the unified motion–contact plans, thereby enhancing robustness against modeling errors. In addition, we introduce several key innovations—warm-starting, trajectory shifting, multi-contact coupling, and adaptive weighting—that effectively coordinate the two levels and multiple tracking objectives.

Extensive simulation and real-world experiments show that our method achieves robust and accurate long-horizon in-hand manipulation, even under disturbances, modeling errors, and sensor noise. These challenges cannot be addressed by improved contact modeling or optimization alone, highlighting the necessity of tactile incorporation. By adopting a bi-level formulation, we enable indirect tactile incorporation in contact-implicit MPC, resolving inherent incompatibilities without disrupting the underlying optimization.
{Beyond these robustness gains, the proposed framework also retains a key model-based advantage: flexible online adaptation to new task specifications without policy retraining, given suitable object and hand models.}

This work aims to provide insights and strives to push the limits of model-based methods for dexterous manipulation. {Future work will include replacing the implicit contact model with learned contact-based dynamics models to further improve the control frequency and enable more dynamic motions.}
We are also planning to incorporate arm-hand coordination and advanced grasping techniques to fully automate the manipulation process.

\begin{dci}
The author(s) declared no potential conflicts of interest with respect to the research, authorship, and/or publication of this article.
\end{dci}

\begin{funding}
This work was supported in part by the Fundamental and Interdisciplinary Disciplines Breakthrough Plan of the Ministry of Education of China under Grant JYB2025XDXM208, in part by the National Natural Science Foundation of China under Grant 62461160307, and in part by Tsinghua University Initiative Scientific Research Program.
\end{funding}

\bibliographystyle{SageH} 
\bibliography{ref}

\section*{Appendix A. Index to multimedia extensions}
\label{sec: appendix_a}
The supplementary video is available on the project website \url{https://director-of-g.github.io/in_hand_manipulation_2/}.

\section*{Appendix B. Additional simulation results}
\label{sec: appendix_b_additional_simulation}

\subsection*{{Unified} Motion-Contact Planning with Contact-Implicit MPC}
\label{subsec: motion_contact_planning_exp}
\begin{figure}[!b]
    \centering
    \vspace{-0.5cm}
    \includegraphics[width=0.95\linewidth]{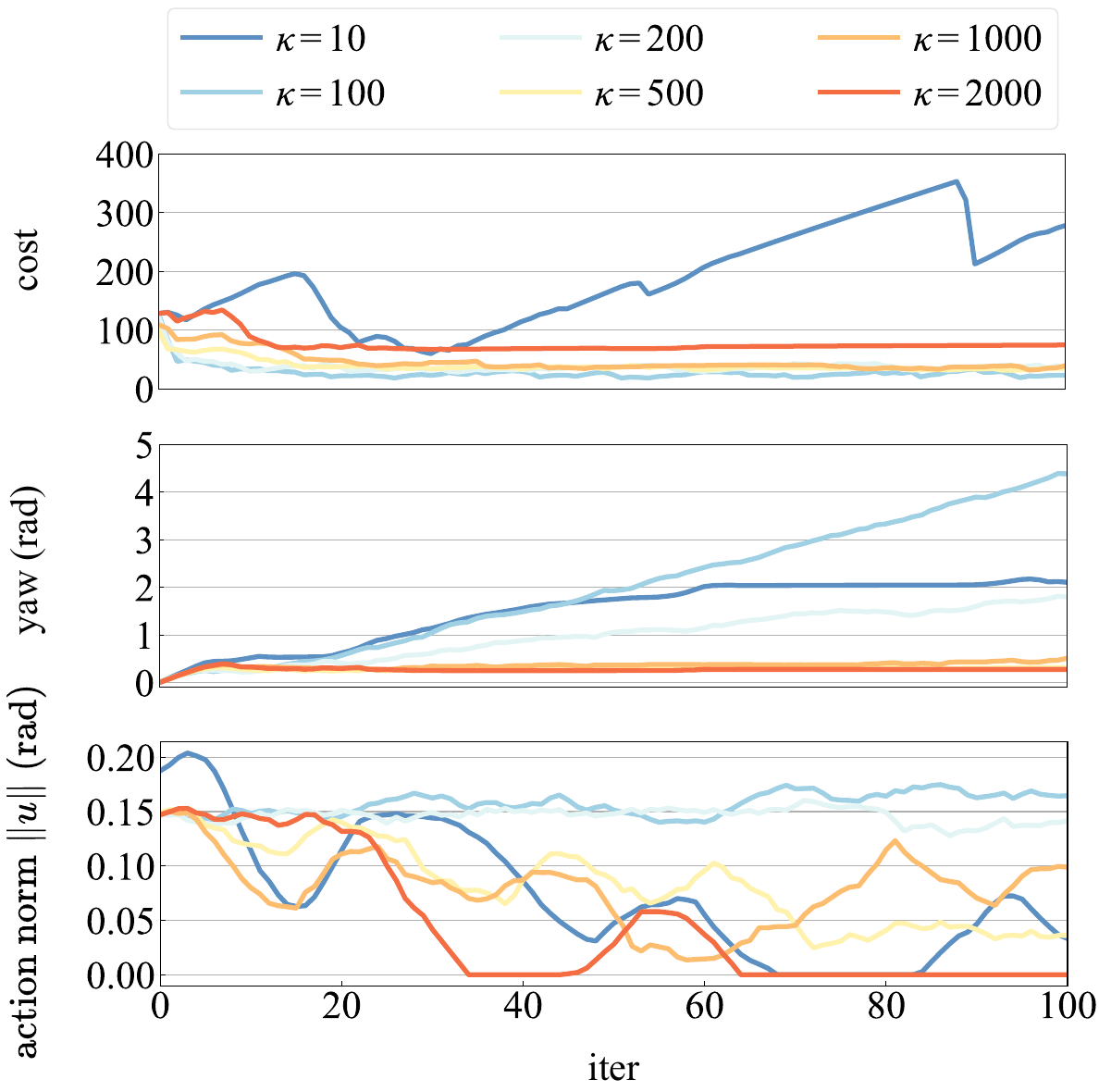}
    \vspace{-0.3cm}
    \caption{Effects of $\kappa$ on optimization performance, including the cost value, the door handle's rotation (yaw angle), and the action norm. Warmer colors indicate a weaker smoothing effect and less nonphysical behavior.}
    \label{fig: high_different_smoothing_kappa}
\end{figure}

{In this section, we solely evaluate the performance of the high-level motion-contact planner, where  the same quasi-dynamic model is used in both planning and simulation.}
The high-level contact-implicit MPC is implemented with the Box-DDP solver in Crocoddyl \citep{mastalli20crocoddyl}.
We choose an MPC horizon $N=10$ and a maximum solver iteration of $2$ to guarantee an adequate solution speed. Moreover, we set $h=0.1s$ and $\kappa=100$ in the CQDC model (\ref{eq: CQDC_model_barrier}). The choice of $\kappa$ is informed by the empirical analysis presented at the end of this subsection (i.e., Fig.~\ref{fig: high_different_smoothing_kappa}).

\begin{figure}[!b]
    \centering
    \vspace{-0.5cm}
    \includegraphics[width=0.85\linewidth]{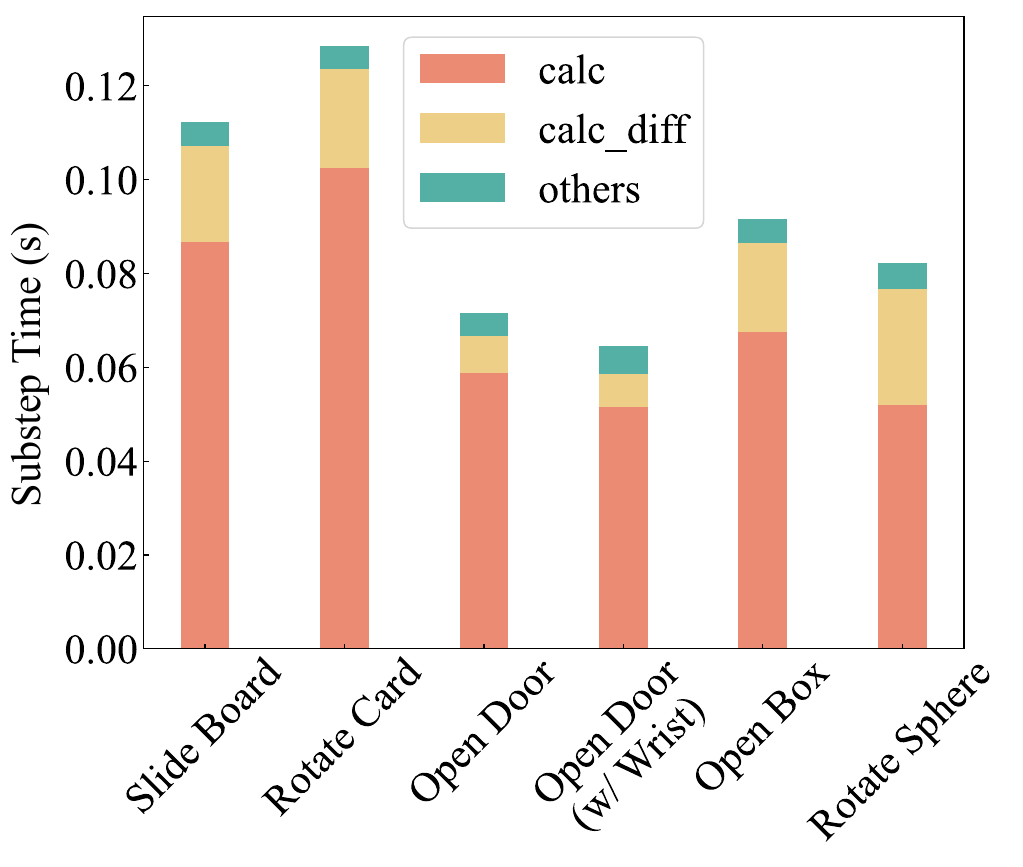}
    \vspace{-0.3cm}
    \caption{Elapsed time of high-level motion-contact planning during key sub-steps. Here, calc and calc\_diff represent the forward dynamics and gradient computation of the CQDC model, respectively. The total height of the bar indicates the average iteration time for each task.}
    \label{fig: high_substep_time}
\end{figure}

\textbf{Solution speed for different tasks}:
The elapsed time for different tasks during key sub-steps is shown in Fig.~\ref{fig: high_substep_time}.
The results represent the time taken for a single iteration, which is averaged over the manipulation sequence. The major time-consuming processes are the forward dynamics \textbf{calc} and the gradient computation \textbf{calc\_diff}. As DDP needs multiple rollouts to determine the proper step size, and it is difficult to truly parallelize these rollouts due to Python's Global Interpreter Lock mechanism \citep{crocoddyl_issue_1042}, the computation of forward dynamics accounts for most of the iteration time. The planning of more complex systems (i.e., with larger $n_x$) tends to take longer time in each iteration.
However, our selected tasks can run at approximately 8 to 15 Hz, {which is sufficient for online replanning in the considered quasi-dynamic settings}.
In real-world experiments, the low-level controller reactively refines the planned motion while synergistically ensuring the intended contacts.

\begin{figure}[!t]
    \centering
    \includegraphics[width=0.95\linewidth]{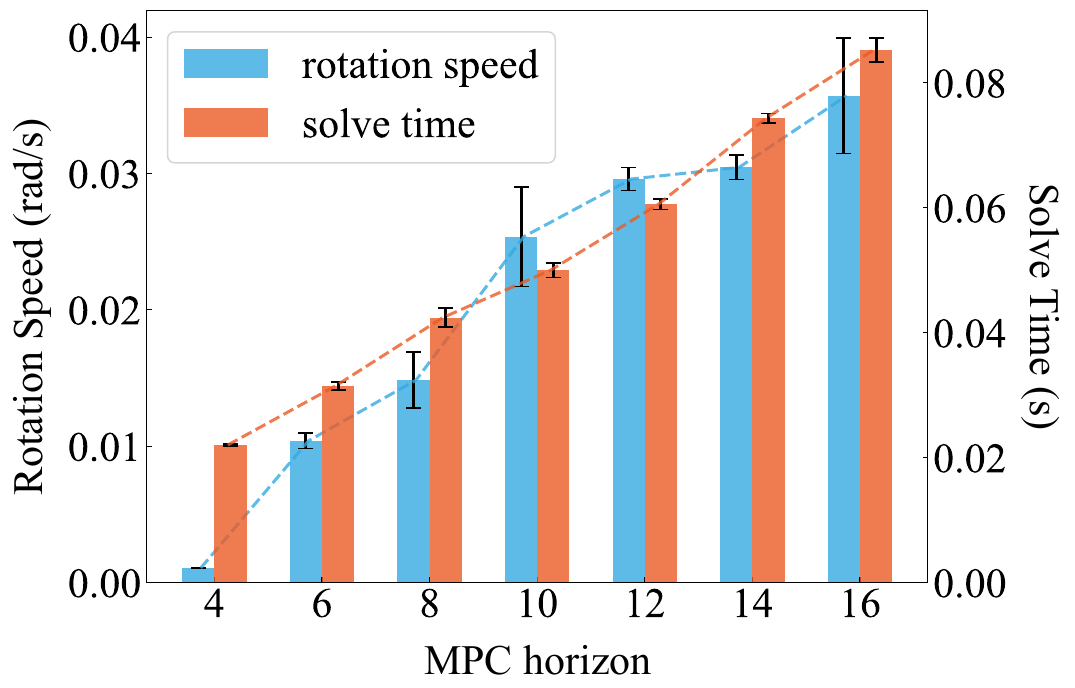}
    \vspace{-0.3cm}
    \caption{Rotation speed and solution time per iteration across MPC horizons in the \textbf{Rotate Valve} task. The error bars indicate standard deviations. The dashed lines represent average values.}
    \vspace{-0.5cm}
    \label{fig: high_stats_t_ctrl}
\end{figure}

\textbf{Effects of the MPC Horizon}:
The \textbf{Rotate Valve} task is used to study the effect of the selected MPC horizon. The fingers need to switch between different movement patterns, such as stepping over and pushing the valve, to accomplish the task. The average rotation speed and the solution time per iteration are shown in Fig.~\ref{fig: high_stats_t_ctrl}. The valve hardly rotates under shorter horizons $N \leq 4$, since the fingers push the valve without stepping over and regrasping it, which indicates that shorter horizons lack predictive ability and readily lead to local optima. On the one hand, the rotation speed increases with longer horizons, which indicates improved solution quality. On the other hand, the solution time increases linearly with the MPC horizon, leading to a reduction in control frequency. We therefore choose $N=10$ for all experimental tasks as a practical trade-off between rotation speed and the computational cost of online replanning.

\textbf{Effects of Smoothness in Forward Dynamics}:
{We investigate the effects of the smoothness factor $\kappa$ in the \textbf{Open Door} task.
Different $\kappa$ values are used in the forward dynamics computation (\ref{eq: CQDC_model_barrier}) to achieve varying smoothness levels.
As shown in Fig.~\ref{fig: high_different_smoothing_kappa}, excessively large values of $\kappa$ ($\kappa \geq 500$) result in less smoothing and minimal door handle rotation due to overly conservative DDP step sizes. This highlights the importance of the smoothed force-at-a-distance effect in planning.
Conversely, very small $\kappa$ ($\kappa=10$) results in excessive smoothing, which enhances nonphysical behavior and leads to poor performance. An appropriate $\kappa$ ($\kappa = 100$) facilitates stable finger-gaiting and a consistent rotation speed.
We thus demonstrate that high-level modeling errors due to smoothing cannot be corrected using more realistic forward dynamics alone. This underscores the necessity of {incorporating tactile feedback for motion correction}, which compensates for high-level errors and improves manipulation performance, as shown in Sec.~\ref{subsec: highlevel_comparison_baselines} and Sec.~\ref{subsec: robustness_noise_modeling_error}.}

\subsection*{{Handling Complex Object Geometries}}
\label{subsec: hand_complex_geometry}

\begin{figure}[!t]
    \centering
    \includegraphics[width=0.95\linewidth]{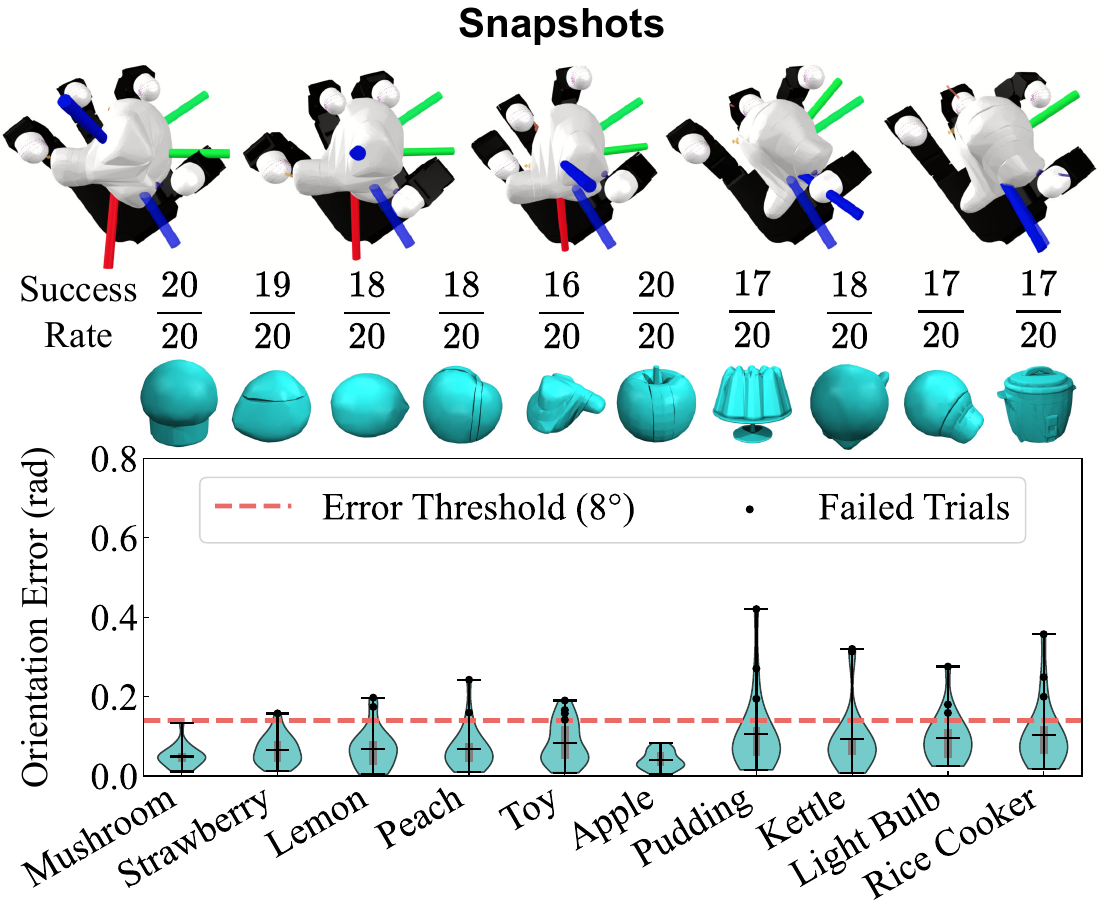}
    \caption{Additional results of the simulated \textbf{Rotate Sphere} task regarding complex geometries. The sphere is replaced with 10 complex objects from the DGN's \citep{wang2023dexgraspnet} dataset and rotated in place toward 20 target orientations. Task snapshots, the object geometries with decomposed convex parts, their associated success rate and error distributions are visualized. Most trials achieve precise manipulation with orientation errors below an 8° threshold.}
    \vspace{-0.5cm}
    \label{fig: any_object_rpy_plot}
\end{figure}

Manipulating objects with complex geometries poses challenges due to degraded modeling accuracy and increased computational complexity. As regards modeling accuracy, we perform convex decomposition to approximate arbitrary object geometries as faithfully as possible. In addition, tactile feedback is leveraged to update the geometry in real time, ensuring that contacts are established as planned even without an accurate prior model. As regards the computational complexity, we simplify the dynamics model by keeping one contact pair with the minimum $\phi$ for each geometry. In addition, existing robotics kinematics libraries, including Drake and Pinocchio, do not reliably provide contact Hessians $\frac{\partial \bm{J}}{\partial \bm{x}}$ for arbitrary object geometries. To overcome this limitation, we approximate the contact Hessian using numerical differentiation:
\begin{equation}
    \frac{\partial \bm{J}}{\partial x_j} \approx \frac{\bm{J}(\bm{x}\oplus\Delta\cdot\bm{e}_j)-\bm{J}(\bm{x})}{\Delta}
    \label{eq: contact_Hessian_by_finitediff}
\end{equation}
where $\bm{e}_j$ is the $j$-th unit vector in the local coordinates and $\Delta$ is a small perturbation. The Jacobian matrix is sparse due to the tree-like mechanical structure of the dexterous hand. These treatments enable a reasonable control frequency (i.e., close to 10Hz) when handling complex objects with tens of convex decompositions.

{We evaluate performance of the proposed method by replacing the sphere with 10 complex objects in the \textbf{Rotate Sphere} task, as shown in Fig.~\ref{fig: any_object_rpy_plot}. These objects are selected from DGN's dataset \citep{wang2023dexgraspnet}, ensuring their aspect ratio remains close to 1:1:1. In addition, we use the first 20 target orientations in Sec.~\ref{subsec: highlevel_comparison_baselines} for evaluation. The proposed method achieves robust and accurate performance when handling complex geometries, with most data points falling below the error threshold.}

\begin{figure}[!t]
    \centering
    \includegraphics[width=0.9\linewidth]{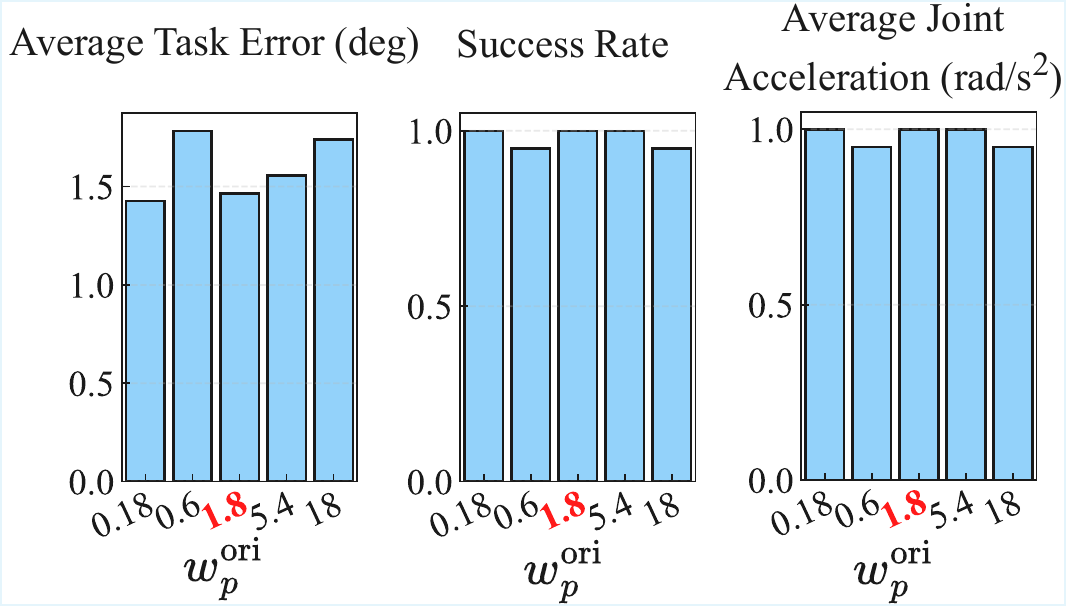}
    \caption{Parameter sweep results for the fingertip orientation tracking weight $w_p^{\text{ori}}$, with the best-performing $w_p$ and $w_\Lambda$ in Fig.~\ref{fig: param_sweep_pos_force} fixed. All three metrics show no significant change as $w_p^{\text{ori}}$ varies, and a moderate value of 1.8 is selected.}
    \vspace{-0.5cm}
    \label{fig: param_sweep_ori}
\end{figure}

\begin{figure*}[!t]
    \centering
    \includegraphics[width=0.9\linewidth]{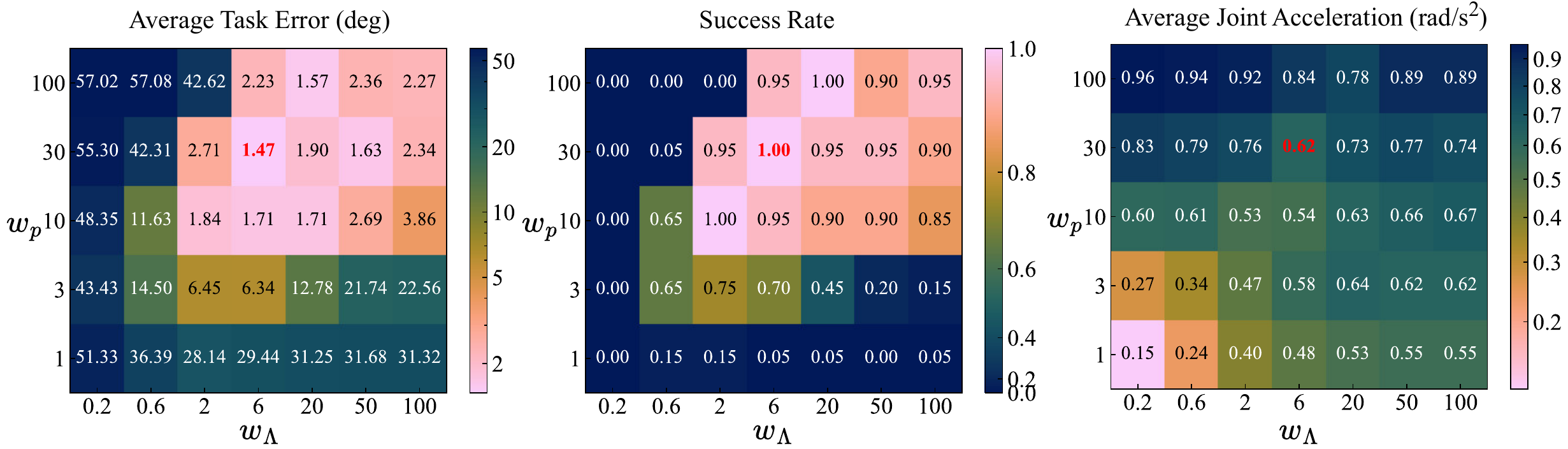}
    \caption{Heatmap of the parameter sweep on the \textbf{Rotate Sphere} task, where $w_p$ and $w_\Lambda$ denote the scalar weights for fingertip position tracking and contact force tracking, respectively, in the fixed-weight baseline. The selected parameter combination, $w_p=30$ and $w_\Lambda=6$, which yields the lowest task error, the highest success rate, and moderate joint acceleration, is marked in red.}
    \vspace{-0.5cm}
    \label{fig: param_sweep_pos_force}
\end{figure*}

\section*{Appendix C. Fixed-Weight Baseline and Parameter Sweep for Comparison with Adaptive Weighting}
\label{sec: appendix_c_param_sweep_fix_weight}

To further examine the necessity and benefit of the proposed adaptive weighting design, we compare it with a fixed-weight baseline in Sec.~\ref{subsec: comparison_weighting_scheme}. This appendix presents the formulation of the fixed-weight scheme and the corresponding parameter sweep used to determine its best-performing setting for a fair comparison.

For the fixed-weight baseline, the weighting matrices in the low-level MPC of~(\ref{eq: low_level_MPC}) are set to constant isotropic forms:
\begin{equation}
\bm{W}_{\bm{\Lambda}} = w_\Lambda \bm{I}, \qquad
\bm{W}_{\bm{p}} =
\begin{bmatrix}
w_p \bm{I} & \bm{0} \\
\bm{0} & w_p^\text{ori}\bm{I}
\end{bmatrix},
\end{equation}
where $w_\Lambda$, $w_p$, and $w_p^\text{ori}$ are scalar constants that remain unchanged throughout execution. In contrast to the proposed adaptive weighting scheme, this baseline uses a single fixed compromise between force tracking and motion tracking.

To ensure a fair comparison, we tune the fixed-weight baseline by parameter sweeping on the \textbf{Rotate Sphere} task. A subset of 20 target orientations is used for tuning, sampled from the same distribution as the 100 target orientations adopted in the full evaluation. We evaluate each parameter setting using average task error, success rate, and average joint acceleration. The sweeping results are shown in Fig.~\ref{fig: param_sweep_ori}--\ref{fig: param_sweep_pos_force}.

The parameter sweep suggests that this task requires a proper trade-off between position and force tracking, while overly large weights tend to reduce execution smoothness. Based on the above three metrics jointly, we select
\[
w_p = 30, \qquad w_\Lambda = 6, \qquad w_p^\text{ori} = 1.8
\]
as the best-performing setting for the fixed-weight baseline. This configuration is then compared with the proposed adaptive weighting scheme reported in Fig.~\ref{fig: comparison_weight_scheme}.

\begin{figure*}[!b]
    \centering
    \vspace{-0.3cm}
    \includegraphics[width=1.0\linewidth]{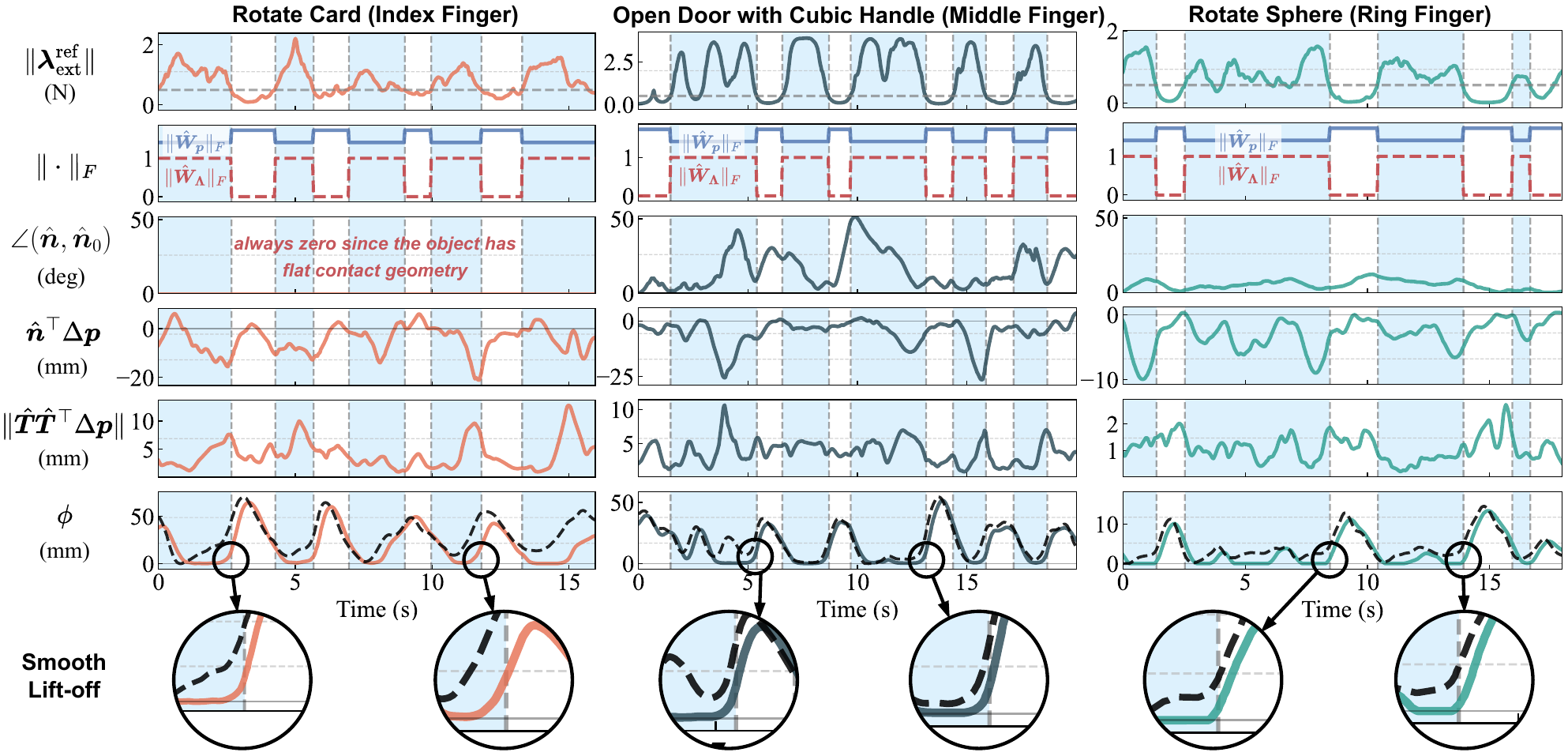}
    \caption{Key variables of the adaptively weighted low-level controller during three representative tasks. From top to bottom: desired contact force magnitude $\Vert \bm{\lambda}_{\mathrm{ext}}^{\mathrm{ref}} \Vert$ (dashed line: threshold $\underline{\bm{\lambda}}$; blue shading: $\Vert \bm{\lambda}_{\mathrm{ext}}^{\mathrm{ref}} \Vert\geq \underline{\bm{\lambda}}$); $\Vert \hat{\bm{W}}_{\bm{p}} \Vert_F$ and $\Vert \hat{\bm{W}}_{\bm{\Lambda}} \Vert_F$, Frobenius norm of the adaptive weights respectively for tracking fingertip pose and contact force; $\angle(\hat{\bm{n}}, \hat{\bm{n}}_0)$, angle in degrees between the current contact normal and its initial value at $t=0$; $\hat{\bm{n}}^\top \Delta \bm{p}$, signed position tracking error along the contact normal; $\Vert \hat{\bm{T}}\hat{\bm{T}}^\top \Delta \bm{p} \Vert$, tangential position tracking error; and contact distance $\phi$ (dashed curve: contact distance when merely following the high-level reference). Here, $\Delta \bm{p}$ is the translational component of $\mathrm{FK}(\bm{q}) \ominus \mathrm{FK}(\bm{q}_{\mathrm{ref}})$.}
    \vspace{-0.5cm}
    \label{fig: plot_hfmc_how_it_works_2}
\end{figure*}

\section*{Appendix D. Further illustration of the adaptive weighting mechanism}
\label{sec: appendix_c_illustration_ada_weighting}

To illustrate the mechanism of the proposed adaptively weighted low-level controller, Fig.~\ref{fig: plot_hfmc_how_it_works_2} visualizes the weight evolution in representative executions of three tasks. We first revisit the mathematical formulation of the adaptive weights:

\begin{equation}
\hat{\bm{W}}_{\bm{\Lambda}} =
    \begin{cases}
    \bm{0}, & \left\lVert \bm{\lambda}_{\mathrm{ext}}^{\mathrm{ref}} \right\rVert < \underline{\bm{\lambda}}, \\[4pt]
    \hat{\bm{n}}\hat{\bm{n}}^\top, & \left\lVert \bm{\lambda}_{\mathrm{ext}}^{\mathrm{ref}} \right\rVert \geq \underline{\bm{\lambda}},
    \end{cases}
\end{equation}
\begin{equation}
\hat{\bm{W}}_{\bm{p}} =
    \begin{cases}
        \begin{bmatrix}
            \bm{I} & \bm{0} \\
            \bm{0} & w_{\mathrm{ori}}\bm{I}
        \end{bmatrix},
    & \left\lVert \bm{\lambda}_{\mathrm{ext}}^{\mathrm{ref}} \right\rVert < \underline{\bm{\lambda}}, \\[10pt]
        \begin{bmatrix}
            \hat{\bm{T}}\hat{\bm{T}}^\top & \bm{0} \\
            \bm{0} & w_{\mathrm{ori}}\bm{I}
        \end{bmatrix},
    & \left\lVert \bm{\lambda}_{\mathrm{ext}}^{\mathrm{ref}} \right\rVert \geq \underline{\bm{\lambda}}.
    \end{cases}
\end{equation}

In Fig.~\ref{fig: plot_hfmc_how_it_works_2}, the Frobenius norms of $\hat{\bm{W}}_{\bm{p}}$ and $\hat{\bm{W}}_{\bm{\Lambda}}$ quantify the weight magnitudes, while the angle between the current contact normal and its initial value at $t=0$ captures the directional evolution of the weighting subspaces. To relate the weight evolution to contact events and low-level behavior, we also plot the desired contact force magnitude $\Vert \bm{\lambda}_{\mathrm{ext}}^{\mathrm{ref}} \Vert$, the signed normal error, the tangential error, and the contact distance $\phi$. Blue shading denotes hybrid force-motion tracking; unshaded intervals denote position-dominant tracking.

The results show that the weight adaptation is closely related to contact events. When the high-level planner indicates a contact intention by setting $\Vert \bm{\lambda}_{\mathrm{ext}}^{\mathrm{ref}} \Vert$ above the threshold $\underline{\bm{\lambda}}$, the controller switches to hybrid tracking: force tracking is prioritized along the contact normal, while position tracking is retained in the tangential subspace. This switch is reflected by the Frobenius norms; within the hybrid phase, the weighting direction further adapts with the contact normal, as shown by $\angle(\hat{\bm{n}}, \hat{\bm{n}}_0)$. Negative normal errors indicate that the controller moves the fingertip closer to the object than the high-level reference to establish the desired contact force. This facilitates contact establishment and maintenance, with $\phi$ approaching and remaining near zero. Conversely, when the desired force falls below the threshold, the controller returns to position-dominant tracking for contact breaking and finger relocation, with $\phi$ increasing accordingly.

Across the three tasks, the overall switching pattern is consistent, but the detailed evolution is task-dependent. In particular, the normal-adaptive component changes mildly for planar contacts, while larger variations appear for curved or non-smooth geometries. These observations show that the adaptive weighting is not merely a mathematical formality; it produces interpretable, task-dependent changes that correlate with contact establishment, maintenance, and breaking during execution.

\section*{Appendix E. Additional real-world experiment results}
\label{sec: appendix_d_additional_real_world}

\begin{figure}[!t]
    \centering
    \includegraphics[width=0.95\linewidth]{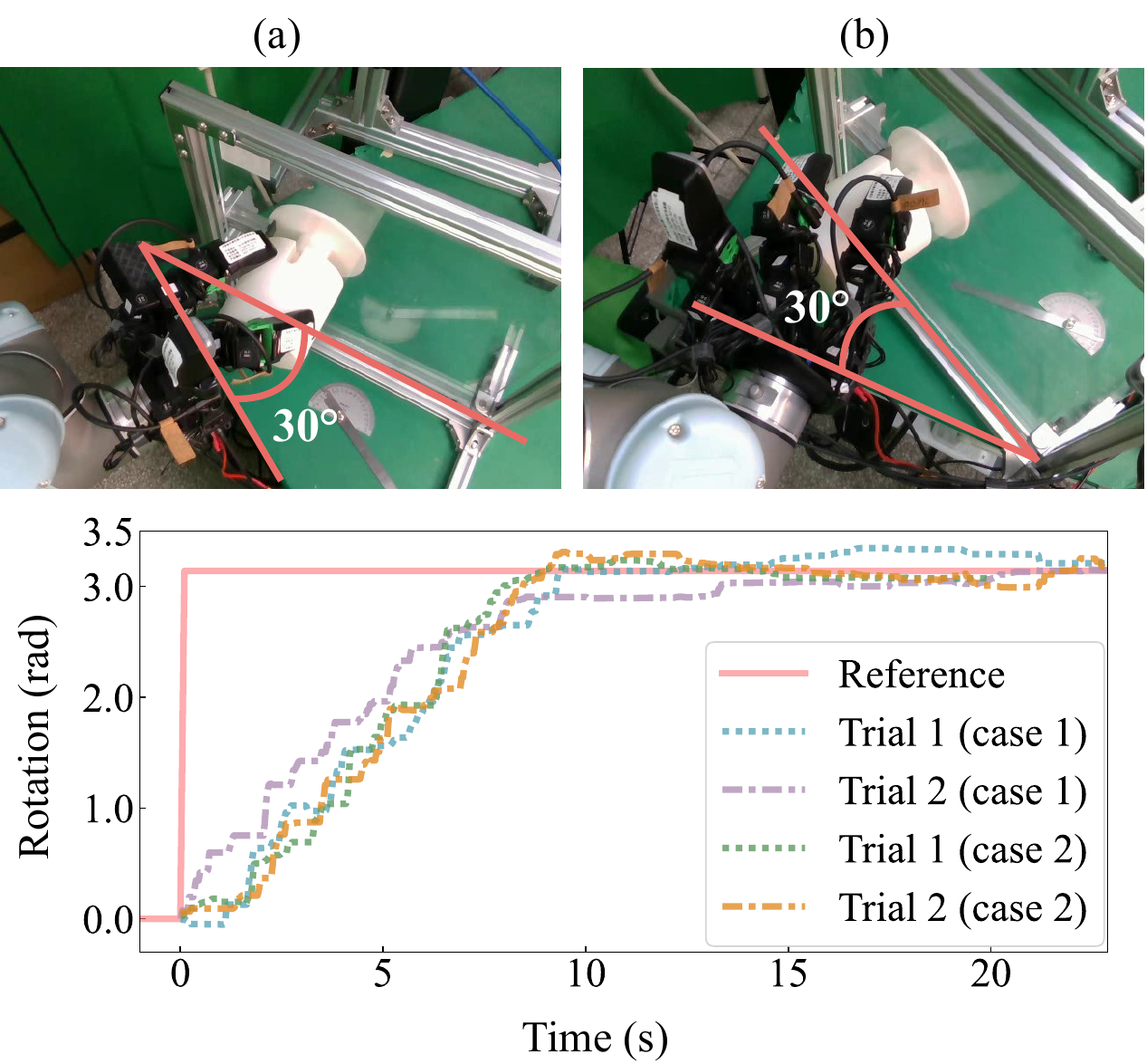}
    \caption{Additional results of the open door task with different relative wrist-door poses. (a) The door is rotated 30 degrees counterclockwise (case 1). The ring finger is not used. (b) The door is rotated 30 degrees clockwise (case 2). The index finger is not used. (c) Door handle rotation in different trials.}
    \vspace{-0.4cm}
    \label{fig: open_door_left_right_real_results}
\end{figure}

\begin{figure*}[!b]
    \centering
    \vspace{-0.4cm}
    \includegraphics[width=0.95\linewidth]{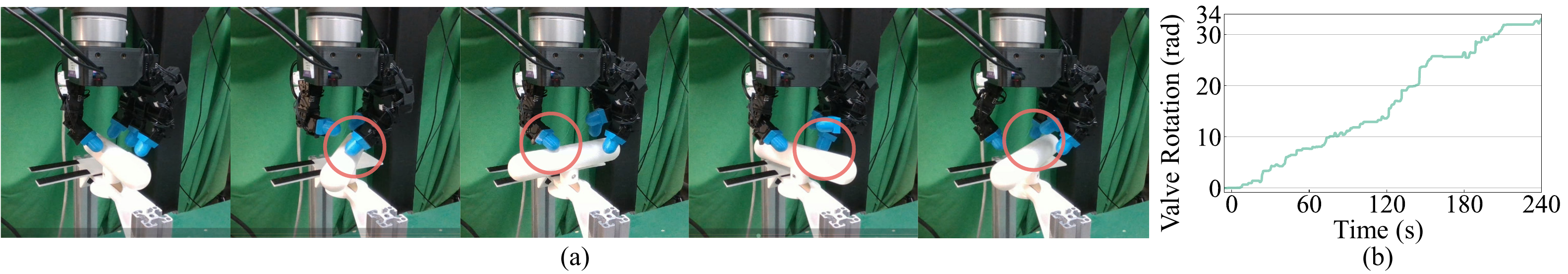}
    \caption{(a) Snapshots of the Rotate Valve task. The red circles mark the fingers in contact. The fingers detect contacts within their reachable space, sequentially touching the valve. Due to the limited sensing regions of the Tac3D tactile sensors, we replace them with spherical fingertips and assume zero contact forces at all fingertips to enable the proposed method. However, this setup does not fully demonstrate the low-level module’s effectiveness in contact tracking. (b) Valve rotation angle relative to time.}
    \label{fig: rotate_valve_real_results}
\end{figure*}

\subsection*{Parameter Selection}

We discuss key parameters used in the real-world experiments. The parameters of the high-level module are set as $h=\SI{0.1}{s}, \kappa=100, N=10$, $\beta_1=2.0$, and $\beta_2=0.75$.
{The high-level cost weights follow the selection procedure in Sec.~\ref{subsec: regulation_cost}, with \(\bm{W}_{\bm{u}}=0.04\bm{I}\) fixed across all tasks; the exact settings used in the experiments are provided with the released code on the project website.}
{The proposed framework does not rely on accurate calibration of dynamic parameters such as mass and friction, demonstrating robustness to model-parameter uncertainty.}
Moreover, the friction coefficient of each contact is set as $\mu_i=1$.
The parameters of the low-level module are set as $\bm{K}_P=4\bm{I}, \bm{K}_D=\bm{I}, k_e=200, k_{\bm{\Lambda}}=0.2, k_{\bm{p}}=10^4, w_\text{ori}=3 \times 10^{-4}, k_{\bm{q}}=20, k_{\bm{u}}=1, T=\SI{0.3}{s}, \Delta t=\SI{0.03}{s}$, and $\underline{\bm{\lambda}}=\SI{0.5}{N}$. To avoid excessive contact forces, the magnitude of the high-level reference force $\Vert \bm{\lambda} \Vert$ is softly thresholded by the function
\begin{equation}
    L\left(\frac{2}{1+e^{-k\Vert\bm{\lambda}\Vert}}-1\right)
    \label{eq: high_force_soft_threshold}
\end{equation}
where $k=0.04, L=2$.

\subsection*{Results of the Open Door Task with Modified Task Setups}

Next, we evaluate the method's capability to adapt to modified task setups.
We change the placement of the whole door by rotating it along the vertical axis 30 degrees counterclockwise
(case 1, Fig.~\ref{fig: open_door_left_right_real_results} (a)) or clockwise (case 2, Fig.~\ref{fig: open_door_left_right_real_results} (b)) while fixing the robot wrist pose. 
We perform the Open Door task twice for each case.
The door handle rotation in different trials is shown in Fig.~\ref{fig: open_door_left_right_real_results} (c). The results show that the handle rotates at almost the same speed and converges to the target orientation under different conditions. 
{This demonstrates the model-based framework's ability to directly adapt to modified task setups through online replanning.}

\subsection*{Demonstration of the Rotate Valve Task}

In Fig.~\ref{fig: rotate_valve_real_results}, we demonstrate the proposed method's capability in executing long-horizon tasks through the \textbf{Rotate Valve} experiment. The valve is continuously rotated via finger gaiting, with all four fingers detecting contacts within their reachable space and sequentially manipulating the valve (Fig.~\ref{fig: rotate_valve_real_results} (b)). This finger gaiting behavior emerges automatically without predefined contact sequences or prior task knowledge. The Tac3D tactile sensor we use provides tactile readings only on the finger pulp, whereas the planned contacts naturally occur on the backs of the fingers, where no tactile sensing is available. This limits the demonstration of the low-level controller’s effectiveness. As shown in Fig.~\ref{fig: rotate_valve_real_results} (a), we use spherical fingertips instead of Tac3D sensors to facilitate manipulation. Assuming zero contact forces from all fingertips encourages contact establishment but results in excessive forces, underscoring the need for our low-level module for effective contact tracking.

\end{document}